\documentclass{article}

\usepackage[accepted]{icml2020}

\icmltitlerunning{Multi-step Greedy Reinforcement Learning Algorithms}

\usepackage[utf8]{inputenc} 
\usepackage[T1]{fontenc}    
\usepackage{hyperref}       
\usepackage{url}            
\usepackage{booktabs}       
\usepackage{amsfonts}       
\usepackage{nicefrac}       
\usepackage{microtype}      
\usepackage{comment}
\usepackage{subcaption}
\usepackage{booktabs} 

\usepackage{times}
\usepackage{url}
\usepackage{algorithm}
\usepackage{algorithmic}
\usepackage{graphicx} 
\usepackage{mathtools}
\usepackage{amsmath}
\usepackage{amssymb}
\usepackage{amsthm} 
\usepackage{thmtools}
\usepackage{hyperref}
\usepackage{float}
\usepackage[export]{adjustbox}
\usepackage{xcolor,colortbl}
\usepackage{amsfonts}
\usepackage{bm}
\usepackage{xpatch}
\usepackage[symbol]{footmisc}
\usepackage{pifont}
\definecolor{Grey}{rgb}{0.91, 0.91, 0.94}

\renewcommand{\cite}[1]{\citep{#1}}
\newtheorem{remark}{Remark}

\usepackage{amsmath,amsfonts,bm}

\def\eqref#1{equation~\ref{#1}}

\def\1{\bm{1}}

\DeclareMathAlphabet{\mathsfit}{\encodingdefault}{\sfdefault}{m}{sl}
\SetMathAlphabet{\mathsfit}{bold}{\encodingdefault}{\sfdefault}{bx}{n}

\newcommand{\E}{\mathbb{E}}

\newcommand*\norm[1]{\left\|#1\right\|}

\DeclareMathOperator*{\argmax}{arg\,max}

\usepackage{color}

\usepackage{multirow}
\def\Vmax{V_{\mathrm{max}}}

\begin{document}

\twocolumn[
\icmltitle{Multi-step Greedy Reinforcement Learning Algorithms}



\icmlsetsymbol{equal}{*}

\begin{icmlauthorlist}
\icmlauthor{Manan Tomar}{equal,to}
\icmlauthor{Yonathan Efroni}{equal,goo}
\icmlauthor{Mohammad Ghavamzadeh}{ed}
\end{icmlauthorlist}

\icmlaffiliation{to}{Facebook AI Research, Menlo Park, USA}
\icmlaffiliation{goo}{Technion, Haifa, Israel}
\icmlaffiliation{ed}{Google Research, Mountain View, USA}

\icmlcorrespondingauthor{Manan Tomar}{manan.tomar@gmail.com}

\icmlkeywords{Reinforcement Learning, Multi-step Greedy Policies}

\vskip 0.3in
]

\printAffiliationsAndNotice{\icmlEqualContribution} 

\begin{abstract}
Multi-step greedy policies have been extensively used in model-based reinforcement learning (RL), both when a model of the environment is available (e.g.,~in the game of Go) and when it is learned. In this paper, we explore their benefits in model-free RL, when employed using multi-step dynamic programming algorithms: $\kappa$-Policy Iteration ($\kappa$-PI) and $\kappa$-Value Iteration ($\kappa$-VI). These methods iteratively compute the next policy ($\kappa$-PI) and value function ($\kappa$-VI) by solving a surrogate decision problem with a shaped reward and a smaller discount factor. We derive model-free RL algorithms based on $\kappa$-PI and $\kappa$-VI in which the surrogate problem can be solved by any discrete or continuous action RL method, such as DQN and TRPO. We identify the importance of a hyper-parameter that controls the extent to which the surrogate problem is solved and suggest a way to set this parameter. When evaluated on a range of Atari and MuJoCo benchmark tasks, our results indicate that for the right range of $\kappa$, our algorithms outperform DQN and TRPO. This shows that our multi-step greedy algorithms are general enough to be applied over any existing RL algorithm and can significantly improve its performance.

\end{abstract}

\section{Introduction}
\label{sec:intro}


Reinforcement learning (RL) algorithms solve sequential decision-making problems through repeated interaction with the environment. By incorporating deep neural networks into RL algorithms, the field has recently witnessed remarkable empirical success (e.g.,~\citealp{mnih2015human,lillicrap2015continuous,levine2016end,silver2017mastering}). Much of this success has been achieved by model-free RL algorithms, such as Q-learning and policy gradients. These algorithms are known to suffer from high variance in their estimations~\citep{greensmith2004variance} and to have difficulties in handling function approximation (e.g.,~\citealp{thrun1993issues,Baird95RA,van2016deep,Lu18ND}). These issues are intensified in decision problems with long horizon, i.e.,~when the discount factor, $\gamma$, is large. Although using smaller values of $\gamma$ addresses the discount factor-dependent issues and leads to more stable algorithms~\citep{petrik2009biasing,jiang2015dependence}, it does not come for free, as the algorithm may return a {\em biased} solution, i.e.,~it may not converge to an optimal (or good) solution for the original decision problem (the one with larger value of $\gamma$).

\citet{efroni2018beyond} recently proposed another approach to mitigate the $\gamma$-dependant instabilities in RL in which they study multi-step greedy versions of the well-known Dynamic Programming (DP) algorithms: Policy Iteration (PI) and Value Iteration (VI)~\citep{bertsekas1996neuro}. They also proposed an alternative formulation of the multi-step greedy policy, called $\kappa$-greedy policy, and studied the convergence of the resulted PI and VI algorithms: $\kappa$-PI and $\kappa$-VI. These algorithms iteratively solve a $\gamma\kappa$-discounted decision problem, whose reward has been shaped by the solution of the decision problem at the previous iteration. Unlike the {\em biased} solution obtained by solving the decision problem with a smaller value of $\gamma$ (discussed above), by iteratively solving decision problems with a shorter $\gamma\kappa$ horizon, the $\kappa$-PI and $\kappa$-VI algorithms could converge to an optimal policy of the original decision problem. 

In this paper, we derive model-free RL algorithms based on the $\kappa$-greedy formulation of multi-step greedy policies. As mentioned earlier, the main component of this formulation is (approximately) solving a surrogate decision problem with a shaped reward and a smaller discount factor. Our algorithms build on $\kappa$-PI and $\kappa$-VI, and solve the surrogate decision problem with the popular deep RL algorithms: Deep Q-Network (DQN)~\citep{mnih2015human} and Trust Region Policy Optimization (TRPO)~\citep{schulman2015trust}. We call the resulting algorithms $\kappa$-PI-DQN, $\kappa$-VI-DQN, $\kappa$-PI-TRPO, and $\kappa$-VI-TRPO, and empirically evaluate and compare them with DQN, TRPO, and Generalized Advantage Estimation (GAE)~\cite{schulman2015high} on Atari~\cite{bellemare2013arcade} and MuJoCo~\cite{todorov2012mujoco} benchmarks. Our results indicate that for the right range of $\kappa$, our algorithms outperform DQN and TRPO. This suggests that the performance of these two deep RL algorithms can be improved by using them as a solver within the multi-step greedy PI and VI schemes. 

Moreover, our results indicate that the success of our algorithms depends on a number of non-trivial design choices. In particular, we identify the importance of a hyper-parameter that controls the extent to which the surrogate decision problem is solved, and use the theory of multi-step greedy DP to derive a recipe for setting this parameter. We show the advantage of using {\em hard} over {\em soft} updates, verifying the theory in~\citet[Thm.~1]{efroni2018multiple}. By hard and soft update, we refer to fully solving the surrogate MDP in a model-free manner and then evaluating the resulting policy (policy improvement and evaluation steps are separated) vs.~changing the policy at each iteration (policy improvement and evaluation steps are concurrent -- each improvement is followed by an evaluation).

We also establish a connection between our multi-step greedy algorithms and GAE. In particular, we show that our $\kappa$-PI-TRPO algorithm coincides with GAE and we can obtain GAE by minor modifications to $\kappa$-PI-TRPO. Finally, we show the advantage of using our multi-step greedy algorithms over lowering the discount factor in DQN (value-based) and TRPO (policy-based) algorithms. Our results indicate that while lowering the discount factor is detrimental to performance, our multi-step greedy algorithms indeed improve over DQN and TRPO.

\section{Preliminaries}
\label{sec:prelim}

In this paper, we assume that the agent's interaction with the environment is modeled as a discrete time $\gamma$-discounted Markov Decision Process (MDP), defined by $\mathcal{M}_\gamma=(\mathcal{S}, \mathcal{A},P,R,\gamma,\mu)$, where $\mathcal{S}$ and $\mathcal{A}$ are the state and action spaces; $P \equiv P(s'|s,a)$ is the transition kernel; $R \equiv r(s,a)$ is the reward function with the maximum value of $R_{\text{max}}$; $\gamma\in(0,1)$ is the discount factor; and $\mu$ is the initial state distribution. Let $\pi: \mathcal{S}\rightarrow \mathcal{P}(\mathcal{A})$ be a stationary Markovian policy, where $\mathcal{P}(\mathcal{A})$ is a probability distribution on the set $\mathcal{A}$. The value function of a policy $\pi$ at any state $s\in\mathcal{S}$ is defined as $V^\pi(s) \equiv \mathbb{E}[\sum_{t\geq 0}\gamma^tr(s_t,a_t)|s_0=s, \pi]$, where the expectation is over all the randomness in the policy, dynamics, and rewards. Similarly, the action-value function of $\pi$ is defined as $Q^{\pi}(s,a) = \mathbb{E}[\sum_{t\geq 0}\gamma^tr(s_t,a_t)|s_0=s,a_0=a,\pi]$. Since the rewards are bounded by $R_{\text{max}}$, both $V$ and $Q$ functions have the maximum value of $V_{\text{max}}=R_{\text{max}}/(1-\gamma)$. An optimal policy $\pi^*$ is the policy with maximum value at every state. We call the value of $\pi^*$ the optimal value, and define it as $V^*(s) = \max_\pi V^\pi(s),\;\forall s\in\mathcal{S}$. We denote by $Q^*(s,a)$, the state-action value of $\pi^*$, and remind that the following relation holds $V^*(s) = \max_a Q^*(s,a)$, for all $s$. 
The algorithms by which we solve an MDP (obtain an optimal policy) are mainly based on two popular DP algorithms: Policy Iteration (PI) and Value Iteration (VI). While VI relies on iteratively computing the optimal Bellman operator $\mathcal T$ applied to the current value function $V$ (Eq.~\ref{eq: 1 step bellman operator}), PI relies on (iteratively) calculating a 1-step greedy policy $\pi_{\text{1-step}}$ w.r.t.~to the value function of the current policy $V$ (Eq.~\ref{eq: 1 step greedy policy}): $\;$for all $s\in\mathcal S$, we have
\begin{align}
\label{eq: 1 step bellman operator}
(\mathcal{T}V)(s) &= \max_{a\in\mathcal{A}} \; \E[r(s_0,a) + \gamma V(s_1) \mid s_0=s], \\
\label{eq: 1 step greedy policy}
\pi_{\text{1-step}}(s) &\in \arg\max_{a\in\mathcal{A}} \; \E[r(s_0,a) + \gamma  V(s_1) \mid s_0=s]. 
\end{align}
It is known that $\mathcal T$ is a $\gamma$-contraction w.r.t.~the max-norm and its unique fixed-point is $V^*$, and the 1-step greedy policy w.r.t.~$V^*$ is an optimal policy $\pi^*$. In practice, the state space is often large, and thus, we can only approximately compute Eqs.~\ref{eq: 1 step bellman operator} and~\ref{eq: 1 step greedy policy}, which results in approximate PI (API) and VI (AVI) algorithms. These approximation errors then propagate through the iterations of the API and AVI algorithms. However, it has been shown that this (propagated) error can be controlled~\citep{munos2003error,munos2005error,farahmand2010error}, and after $N$ steps, the algorithms approximately converge to a solution $\pi_N$, whose difference with the optimal value is bounded (see e.g.,~\citealp{scherrer2014approximate} for API): 
\begin{equation}
\label{eq: approximate 1 step PI}
\eta(\pi^*) - \eta(\pi_N) \leq C\delta/(1-\gamma)^2 + \gamma^N \Vmax. 
\end{equation}
In Eq.~\ref{eq: approximate 1 step PI}, the scalar $\eta(\pi) = \E_{s\sim\mu}[V^\pi(s)]$ is the expected value function at the initial state,\footnote[1]{Note that the LHS of Eq.~\ref{eq: approximate 1 step PI} is the $\ell_1$-norm of $(V^{\pi^*}-V^{\pi_N})$ w.r.t.~the initial state distribution $\mu$.} $\delta$ represents the per-iteration error, and $C$ upper-bounds the mismatch between the sampling distribution and the distribution according to which the final value function is evaluated ($\mu$ in Eq.~\ref{eq: approximate 1 step PI}), depending heavily on the dynamics. Finally, the second term on the RHS of Eq.~\ref{eq: approximate 1 step PI} is the error due to initial values of policy/value and decays with the number of iterations $N$.

%


\section{$\kappa$-Greedy Policy: $\kappa$-PI and $\kappa$-VI Algorithms} 
\label{sec:dp_multiple_step}

The optimal Bellman operator $\mathcal T$ (Eq.~\ref{eq: 1 step bellman operator}) and 1-step greedy policy $\pi_{\text{1-step}}$ (Eq.~\ref{eq: 1 step greedy policy}) can be generalized to their multi-step versions. The most straightforward form of this generalization is realized by replacing $\mathcal T$ and $\pi_{\text{1-step}}$ with $h$-optimal Bellman operator and $h$-step greedy policy (i.e.,~a lookahead of horizon $h$), respectively. This is done by substituting the 1-step return in Eqs.~\ref{eq: 1 step bellman operator} and~\ref{eq: 1 step greedy policy}, $r(s_0,a)+\gamma V(s_1)$, with the $h$-step return, $\sum_{t=0}^{h-1}r(s_t,a_t) + \gamma^h  V(s_h)$, and computing the maximum over actions $a_0,\ldots,a_{h-1}$, instead of just $a_0$~\citep{bertsekas1996neuro}.~\citet{efroni2018beyond} proposed an alternative form for the multi-step optimal Bellman operator and greedy policy, called $\kappa$-optimal Bellman operator, $\mathcal T_\kappa$, and $\kappa$-greedy policy, $\pi_\kappa$, for $\kappa\in [0,1]$, i.e.,
\begin{align}
\label{eq: kappa optimal bellman}
&(\mathcal T_\kappa V)(s) = \max_\pi \; \E[\sum_{t\geq 0} (\gamma\kappa)^t r_t(\kappa,V) \mid s_0=s,\pi], \\
&\pi_\kappa(s) \in \arg\max_{\pi} \; \E[\sum_{t\geq 0} (\gamma\kappa)^t r_t(\kappa,V) \mid s_0=s, \pi],
\label{eq: kappa greedy policy}
\end{align}
for all $s\in \mathcal S$. In Eqs.~\ref{eq: kappa optimal bellman} and~\ref{eq: kappa greedy policy}, the {\em shaped reward} $r_t(\kappa,V)$ w.r.t.~the value function $V$ is defined as 
\begin{equation}
r_t(\kappa,V) \equiv r_t + \gamma(1-\kappa) V(s_{t+1}).
\label{eq: shaped reward}
\end{equation}
%
%
It can be shown that the $\kappa$-greedy policy w.r.t.~the value function $V$ is the optimal policy w.r.t.~a $\kappa$-weighted geometric average of all future $h$-step returns (from $h=0$ to $\infty$). This can be interpreted as TD($\lambda$)~\citep{sutton2018reinforcement} for policy improvement (see~\citealp[Sec.~6]{efroni2018beyond}). The important difference is that TD($\lambda)$ is used for policy evaluation and not for policy improvement.

It is easy to see that solving Eqs.~\ref{eq: kappa optimal bellman} and~\ref{eq: kappa greedy policy} is equivalent to solving a surrogate $\gamma\kappa$-discounted MDP with the shaped reward $r_t(\kappa,V)$, which we denote by $\mathcal{M}_{\gamma\kappa}(V)$ throughout the paper. The optimal value and policy of the surrogate MDP $\mathcal{M}_{\gamma\kappa}(V)$ are $\mathcal T_\kappa V$ and the $\kappa$-greedy policy $\pi_\kappa$, respectively. Using the notions of $\kappa$-optimal Bellman operator, $\mathcal T_\kappa$, and $\kappa$-greedy policy, $\pi_\kappa$,~\citet{efroni2018beyond} derived $\kappa$-PI and $\kappa$-VI algorithms, whose pseudocodes are shown in Algorithms~\ref{alg:kappaPI} and~\ref{alg:kappaVI}. 
%
%
%
$\kappa$-PI iteratively {\em (i)} evaluates the value $V^{\pi_i}$ of the current policy $\pi_i$, and {\em (ii)} sets the new policy, $\pi_{i+1}$, to the $\kappa$-greedy policy w.r.t.~the value of the current policy $V^{\pi_i}$, by solving Eq.~\ref{eq: kappa greedy policy}. On the other hand, $\kappa$-VI repeatedly applies the $\mathcal T_\kappa$ operator to the current value function $V_i$ (solves Eq.~\ref{eq: kappa optimal bellman}) to obtain the next value function, $V_{i+1}$, and returns the $\kappa$-greedy policy w.r.t.~the final value $V_{N_\kappa}$. Note that for $\kappa=0$, the $\kappa$-optimal Bellman operator and $\kappa$-greedy policy are equivalent to their 1-step counterparts, defined by Eqs.~\ref{eq: 1 step bellman operator} and~\ref{eq: 1 step greedy policy}, 
which indicates that $\kappa$-PI and $\kappa$-VI are generalizations of the seminal PI and VI algorithms.

\begin{minipage}{0.48\textwidth}
\begin{algorithm}[H]
\begin{small}
	\caption{$\kappa$-Policy Iteration}
	\label{alg:kappaPI}
	\begin{algorithmic}[1]
		\STATE {\bfseries Initialize:} $\kappa \in [0,1]$, $\pi_0$, $N_\kappa$
		\FOR{$i = 0,1,\ldots,N_\kappa-1$}
		\STATE $V^{\pi_i} = \E [\sum_{t\geq 0} \gamma^t r_t \mid \pi_i]$
		\STATE  $\pi_{i+1} \gets \arg\max\limits_{\pi}\;\E[\sum_{t\geq 0}(\kappa\gamma)^t r_{t}(\kappa,V^{\pi_i}) \mid \pi]$
		\ENDFOR
		\STATE {\bfseries Return $\pi_{N_\kappa}$}
	\end{algorithmic}
\end{small}	
\end{algorithm}
\end{minipage}
\hspace{0.3cm}
\begin{minipage}{0.48\textwidth}
\begin{algorithm}[H]
\begin{small}
	\caption{$\kappa$-Value Iteration}
	\label{alg:kappaVI}
	\begin{algorithmic}[1]
		\STATE {\bfseries Initialize:} $\kappa \in [0,1]$, $V_0$ , $N_\kappa$
		\FOR{$i = 0,1,\ldots,N_\kappa-1$}
		\STATE $V_{i+1} = \max_{\pi} \E[\sum_{t\geq 0} (\gamma\kappa)^t r_t(\kappa,V_{i}) \mid \pi]$
		\ENDFOR
		\STATE $\pi_{N_\kappa} \gets \arg\max\limits_{\pi}\;\E[\sum_{t\geq 0}(\kappa\gamma)^t r_{t}(\kappa,V_{N_\kappa}) \mid \pi]$
		\STATE {\bfseries Return $\pi_{N_\kappa}$}
	\end{algorithmic}
\end{small}
\end{algorithm}
\end{minipage}

It has been shown that both PI and VI converge to the optimal value with an exponential rate that depends on the discount factor $\gamma$, i.e.,~$\norm{V^*- V^{\pi_N}}_\infty \leq O(\gamma^N)$ (see e.g.,~\citealp{bertsekas1996neuro}). Analogously,~\citet{efroni2018beyond} showed that $\kappa$-PI and $\kappa$-VI converge with a faster exponential rate $\xi_\kappa = \frac{\gamma(1-\kappa)}{1-\gamma\kappa}\leq \gamma$, i.e.,~$\norm{V^*- V^{\pi_{N_\kappa}}}_\infty \leq O(\xi_\kappa^{N_\kappa})$, with the cost that each iteration of these algorithms is computationally more expensive than that of PI and VI. Finally, we state the following property of $\kappa$-PI and $\kappa$-greedy policies that we use in our $\kappa$-PI and $\kappa$-VI based RL algorithms described in Section~\ref{sec:RL-kPI-kVI}:


{\em Asymptotic performance depends on $\kappa$.}~\citet[Thm.~5]{efroni2018multiple} proved the following bound on the performance of $\kappa$-PI that is similar to the one in Eq.~\ref{eq: approximate 1 step PI} for API: 
%
%
%
\begin{equation}
\eta(\pi^*) - \eta(\pi_{N_\kappa}) \leq \underbrace{C_\kappa\delta_\kappa/(1-\gamma)^2}_{\text{Asymptotic Term}} + \underbrace{\xi_\kappa^{N_\kappa} \Vmax}_{\text{Decaying Term}}, 
\label{eq: approximate kappa PI}
\end{equation}
where $\delta_\kappa$ and $C_\kappa$ are quantities similar to $\delta$ and $C$ in Eq.~\ref{eq: approximate 1 step PI}. Note that while the second term on the RHS of Eq.~\ref{eq: approximate kappa PI} decays with $N_\kappa$, the first one is independent of $N_\kappa$. 
    

\section{$\kappa$-PI and $\kappa$-VI based RL Algorithms}
\label{sec:RL-kPI-kVI}

As described in Section~\ref{sec:dp_multiple_step}, implementing $\kappa$-PI and $\kappa$-VI requires iteratively solving a $\gamma\kappa$-discounted surrogate MDP with a shaped reward. If a model of the problem is given, the surrogate MDP can be solved using a DP algorithm (see \citealp[Sec.~7]{efroni2018beyond}). When a model is not available, we should approximately solve the surrogate MDP using a model-free RL algorithm. In this paper, we focus on the latter case and propose RL algorithms inspired by $\kappa$-PI and $\kappa$-VI. In our algorithms, we use model-free RL algorithms DQN~\citep{mnih2015human} and TRPO~\citep{schulman2015trust} (for discrete and continuous action problems, respectively) as subroutines for estimating a $\kappa$-greedy policy (Line~4 in Alg.~\ref{alg:kappaPI}, $\kappa$-PI, and Line~5 in Alg.~\ref{alg:kappaVI}, $\kappa$-VI) and an optimal value of the surrogate MDP (Line~3 in Alg.~\ref{alg:kappaVI}, $\kappa$-VI). 
We refer to the resulting algorithms as $\kappa$-PI-DQN, $\kappa$-VI-DQN, $\kappa$-PI-TRPO, and $\kappa$-VI-TRPO. 

In order to have an efficient implementation of our $\kappa$-PI and $\kappa$-VI based algorithms, the main question to answer is how a fixed number of samples $T$ should be allocated to different parts of the $\kappa$-PI and $\kappa$-VI algorithms? More precisely, how shall we set $N_\kappa\in \mathbb{N}$, i.e.,~the total number of iterations of our algorithms, and determine the number of samples to solve the surrogate MDP at each iteration? To answer these questions, we devise a heuristic approach based on the theory of $\kappa$-PI and $\kappa$-VI algorithms, and in particular Eq.~\ref{eq: approximate kappa PI}. Since $N_\kappa$ only appears explicitly in the second term on the RHS of Eq.~\ref{eq: approximate kappa PI}, an appropriate choice of $N_\kappa$ is such that $C_\kappa\delta_\kappa/(1-\gamma)^2 \simeq \xi_\kappa^{N_\kappa} \Vmax$. Note that setting $N_\kappa$ to a higher value would {\em not} significantly improve the performance, because the asymptotic term in Eq.~\ref{eq: approximate kappa PI} is independent of $N_\kappa$. In practice, since $\delta_\kappa$ and $C_\kappa$ are unknown, we set $N_\kappa$ to satisfy the following equality:
\begin{align}
\xi_\kappa^{N_\kappa} = C_{\text{FA}}, 
\label{eq: final accuracy}
\end{align}
where $C_{\text{FA}}$ is a hyper-parameter that depends on the \emph{final-accuracy} we aim for. For example, if our goal is to obtain $90\%$ accuracy, we would set $C_{\text{FA}}=0.1$, which results in $N_{\kappa=0.99} \simeq 4$ and $N_{\kappa=0.5}\simeq 115$, for $\gamma=0.99$. Our experimental results in Section~\ref{sec:experiments} suggest that this approach leads to a reasonable choice for the total number of iterations $N_\kappa$. It is important to note the following facts: {\bf 1)} as we increase $\kappa$, we expect less iterations are needed for $\kappa$-PI and $\kappa$-VI to converge to a good policy, and {\bf 2)} the {\em effective horizon}\footnote[2]{The effective horizon of a $\gamma\kappa$-discounted MDP is $1/(1-\gamma\kappa)$.} of the surrogate MDP that $\kappa$-PI and $\kappa$-VI solve at each iteration increases with $\kappa$.
%

Lastly, we need to determine the number of samples for each iteration of our $\kappa$-PI and $\kappa$-VI based algorithms. We allocate equal number of samples per iteration, denoted by $T_\kappa$. Since the total number of samples, $T$, is known beforehand, we set the number of samples per iteration to 

\begin{equation}
\label{eq: number of samples per iteration}
T_\kappa=T/N_\kappa.
\end{equation}

In the rest of the paper, we first derive our DQN-based and TRPO-based algorithms in Sections~\ref{subsec:DQN-Algos} and~\ref{subsec:TRPO-Algos}. 
It is important to note that for $\kappa=1$, our algorithms are reduced to DQN and TRPO. We then conduct a set of experiments with our algorithms in Sections~\ref{subsec:DQN-Experiments} and~\ref{subsec:TRPO-Experiments} in which we carefully study the effect of $\kappa$ and $N_\kappa$ (or equivalently the hyper-parameter $C_{\text{FA}}$, defined by Eq.~\ref{eq: final accuracy}) on their performance.




\subsection{$\kappa$-PI-DQN and $\kappa$-VI-DQN Algorithms} 
\label{subsec:DQN-Algos}

\begin{algorithm}[t]
\begin{small}
    \caption{$\kappa$-PI-DQN}
    \label{alg:kappaPI-DQN}
    \begin{algorithmic}[1]
        \STATE {\bfseries Initialize} replay buffer $\mathcal{D}$; $Q$-networks $Q_{\theta}$, $Q_{\phi}$; target networks $Q'_{\theta}$, $Q'_{\phi}$; \\
        \FOR{ $i = 0, \ldots, N_\kappa-1$}
            \STATE {\color{gray}\# Policy Improvement}
            \FOR{$t =1,\ldots,T_\kappa$}
                \STATE Act by an $\epsilon$-greedy policy w.r.t.~$Q_{\theta}(s_t, a)$, observe $r_t, s_{t+1}$, and store $(s_t,a_t,r_t,s_{t+1})$ in $\mathcal{D}$;
                \STATE Sample a batch $\{(s_j, a_j, r_j, s_{j+1})\}_{j=1}^N$ from $\mathcal{D}$;
                \vspace{0.4mm} 
                \STATE Update $\theta$ using DQN with 
                \vspace{0.4mm} 
                \STATE $\;\;\{(s_j, a_j, r_j(\kappa,V_\phi), s_{j+1})\}_{j=1}^N$, $\;$ where
                \vspace{0.4mm} 
                \STATE $\;\;V_\phi(s_{j+1})=Q_\phi(s_{j+1},\pi_{i-1}(s_{j+1}))\;$ and 
                \vspace{0.4mm}
                \STATE $\;\;\pi_{i-1}(\cdot)\in\arg\max_a Q'_\theta(\cdot,a)$; 
                \vspace{0.4mm}
                \STATE Copy $\theta$ to $\theta'$ occasionally $\quad (\theta'\gets\theta)$;
            \ENDFOR
            \STATE {\color{gray}\# Policy Evaluation of  $\pi_i(s)\in \arg\max_a Q'_{\theta}(s,a)$} \label{line:temp2}
            \FOR{$t=1,\ldots,T_\kappa$}
                \STATE Sample a batch $\{(s_j, a_j, r_j, s_{j+1})\}_{j=1}^N$ from $\mathcal{D}$;
                \STATE Update $\phi$ using this data and off-policy TD($0$) to estimate the $Q$-function of the current policy $\pi_i$;
                \STATE  Copy $\phi$ to $\phi'$ occasionally $\quad (\phi'\gets\phi)$;
            \ENDFOR
            \ENDFOR
    \end{algorithmic}
\end{small}    
\end{algorithm}

\begin{algorithm}[t]
\begin{small}
    \caption{$\kappa$-PI-TRPO}
    \label{alg:kappaPI-TRPO}
    \begin{algorithmic}[1]
        \STATE {\bfseries Initialize} $V$-networks $V_{\theta}$ and $V_{\phi}$; $\;$ policy network $\pi_\psi$;
        \FOR{ $i = 0, \ldots, N_\kappa-1$}
            \FOR{$t =1,\ldots,T_\kappa$}
                \STATE Simulate the current policy $\pi_\psi$ for $M$ steps and calculate the following two returns for all steps $j$: \\
                $R_j(\kappa,V_\phi) = \sum_{t=j}^M (\gamma \kappa)^{t-j} r_t(\kappa,V_\phi)\;$ and \\
                $\rho_j = \sum_{t=j}^M \gamma^{t-j} r_t$;
                \STATE Update $\theta$ by minimizing the batch loss function: \\ 
                $\mathcal{L}_{V_\theta}$ = $\frac{1}{N} \sum_{j=1}^{N}(V_\theta(s_j) - R_{j}(\kappa,V_\phi))^2$;
                \STATE {\color{gray}\# Policy Improvement}
                \STATE Update $\psi$ using TRPO and the batch \\ 
                $\{(R_j(\kappa,V_\phi),V_\theta(s_j))\}_{j=1}^N$;
            \ENDFOR
            \STATE {\color{gray}\# Policy Evaluation}
            \STATE Update $\phi$ by minimizing the batch loss function: \\ 
            $\mathcal{L}_{V_{\phi}}=\frac{1}{N} \sum_{j=1}^N(V_\phi(s_j) - \rho_j)^2$;
        \ENDFOR
    \end{algorithmic}
\end{small}
\end{algorithm}

Algorithm~\ref{alg:kappaPI-DQN} presents the pseudo-code of $\kappa$-PI-DQN. Due to space constraints, we report the detailed pseudo-code in Appendix~\ref{supp: pseudo code k-VI k-PI DQN} (Alg.~\ref{alg:kappaPI-DQN-Full}). We use four neural networks in this algorithm, two to represent the $Q$-function of the original MDP (with discount factor $\gamma$ and reward $r$), $Q_\phi$ ($Q$-network) and $Q'_\phi$ (target network), and two for the $Q$-function of the surrogate MDP, $Q_\theta$ ($Q$-network) and $Q'_\theta$ (target network). In the {\em policy improvement step}, we use DQN to solve the $\gamma\kappa$-discounted surrogate MDP with the shaped reward $r_j(\kappa,V_\phi)=r_j+\gamma(1-\kappa)V_\phi(s_{j+1})$, i.e.,~$\mathcal{M}_{\gamma\kappa}(V_\phi)$, where $V_\phi\simeq V^{\pi_{i-1}}$ and is computed as $V_\phi(s)=Q_\phi(s,\argmax_a Q'_\theta(s,a))$. The output of DQN is (approximately) the optimal $Q$-function of $\mathcal{M}_{\gamma\kappa}(V_\phi)$, and thus, the new policy $\pi_i$, which is the (approximate) $\kappa$-greedy policy w.r.t.~$V_\phi$ is equal to $\pi_i(\cdot) = \arg\max_a Q'_\theta(\cdot,a)$. In the {\em policy evaluation step}, we use off-policy TD($0$) to evaluate the $Q$-function of the current policy $\pi_i$, i.e.,~$Q_\phi \simeq Q^{\pi_i}$. Although what is needed is an estimate of the value function of the current policy, $V_\phi\simeq V^{\pi_i}$, we chose to evaluate its $Q$-function, because the available data (the transitions stored in the replay buffer) is off-policy, and unlike the value function, the $Q$-function of a fixed policy can be easily evaluated with this type of data using off-policy TD($0$).

\begin{remark}
In the policy improvement phase, $\pi_{i-1}$ is computed as the $\arg\max$ of $Q'_\theta(\cdot,a)$ (Line~10), a quantity that is not constant and is (slowly) changing during this phase. Addressing this issue requires using an additional target network that is set to $Q_\theta$ only at the end of each improvement step and its $\argmax$ is used to compute $\pi_{i-1}$ throughout the improvement step of the next iteration. We tested using this additional network in our experiments, but it did not improve the performance, and thus, we decided to report the algorithm without it.    
\end{remark}



\begin{figure*}
    \centering
    \begin{subfigure}[t]{\textwidth}
        \centering
        \includegraphics[scale=0.22]{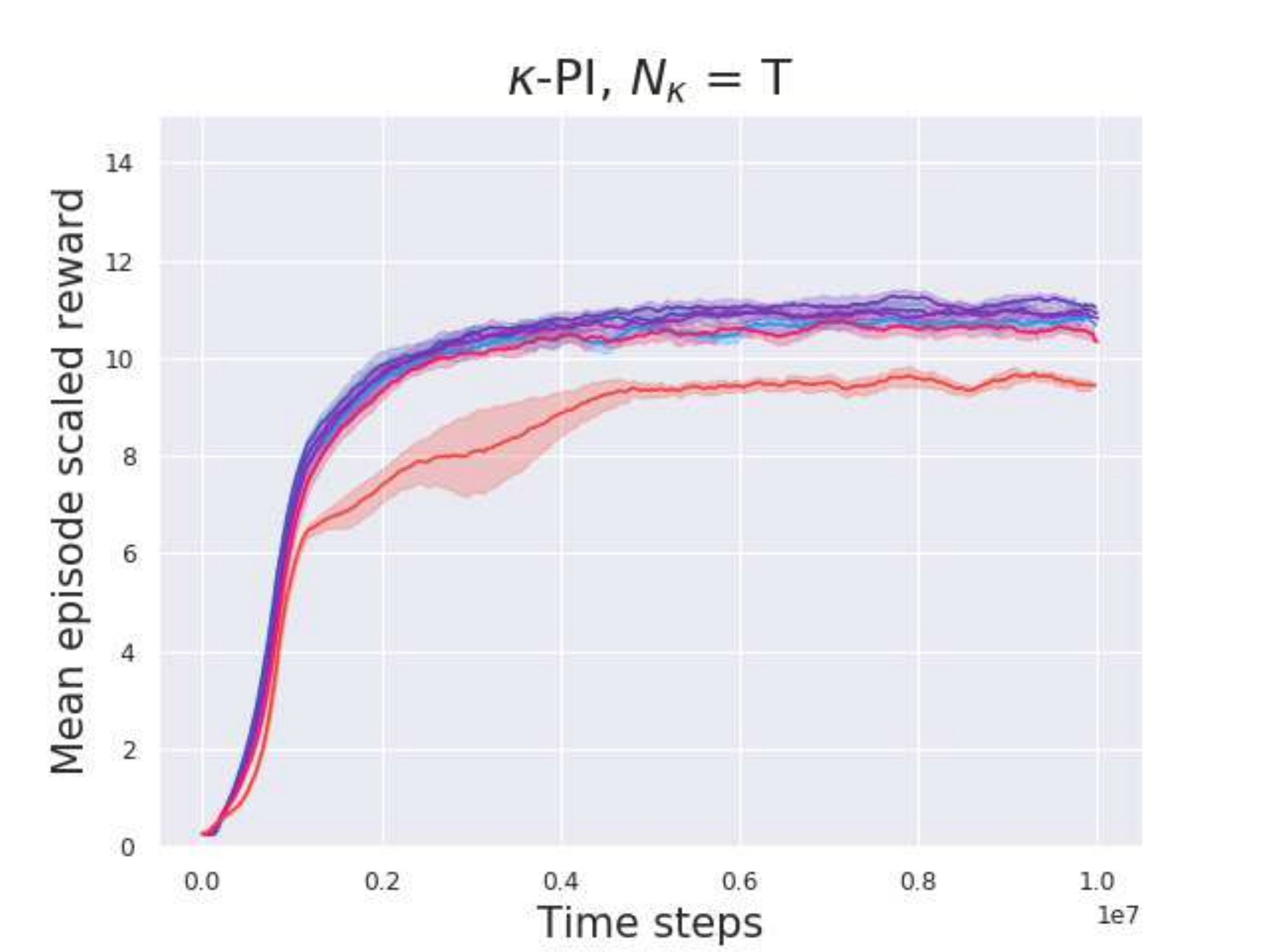}
        \hspace{-0.75cm} \includegraphics[scale=0.22]{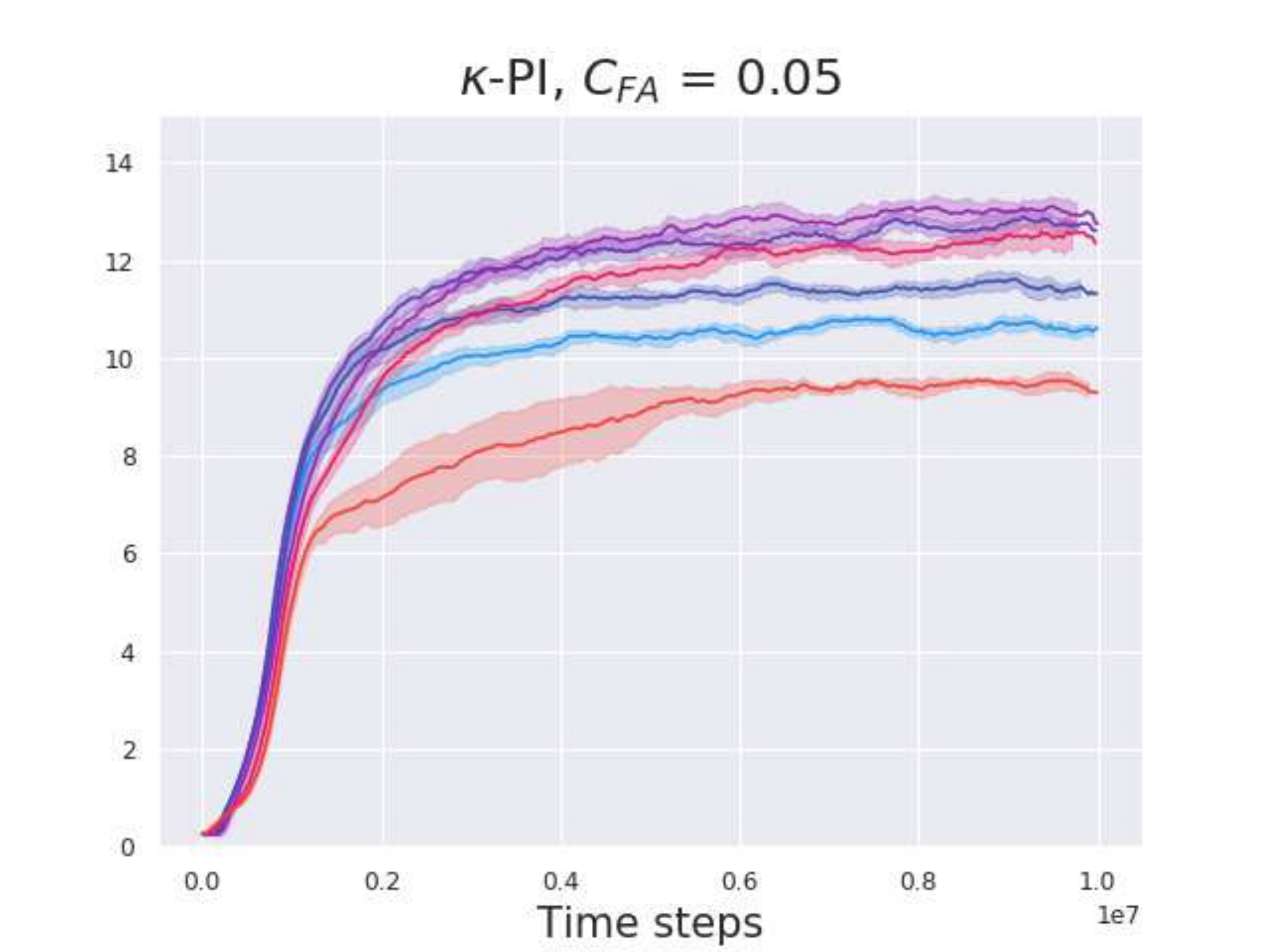}
        \hspace{-0.75cm} \includegraphics[scale=0.22]{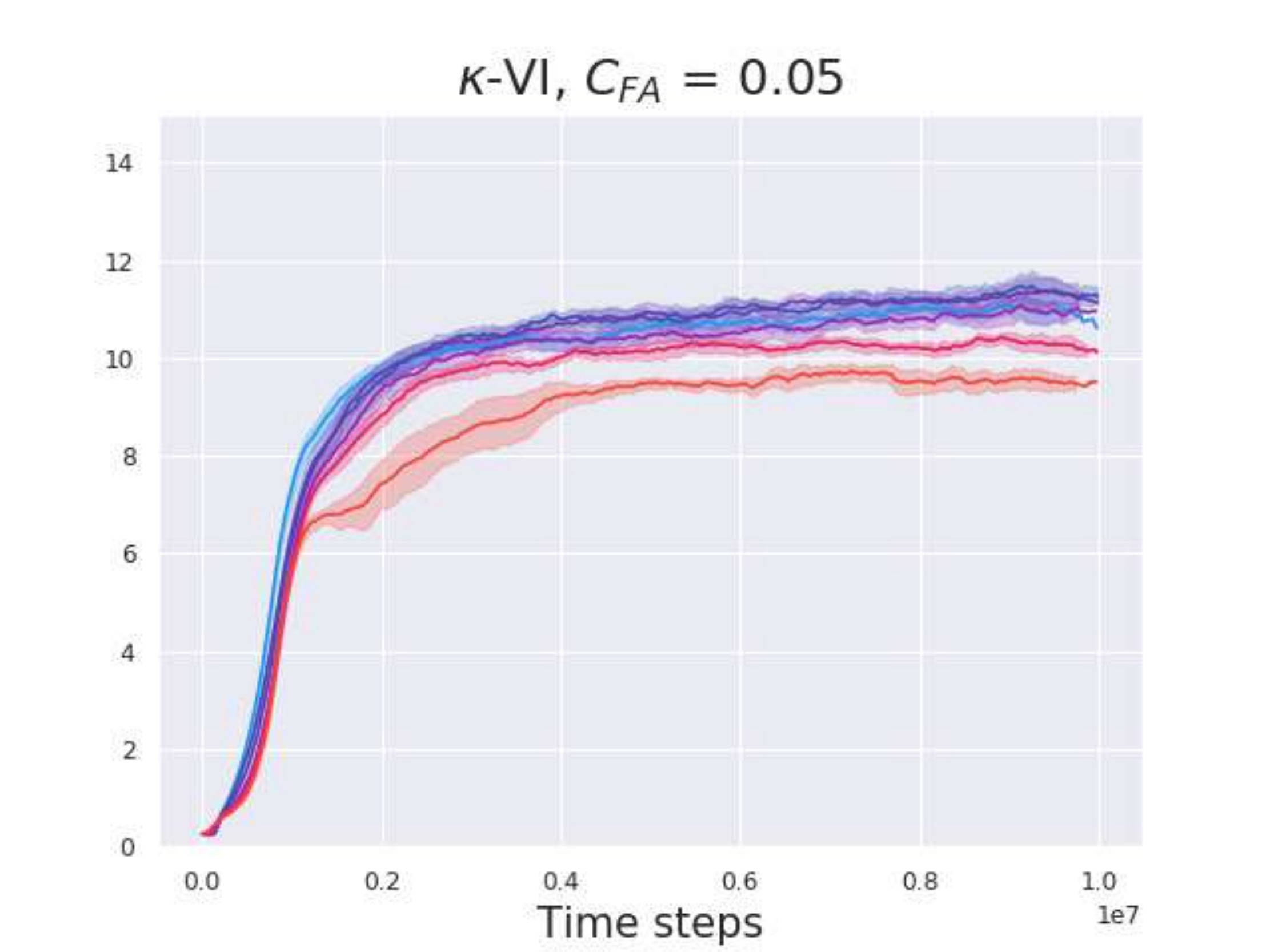} \\
        \includegraphics[scale=0.3]{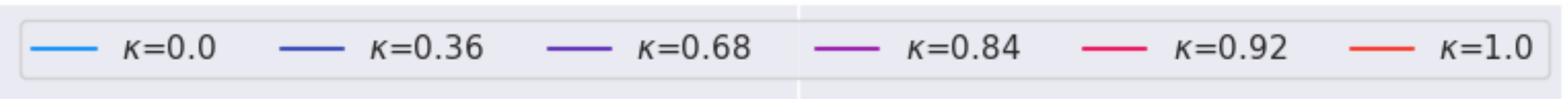}
        \caption*{\textit{Breakout}}
    \end{subfigure} \\
    \begin{subfigure}[t]{\textwidth}
        \centering
        \includegraphics[scale=0.20]{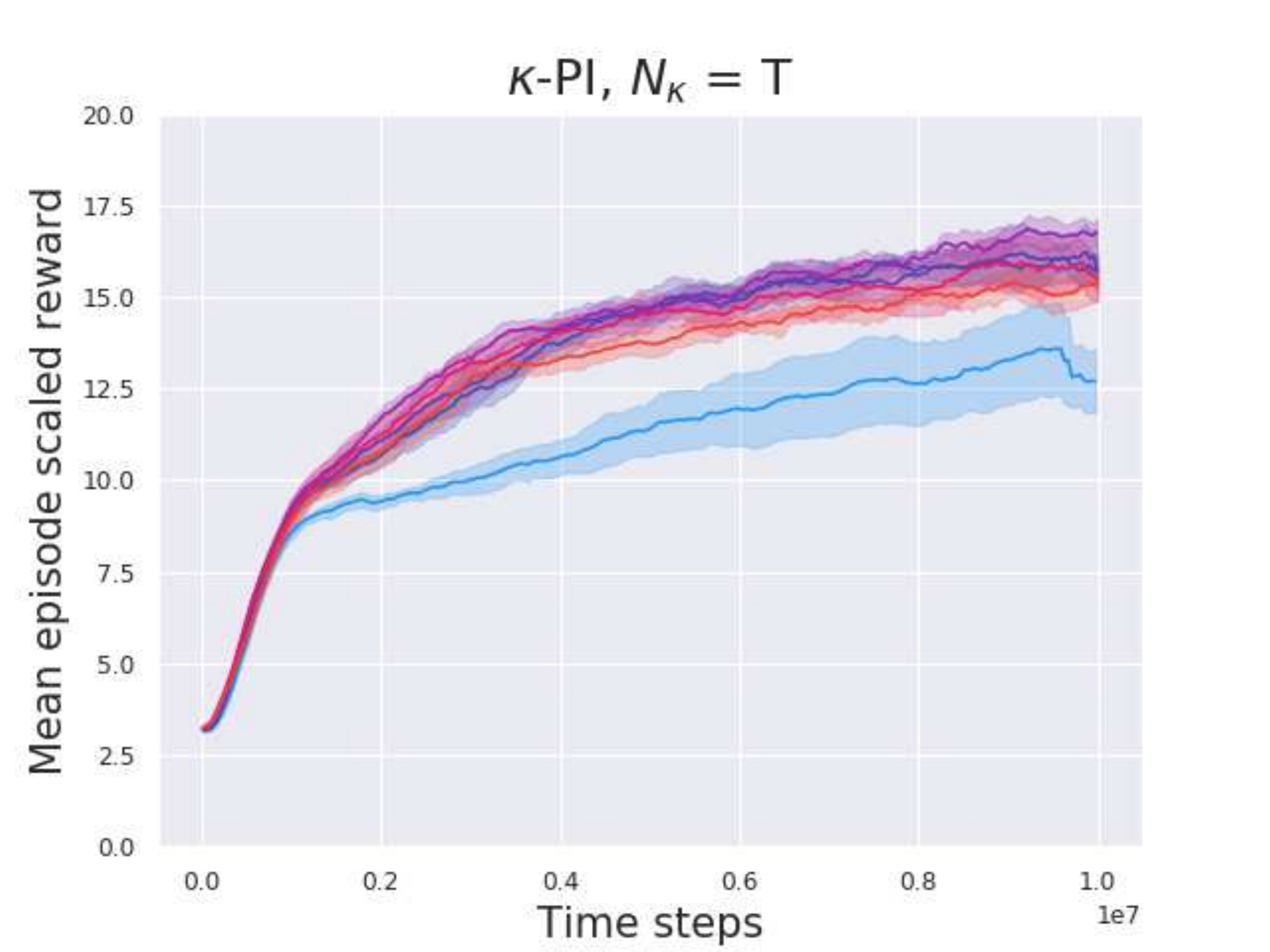}
        \includegraphics[scale=0.20]{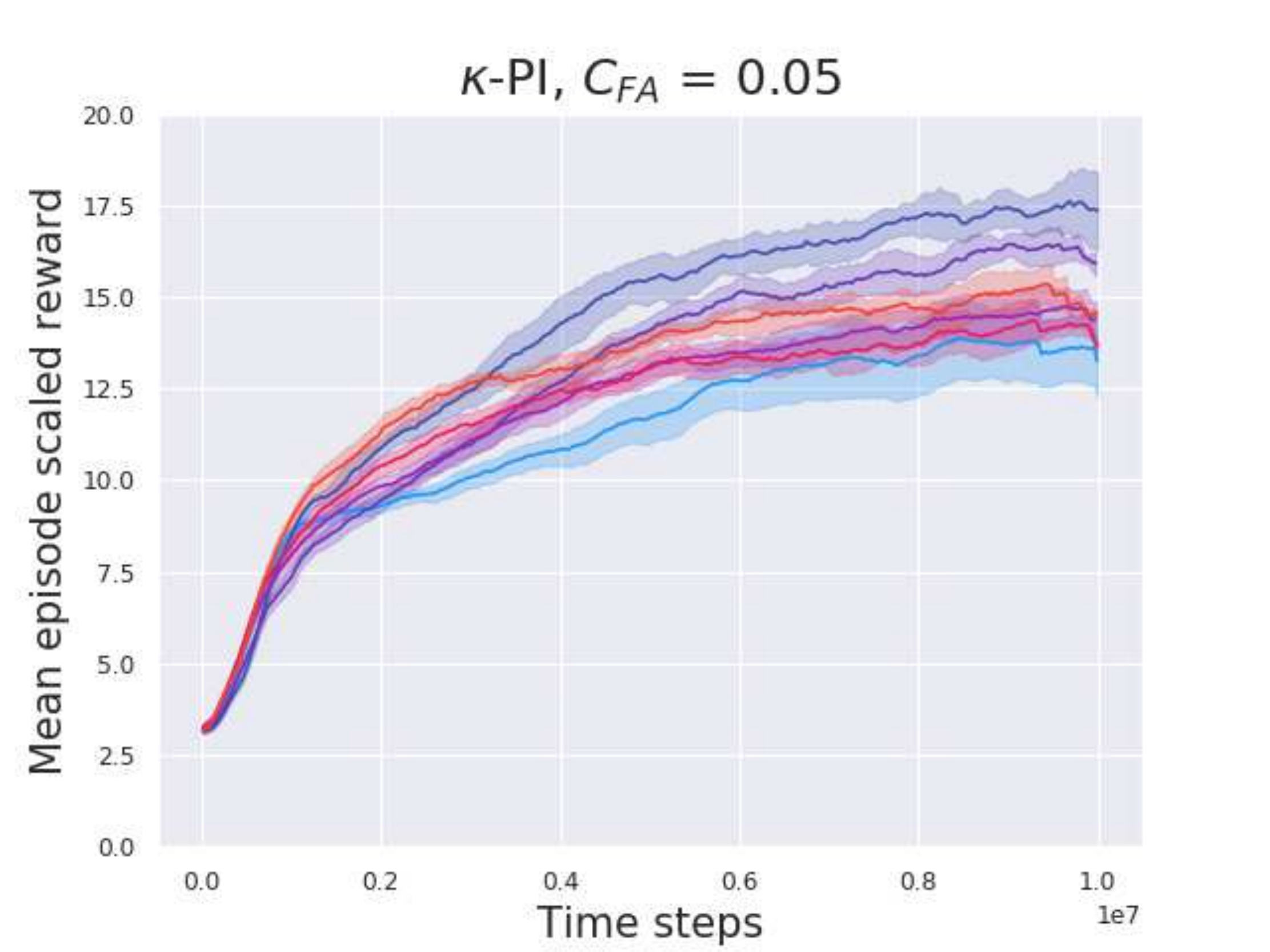}
        \includegraphics[scale=0.20]{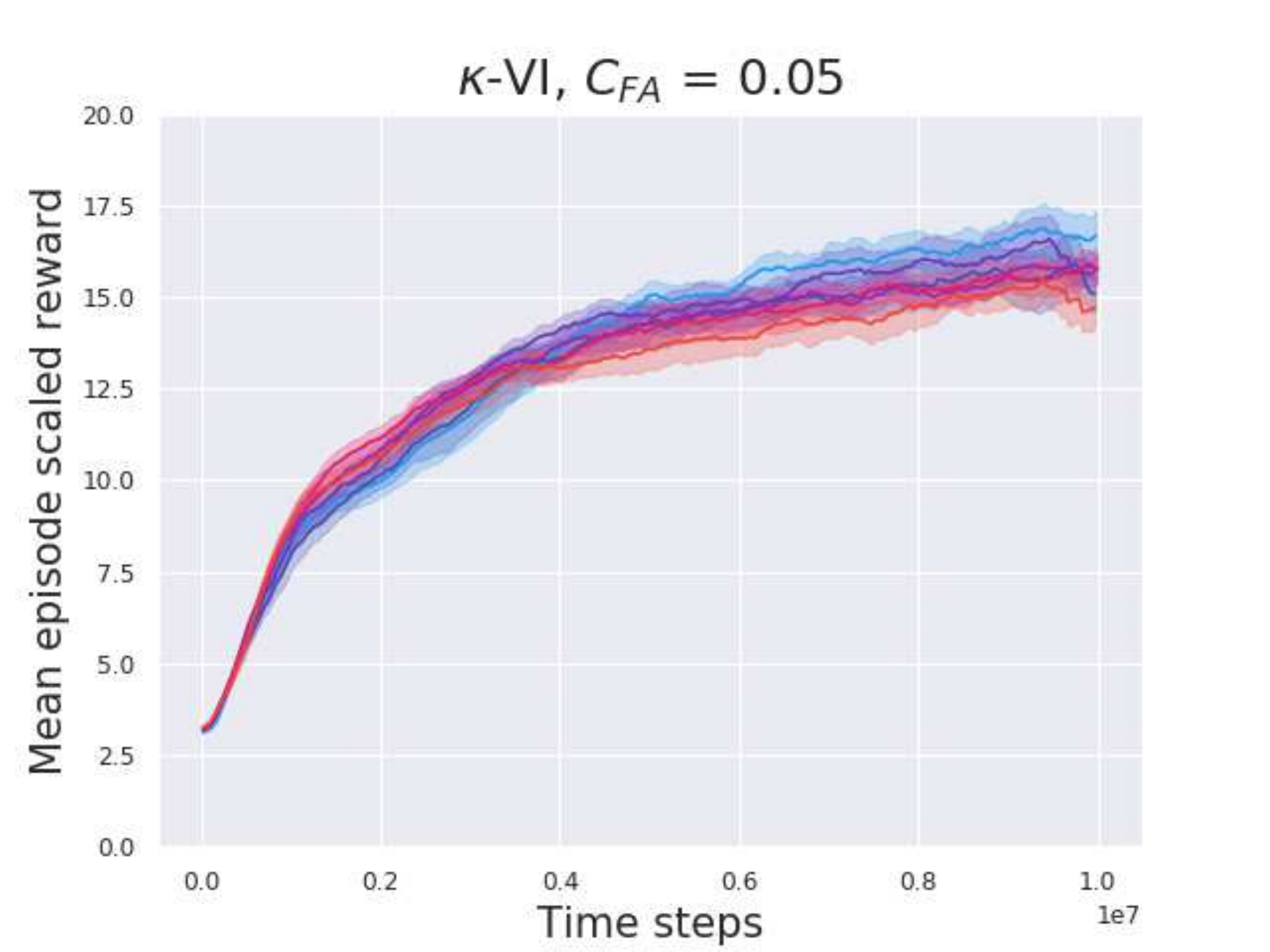}
        \includegraphics[scale=0.3]{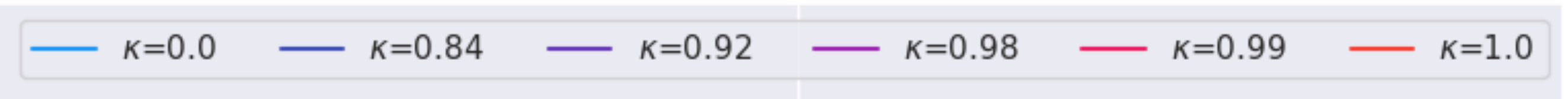}
        \caption*{\textit{SpaceInvaders}}
    \end{subfigure}
    \caption{Training performance of the `naive' baseline $N_\kappa=T$ and $\kappa$-PI-DQN, $\kappa$-VI-DQN for $C_{FA}=0.05$ on Breakout (top) and SpaceInvaders (bottom). See Appendix~\ref{subsec: appendix CFA} for performance w.r.t. different $C_{FA}$ values.}
    \label{fig:Breakout-kPI-kVI}
\end{figure*}

We report the pseudo-code of $\kappa$-VI-DQN in Appendix~\ref{supp: pseudo code k-VI k-PI DQN} (Alg.~\ref{alg:kappaVI-DQN-Full}). Note that $\kappa$-VI simply repeats $V \gets \mathcal{T}_\kappa V$ and computes $\mathcal{T}_\kappa V$, which is the optimal value of the surrogate MDP $\mathcal{M}_{\gamma\kappa}(V)$. In $\kappa$-VI-DQN, we repeatedly solve $\mathcal{M}_{\gamma\kappa}(V)$ by DQN and use its (approximately) optimal $Q$-function to shape the reward of the next iteration. 
The algorithm uses three neural networks, two to solve the surrogate MDP by DQN, $Q_\theta$ ($Q$-network) and $Q'_\theta$ (target network), and one to store its optimal $Q$-function to use it for the shaped reward in the next iteration, $Q_\phi$. Let $Q^*_{\gamma\kappa,V}$ and $V^*_{\gamma\kappa,V}$ be the optimal $Q$ and $V$ functions of $\mathcal{M}_{\gamma\kappa}(V)$. Then, we have $\max_{a}Q^*_{\gamma\kappa,V}(s,a) = V^*_{\gamma\kappa,V}(s) = (\mathcal T_\kappa V)(s)$, where the first equality is by definition (Sec.~\ref{sec:prelim}) and the second one holds since $\mathcal T_\kappa V$ is the optimal value of $\mathcal{M}_{\gamma\kappa}(V)$ (Sec.~\ref{sec:dp_multiple_step}). Therefore, in $\kappa$-VI-DQN, we shape the reward at each iteration by $\max_a Q_\phi(s,a)$, where $Q_\phi$ is the output of the DQN from the previous iteration, i.e.,~$\max_a Q_\phi(s,a) \simeq \mathcal T_\kappa V_{i-1}$ .


\subsection{$\kappa$-PI-TRPO and $\kappa$-VI-TRPO Algorithms}
\label{subsec:TRPO-Algos}

Algorithm~\ref{alg:kappaPI-TRPO} presents the pseudo-code of $\kappa$-PI-TRPO (a detailed pseudo-code is reported in Appendix~\ref{supp: pseudo code k-VI k-PI TRPO}, Alg.~\ref{alg:kappaPI-TRPO-Full}). Recall that TRPO iteratively updates the current policy using its return and an estimate of its value function. At each iteration $i$ of $\kappa$-PI-TRPO: {\bf 1)} we use the estimate of the value of the current policy $V_\phi \simeq V^{\pi_{i-1}}$ to calculate the return $R(\kappa,V_\phi)$ and the estimate of the value function $V_\theta$ of the surrogate MDP $\mathcal{M}_{\gamma\kappa}(V_\phi)$, {\bf 2)} we use $R(\kappa,V_\phi)$ and $V_\theta$ to compute the new policy $\pi_i$ (TRPO style), and {\bf 3)} we estimate the value of the new policy $V_\phi \simeq V^{\pi_i}$ in the original MDP (with discount factor $\gamma$ and reward $r$). The algorithm uses three neural networks, one for the value function of the original MDP, $V_\phi$, one for the value function of the surrogate MDP, $V_\theta$, and one for the policy, $\pi_\psi$.

We report the pseudo-code of $\kappa$-VI-TRPO in Appendix~\ref{supp: pseudo code k-VI k-PI TRPO}, Alg.~\ref{alg:kappaVI-TRPO-Full}. As previously noted, $\kappa$-VI iteratively solves the surrogate MDP and uses its optimal value $\mathcal T_\kappa V_{i-1}$ to shape the reward of the surrogate MDP in the next iteration. In $\kappa$-VI-TRPO, we solve the surrogate MDP $\mathcal{M}_{\gamma\kappa}(V_{i-1}=V_\phi)$ with TRPO until its policy $\pi_\psi$ converges to the optimal policy of $\mathcal{M}_{\gamma\kappa}(V_{i-1}= V_\phi)$ and its value $V_\theta$ converges to $\mathcal T_\kappa V_{i-1} = \mathcal T_\kappa V_\phi$. We then replace $V_i$ with $V_\theta$ ($V_i=V_\phi\leftarrow V_\theta$) and repeat this process. 


\section{Experimental Results} 
\label{sec:experiments}

\begin{figure*}
    \centering
    \begin{subfigure}[t]{\textwidth}
        \centering
        \includegraphics[scale=0.22]{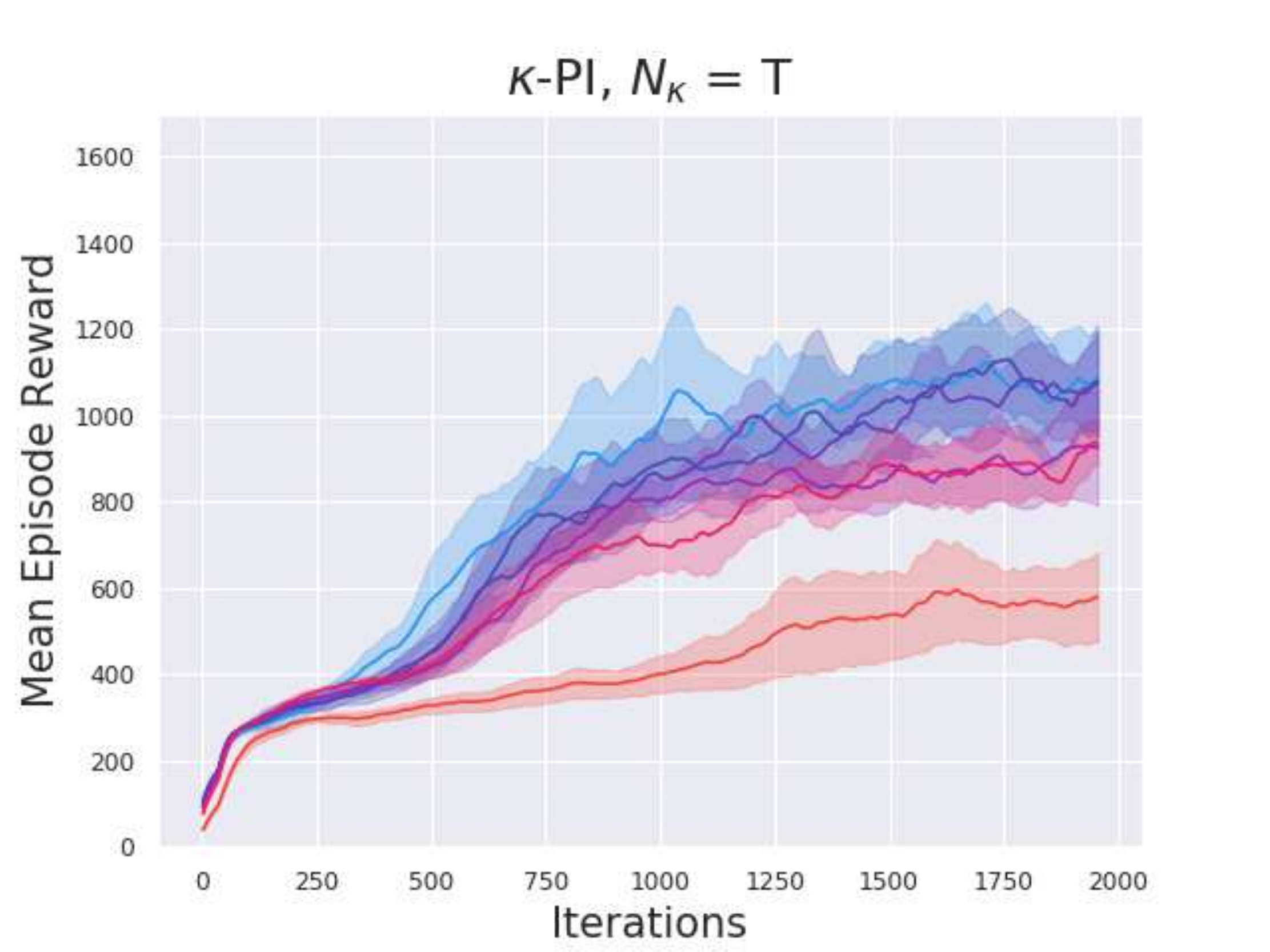}
        \hspace{-0.75cm} \includegraphics[scale=0.22]{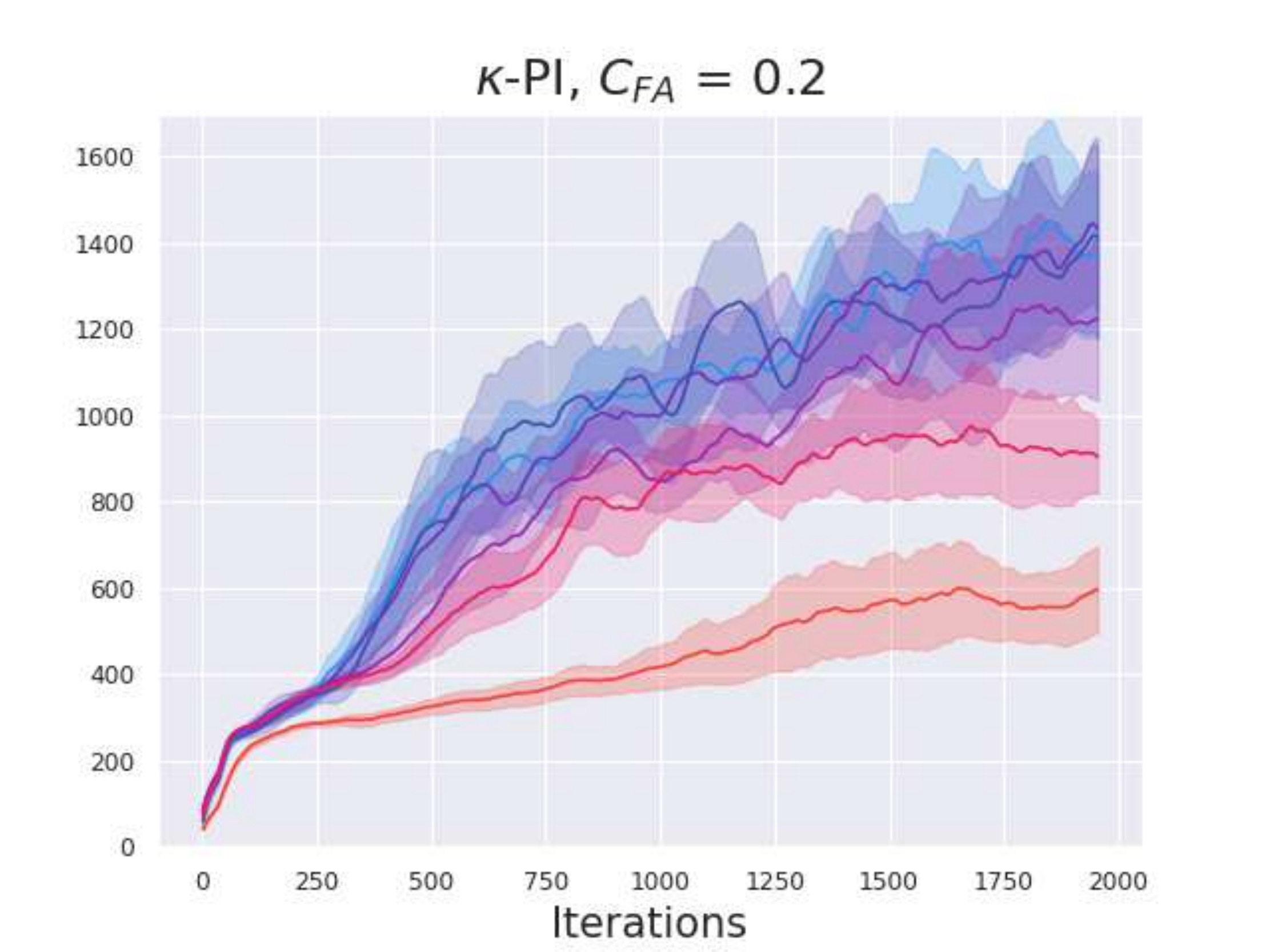}
        \hspace{-0.75cm} \includegraphics[scale=0.22]{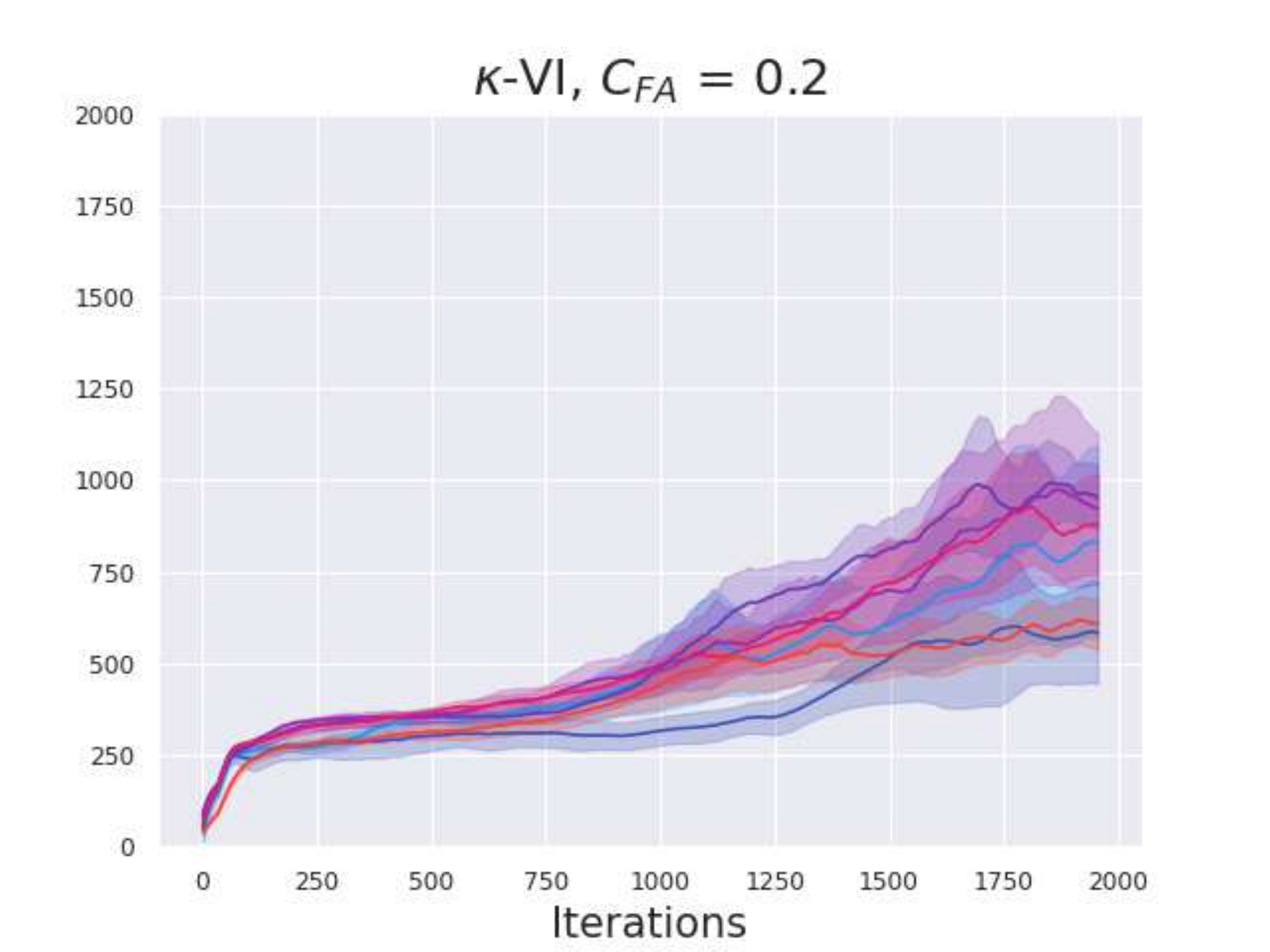} \\
        \includegraphics[scale=0.30]{Images/TRPO/legend_horizontal.pdf}
        \caption*{\textit{Walker-v2}}
    \end{subfigure} \\
    \begin{subfigure}[t]{\textwidth}
        \centering
        \includegraphics[scale=0.22]{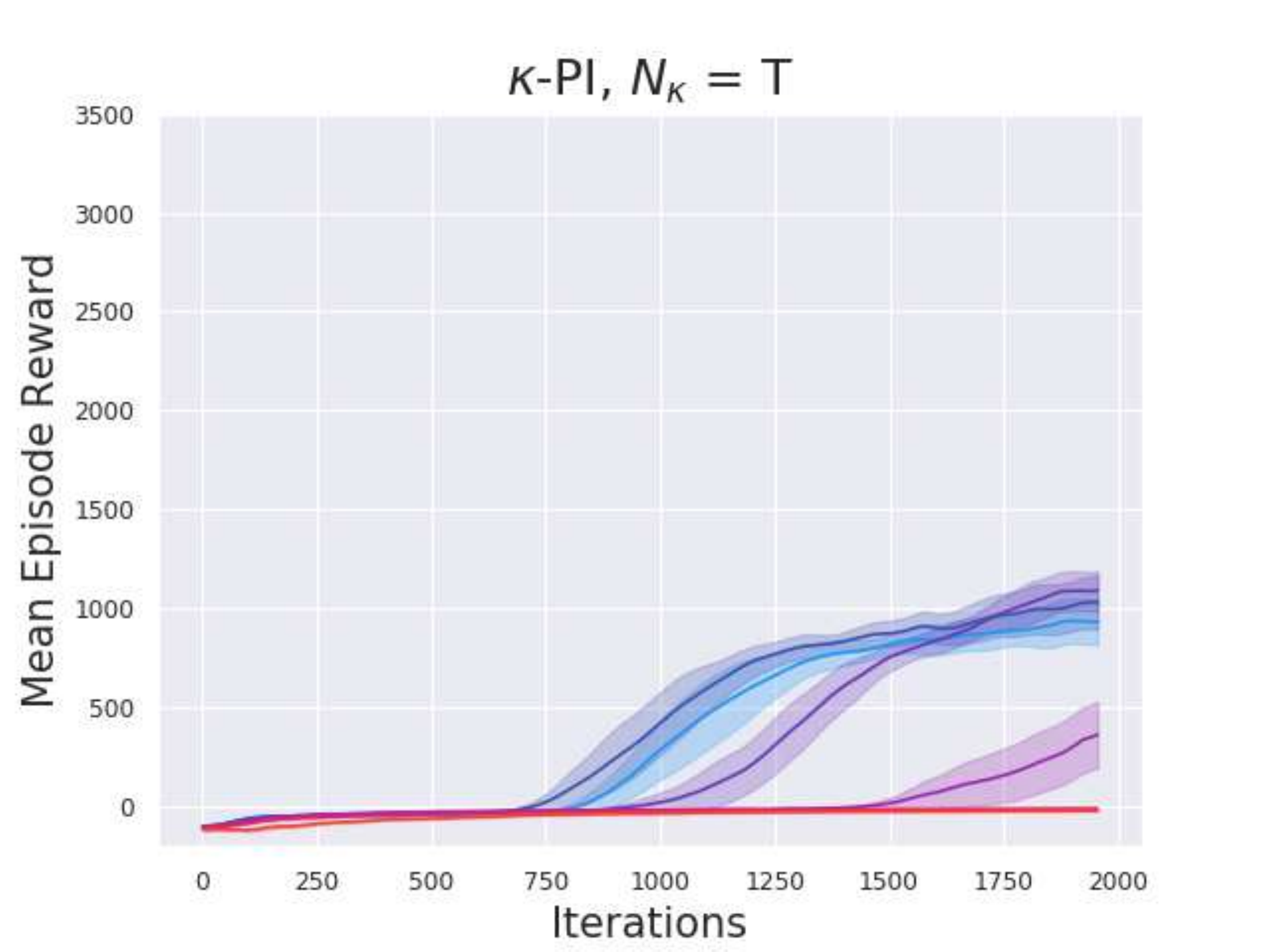}
        \hspace{-0.75cm} \includegraphics[scale=0.22]{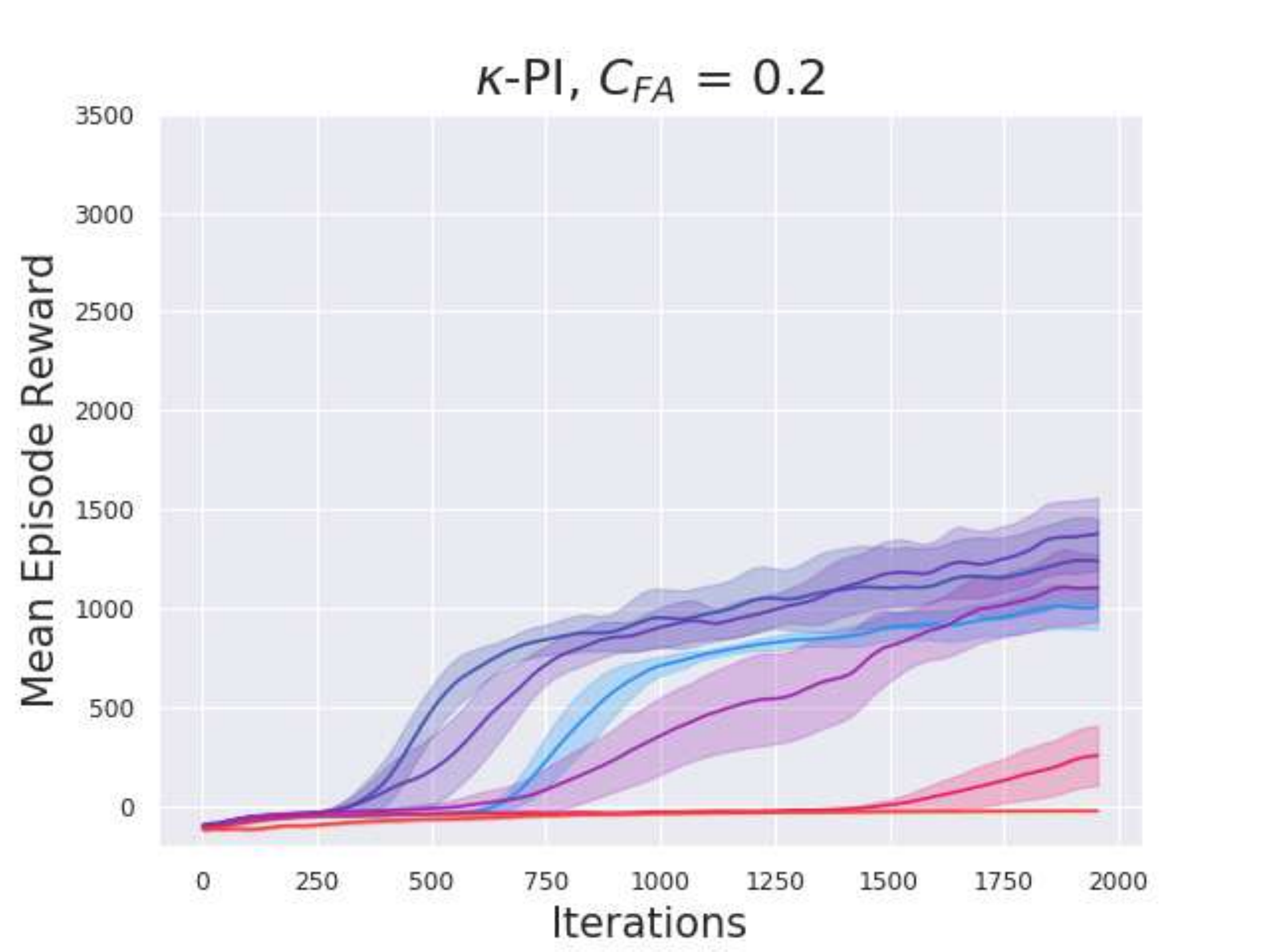}
        \hspace{-0.75cm} \includegraphics[scale=0.22]{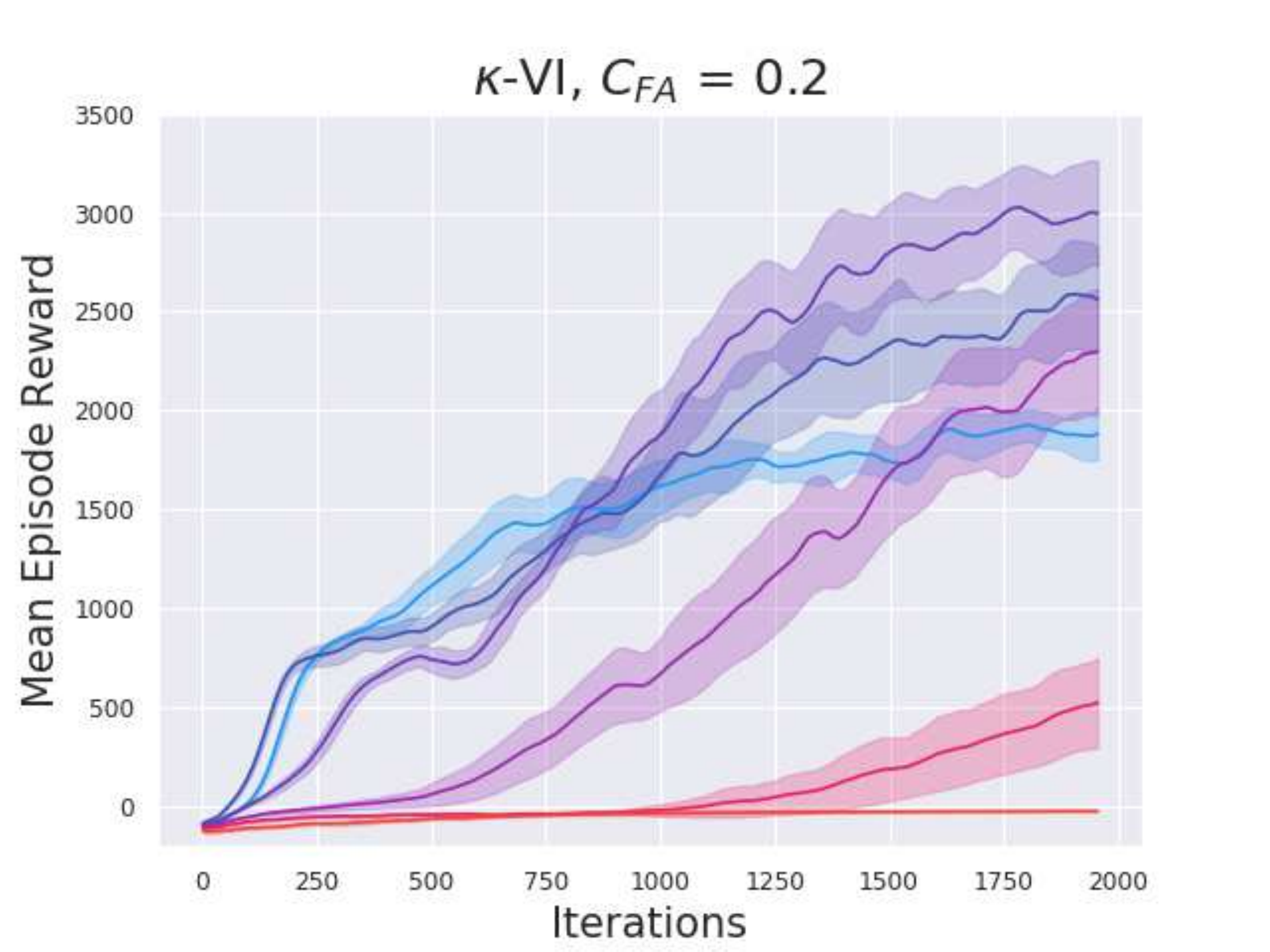} \\
        \includegraphics[scale=0.30]{Images/TRPO/legend_horizontal.pdf}
        \caption*{\textit{Ant-v2}}
    \end{subfigure}
    \caption{Training performance of the `naive' baseline $N_\kappa=T$ and $\kappa$-PI-TRPO, $\kappa$-VI-TRPO for $C_{FA}=0.2$ on Walker (top) and Ant (bottom). See Appendix~\ref{subsec: TRPO appendix CFA} for performance w.r.t.~different $C_{FA}$ values. Each iteration corresponds to roughly 1000 environment samples, and thus, the total number of training samples is 2 millions.}
    \label{fig:Walker-kPI-kVI}
\end{figure*}

In our experiments, we specifically focus on answering the following questions:
\begin{enumerate}
    \item Does the performance of DQN and TRPO improve when using them as $\kappa$-greedy solvers in $\kappa$-PI and $\kappa$-VI? Is there a performance tradeoff w.r.t.~to $\kappa$? 
    \item Following $\kappa$-PI and $\kappa$-VI, our DQN and TRPO implementations of these algorithms devote a significant $T_\kappa$ number of samples to each iteration. Is this needed or a `naive' choice of $T_\kappa=1$, or equivalently $N_\kappa=T$, works just well for all values of $\kappa$?
\end{enumerate}

We choose to test our $\kappa$-DQN and $\kappa$-TRPO algorithms on the Atari and MuJoCo benchmarks, respectively. Both of these algorithms use standard setups, including the use of the Adam optimizer for performing gradient descent, a discount factor of 0.99 across all tasks, target $Q$ value networks in the case of $\kappa$-DQN and an entropy regularizer with a coefficient of 0.01 in the case of $\kappa$-TRPO. We choose to run our experiments for multiple values of $\kappa$ between zero and one, which roughly follow a logarithmic scale. Note that just by using the definition of $\kappa$-greedy algorithms, we reduce to the base cases of DQN and TRPO when setting $\kappa=1.0$ for all experiments. This forms as one of the two baselines we consider in this work. The second baseline essentially refers to using our $\kappa$-greedy algorithms for a fixed $N_{\kappa}=T$ value. Thus, independent of the value of $\kappa$, the surrogate MDP is solved for a single time-step per each iteration. Below, we describe the experiments and results in further detail. The implementation details for the $\kappa$-DQN and $\kappa$-TRPO cases are provided in Appendix~\ref{supp: DQN}, Table~\ref{tab:Hyperparams-DQN} and Appendix~\ref{supp: TRPO}, Table~\ref{tab:Hyperparams-TRPO}, respectively.


\subsection{$\kappa$-PI-DQN and $\kappa$-VI-DQN Experiments}
\label{subsec:DQN-Experiments}


\begin{table*}[h]
\footnotesize
\begin{center}
\begin{tabular}{  c | c | c | c| c | c }\hline
 Domain & Alg. & $\kappa_{\mathrm{best}}$ &  $\kappa=0$ & DQN, $\kappa=1$ &  $N_\kappa=T$, $\kappa_{\mathrm{best}}$ ($\kappa$-PI)\\ \hline 
  \rowcolor{Grey}
  & $\kappa$-PI &\textbf{223{\tiny (\boldmath$\pm$7)}, \boldmath$\kappa$=0.84} &154{\tiny($\pm$3)}& &\\
  \rowcolor{Grey}
  \multirow{-2}{*}{Breakout} &$\kappa$-VI & 181{\tiny($\pm$7)}, $\kappa$=0.68& 174{\tiny($\pm$5)} & \multirow{-2}{*}{134{\tiny($\pm$4)}}& \multirow{-2}{*}{170{\tiny($\pm$2)}, $\kappa$=0.68} \\
  \multirow{2}{*}{SpaceInv.}& $\kappa$-PI&\textbf{755{\tiny(\boldmath$\pm$23)}, \boldmath$\kappa$=0.84}&613{\tiny($\pm$20)}& \multirow{2}{*}{656{\tiny($\pm$17)}}&\multirow{2}{*}{700{\tiny($\pm$21)}, $\kappa$=0.98}\\
  & $\kappa$-VI&712{\tiny($\pm$25)}, $\kappa$=0.92&687{\tiny($\pm$32)}&&\\
 \rowcolor{Grey}
 & $\kappa$-PI&\textbf{5159{\tiny(\boldmath$\pm$292)}, \boldmath$\kappa$=0.92}&2612{\tiny($\pm$238)}& &\\
 \rowcolor{Grey}
 \multirow{-2}{*}{Seaquest}& $\kappa$-VI&3253{\tiny($\pm$402)}, $\kappa$=0.84& 2680{\tiny($\pm$382)}&\multirow{-2}{*}{3099{\tiny($\pm$191)}}&\multirow{-2}{*}{3897{\tiny($\pm$218)}, $\kappa$=0.68}\\
 \multirow{2}{*}{Enduro}& $\kappa$-PI& \textbf{533{\tiny(\boldmath$\pm$12)}, \boldmath$\kappa$=0.84}& 478{\tiny($\pm$10)}& \multirow{2}{*}{224{\tiny($\pm$110)}}& \multirow{2}{*}{\textbf{535{\tiny(\boldmath$\pm$13)}, \boldmath$\kappa$=0.68}}\\
 & $\kappa$-VI& 486{\tiny($\pm$23)}, $\kappa$=0.84 & 443 {\tiny($\pm$90)} & & \\
 \rowcolor{Grey}
 & $\kappa$-PI&3849{\tiny($\pm$110)}, $\kappa$=1.0& 3103{\tiny($\pm$279)}& &\\
 \rowcolor{Grey}
 \multirow{-2}{*}{BeamRider}& $\kappa$-VI& \textbf{4277 {\tiny(\boldmath$\pm$269)}, \boldmath$\kappa$=0.84} & 3714 {\tiny($\pm$437)} &\multirow{-2}{*}{3849{\tiny($\pm$110)}}& \multirow{-2}{*}{3849{\tiny($\pm$110)}, $\kappa$=1.0} \\
 \multirow{2}{*}{Qbert}& $\kappa$-PI&\textbf{8157{\tiny(\boldmath$\pm$265)}, \boldmath$\kappa$=0.84}& 6719{\tiny($\pm$520)}& \multirow{2}{*}{7258{\tiny($\pm$385)}}& \multirow{2}{*}{\textbf{7968{\tiny(\boldmath$\pm$218)}, \boldmath$\kappa$=0.98}}\\
 & $\kappa$-VI& \textbf{8060 {\tiny(\boldmath$\pm$158)}, \boldmath$\kappa$=0.84} & 7563 {\tiny($\pm$398)} & & \\
 \hline
\end{tabular}
\end{center}
\caption{The final training performance of $\kappa$-PI-DQN and $\kappa$-VI-DQN on the Atari domains, for the hyper-parameter $C_{FA}=0.05$. The values are reported for a 95\% confidence interval across 10 random runs (\textit{empirical mean} $\pm$ 1.96 $\times$ \textit{empirical standard deviation} / $\sqrt{n=10}$). The best scores are in bold and multiple bold values for a domain denote an insignificant statistical difference between them.}
\label{tab:k-DQN}
\end{table*}

\begin{table*}
\footnotesize
\begin{tabular}{  c | c | c | c| c | c |c }\hline
 Domain & Alg. & $\kappa_{\mathrm{best}}$ &  $\kappa=0$ & TRPO, $\kappa=1$ &  $N_\kappa=T$, $\kappa_{\mathrm{best}}$ ($\kappa$-PI)& GAE, $\lambda_{\mathrm{best}}$\\ \hline 
 \rowcolor{Grey}
  & $\kappa$-PI & \textbf{1438{\tiny(\boldmath$\pm$188)}, \boldmath$\kappa$=0.68} & 1371 {\tiny($\pm$192)}& & &\\
  \rowcolor{Grey}
  \multirow{-2}{*}{Walker} &$\kappa$-VI & 954{\tiny($\pm$88)}, $\kappa$=0.68 & 830{\tiny($\pm$262)} & \multirow{-2}{*}{594 {\tiny($\pm$99)}}& \multirow{-2}{*}{1082{\tiny($\pm$110)}, $\kappa$=0.0} & \multirow{-2}{*}{\textbf{1601{\tiny(\boldmath$\pm$190)}, \boldmath$\lambda$=0.0}}\\
  \multirow{2}{*}{Ant}& $\kappa$-PI&1377{\tiny($\pm$183)}, $\kappa$=0.68& 1006{\tiny($\pm$106)} &\multirow{2}{*}{-19{\tiny($\pm$1)}}&\multirow{2}{*}{1090{\tiny($\pm$99)}, $\kappa$=0.68}&\multirow{2}{*}{1094{\tiny($\pm$139)}, $\lambda$=0.0}\\
  & $\kappa$-VI& \textbf{2998{\tiny(\boldmath$\pm$264)}, \boldmath$\kappa$=0.68}&1879{\tiny($\pm$128)} & & & \\
 \rowcolor{Grey}
 & $\kappa$-PI& \textbf{1334{\tiny(\boldmath$\pm$151)}, \boldmath$\kappa$=0.36} & 907{\tiny($\pm$176)} &&&\\
 \rowcolor{Grey}
  \multirow{-2}{*}{HalfCheetah}& $\kappa$-VI& \textbf{1447{\tiny(\boldmath$\pm$346)}, \boldmath$\kappa$=0.36} &1072{\tiny($\pm$30)} &\multirow{-2}{*}{-18{\tiny($\pm$87)}}&\multirow{-2}{*}{1195{\tiny($\pm$218)}, $\kappa$=0.36}&\multirow{-2}{*}{\textbf{1322{\tiny(\boldmath$\pm$213)}, \boldmath$\lambda$=0.36}} \\
 \multirow{2}{*}{HumanoidStand}& $\kappa$-PI& \textbf{72604{\tiny(\boldmath$\pm$1219)}, \boldmath$\kappa$=0.99} & 52936{\tiny($\pm$1529)} &\multirow{2}{*}{68143{\tiny($\pm$1031)}}&\multirow{2}{*}{\textbf{71331{\tiny(\boldmath$\pm$1149)}, \boldmath$\kappa$=0.98}}&\multirow{2}{*}{\textbf{71932{\tiny(\boldmath$\pm$2122)}, \boldmath$\lambda$=0.98}}\\
  & $\kappa$-VI& \textbf{72821{\tiny(\boldmath$\pm$908)}, \boldmath$\kappa$=0.99} &51148{\tiny($\pm$1377)} & & & \\
 \rowcolor{Grey}
 & $\kappa$-PI& \textbf{107{\tiny(\boldmath$\pm$12)}, \boldmath$\kappa$=1.0} & 42{\tiny($\pm$3)} &&&\\
 \rowcolor{Grey}
 \multirow{-2}{*}{Swimmer} & $\kappa$-VI& \textbf{114{\tiny(\boldmath$\pm$15)}, \boldmath$\kappa$=1.0}& 46{\tiny($\pm$1)} & \multirow{-2}{*}{\textbf{107{\tiny(\boldmath$\pm$12)}}}&\multirow{-2}{*}{\textbf{107{\tiny(\boldmath$\pm$12)}, \boldmath$\kappa$=1.0}}&\multirow{-2}{*}{\textbf{103{\tiny(\boldmath$\pm$13)}, \boldmath$\lambda=1.0$}} \\
 \multirow{2}{*}{Hopper}& $\kappa$-PI& \textbf{1486{\tiny(\boldmath$\pm$324)}, \boldmath$\kappa$=0.68} & 1012{\tiny($\pm$263)} &\multirow{2}{*}{1142{\tiny($\pm$141)}}&\multirow{2}{*}{\textbf{1434{\tiny(\boldmath$\pm$129)}, \boldmath$\kappa$=0.98}}&\multirow{2}{*}{\textbf{1600{\tiny(\boldmath$\pm$134)}, \boldmath$\lambda=0.84$}}\\
  & $\kappa$-VI& 1069{\tiny($\pm$76)}, $\kappa$=0.92& 531{\tiny($\pm$125)} & & & \\
 \hline
\end{tabular}
\caption{The final training performance of $\kappa$-PI-TRPO and $\kappa$-VI-TRPO on the MuJoCo domains, for the hyper-parameter $C_{FA}=0.2$. The values are reported for a 95\% confidence interval across 10 random runs (\textit{empirical mean} $\pm$ 1.96 $\times$ \textit{empirical standard deviation} / $\sqrt{n=10}$). The best scores are in bold and multiple bold values for a domain denote an insignificant statistical difference between them.}
\label{tab:k-TRPO}
\end{table*}

In this section, we empirically analyze the performance of the $\kappa$-PI-DQN and $\kappa$-VI-DQN algorithms on the Atari domains: Breakout, SpaceInvaders, Seaquest, Enduro, BeamRider, and Qbert~\citep{bellemare2013arcade}. We start by performing an ablation test on three values of hyper-parameter $C_{FA}=\{0.001,0.05,0.2\}$ on the Breakout domain. The value of $C_{FA}$ sets the number of samples per iteration $T_\kappa$ (Eq.~\ref{eq: final accuracy}) and the total number of iterations $N_\kappa$ (Eq.~\ref{eq: number of samples per iteration}). The total number of samples is set to $T \simeq 10^6$. This value represents the number of samples after which our DQN-based algorithms approximately converge. For each value of $C_{FA}$, we test $\kappa$-PI-DQN and $\kappa$-VI-DQN for several $\kappa$ values. In both algorithms, the best performance was obtained with $C_{FA}=0.05$. Therefore, $C_{FA}$ is set to $0.05$ for all our experiments with other Atari domains. 

Figure~\ref{fig:Breakout-kPI-kVI} shows the training performance of $\kappa$-PI-DQN and $\kappa$-VI-DQN on Breakout and SpaceInvaders for the best value of $C_{FA}=0.05$, as well as for the `naive' baseline $T_\kappa=1$, or equivalently $N_\kappa=T$. The results on Breakout for the other values of $C_{FA}$ and the training plots for all other Atari domains (with $C_{FA}=0.05$) are reported in Appendices~\ref{subsec: appendix CFA} and~\ref{subsec:add-plot-DQN}, respectively. Table~\ref{tab:k-DQN} shows the final training performance of $\kappa$-PI-DQN and $\kappa$-VI-DQN on the Atari domains with $C_{FA}=0.05$. Note that the scores reported in~Table~\ref{tab:k-DQN} are the actual returns on the Atari domains, while the vertical axis in the plots of Figure~\ref{fig:Breakout-kPI-kVI} corresponds to a scaled return. We plot the scaled return, since this way it can be easier to reproduce our results using the OpenAI Baselines codebase~\citep{stable-baselines}.
%
%
%
%

Our results exhibit that both $\kappa$-PI-DQN and $\kappa$-VI-DQN improve the performance of DQN ($\kappa=1$). Moreover, they show that setting $N_\kappa=T$ leads to a clear degradation of the final training performance on all of the domains except for Enduro, which gets to approximately the same score. Although the performance degrades, the results for $N_\kappa=T$ are still better than for DQN.



\subsection{$\kappa$-PI-TRPO and $\kappa$-VI-TRPO Experiments}
\label{subsec:TRPO-Experiments}

In this section, we empirically analyze the performance of the $\kappa$-PI-TRPO and $\kappa$-VI-TRPO algorithms on the MuJoCo \citep{todorov2012mujoco} based OpenAI Gym domains: Walker2d-v2, Ant-v2, HalfCheetah-v2, HumanoidStandup-v2, Swimmer-v2, and Hopper-v2 \cite{brockman2016openai}. As in Section~\ref{subsec:DQN-Experiments}, we start by performing an ablation test on the parameter $C_{FA}=\{0.001, 0.05, 0.2\}$ on the Walker domain. We set the total number of iterations to $2000$, with each iteration consisting $1000$ samples. Thus, the total number of samples is $T\simeq 2\times 10^6$. This is the number of samples after which our TRPO-based algorithms approximately converge. For each value of $C_{FA}$, we test $\kappa$-PI-TRPO and $\kappa$-VI-TRPO for several $\kappa$ values. In both algorithms, the best performance was obtained with $C_{FA}=0.2$, and thus, we set $C_{FA}=0.2$ in our experiments with other MuJoCo domains. 

\begin{figure*}[t]
    \centering
    \includegraphics[scale=0.22]{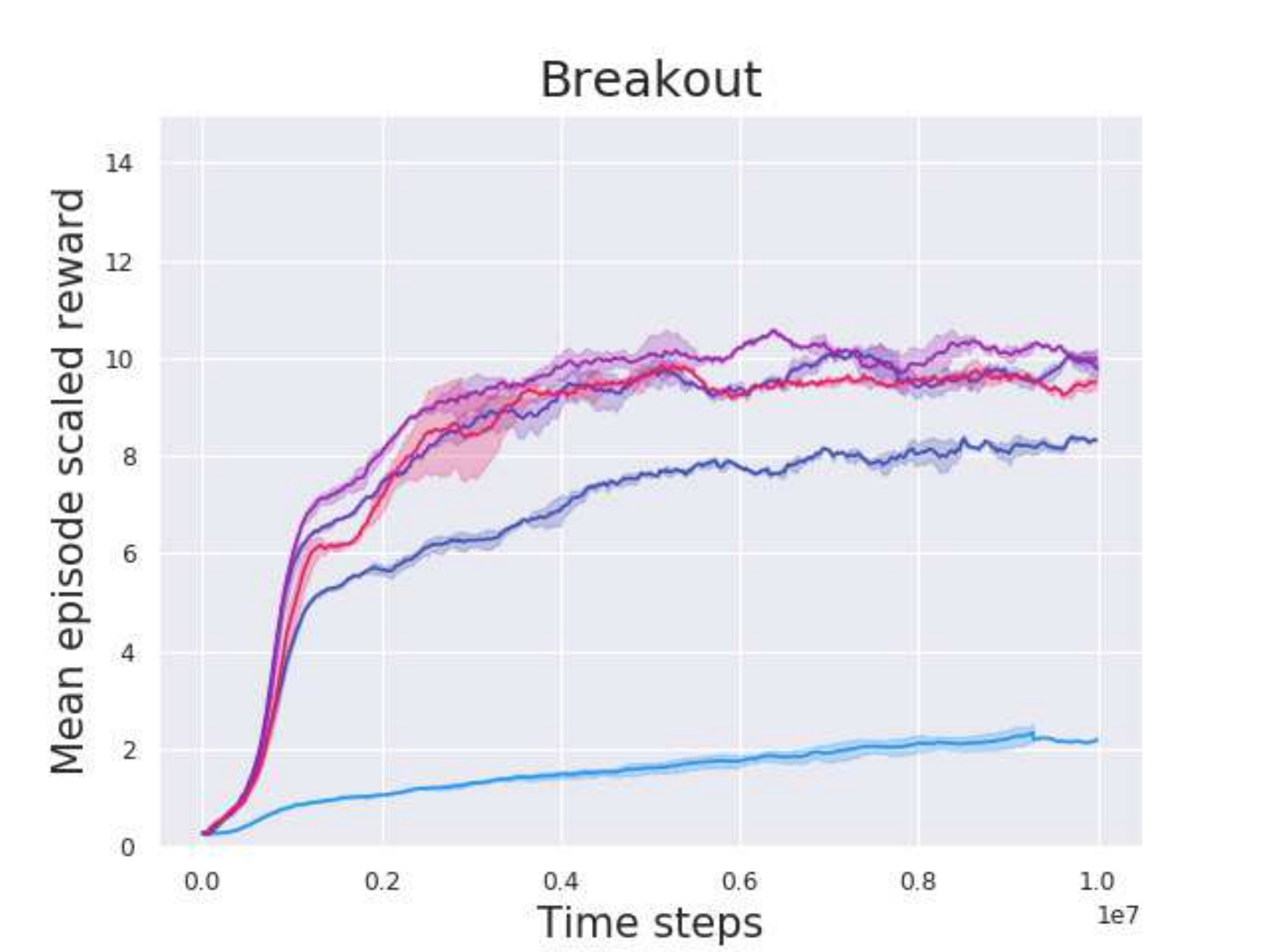}
    \includegraphics[scale=0.22]{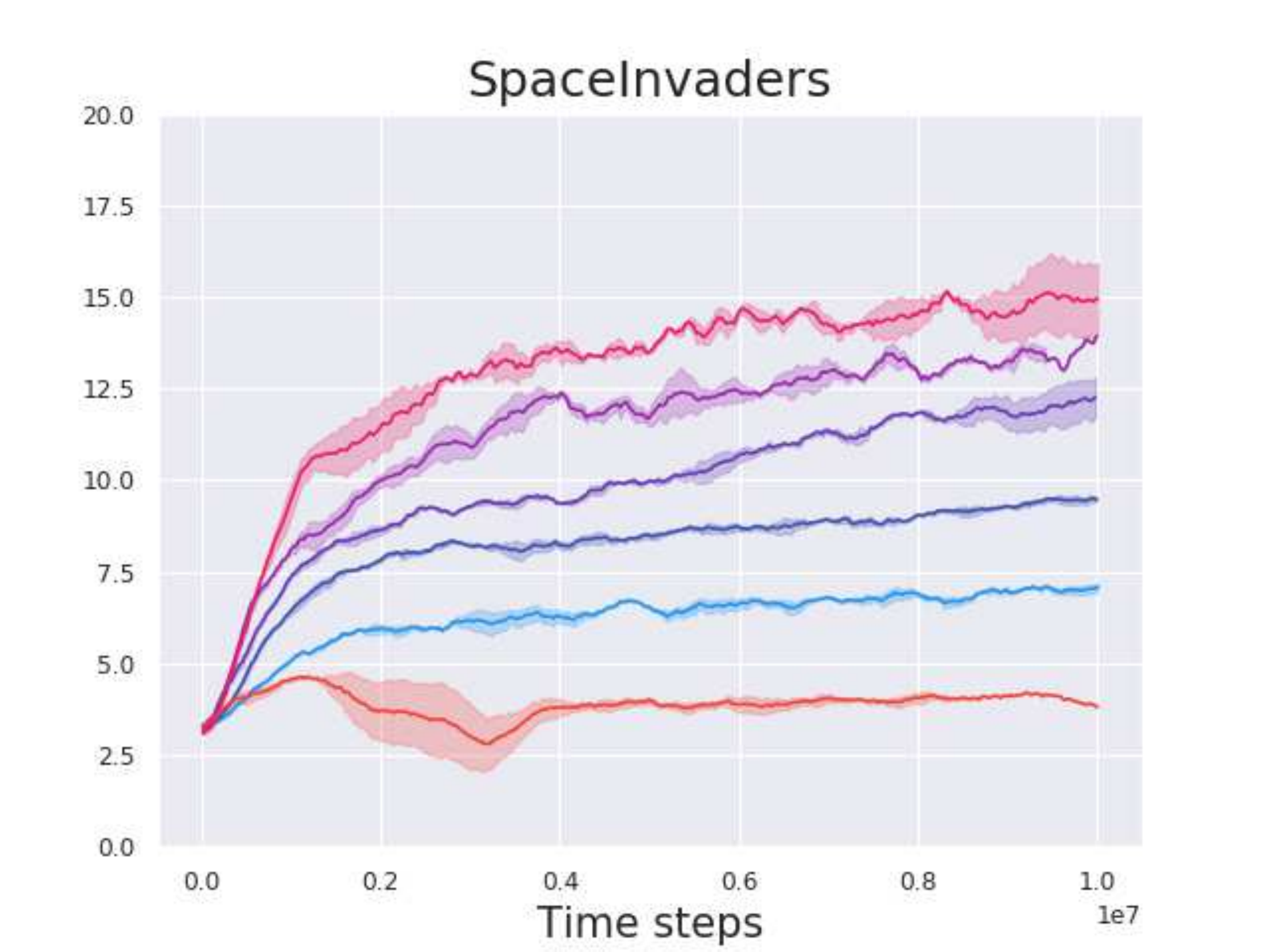} \\
    \includegraphics[scale=0.2]{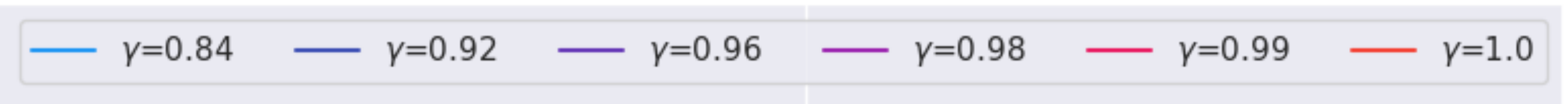} \\
    \includegraphics[scale=0.22]{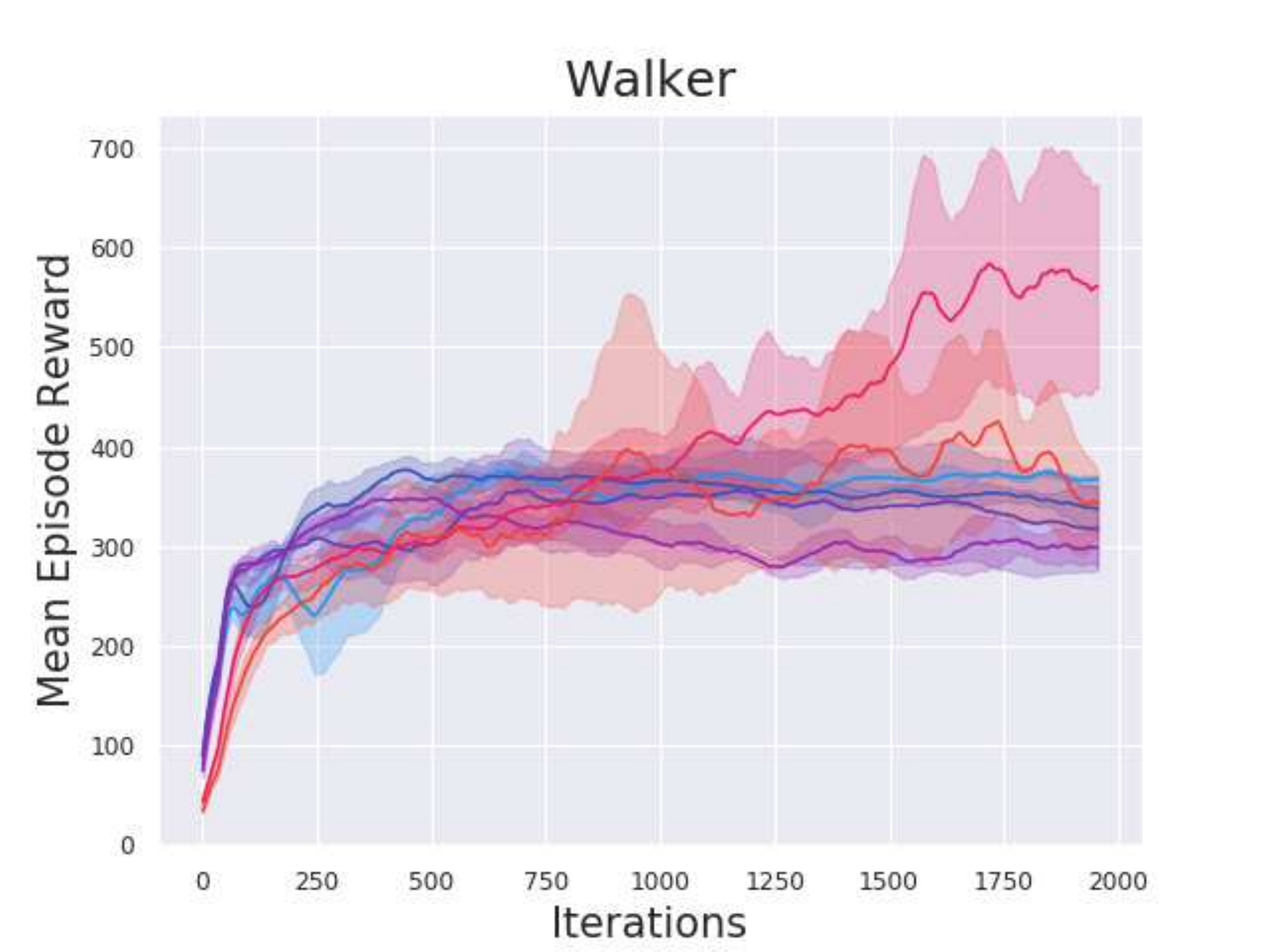} 
    \includegraphics[scale=0.22]{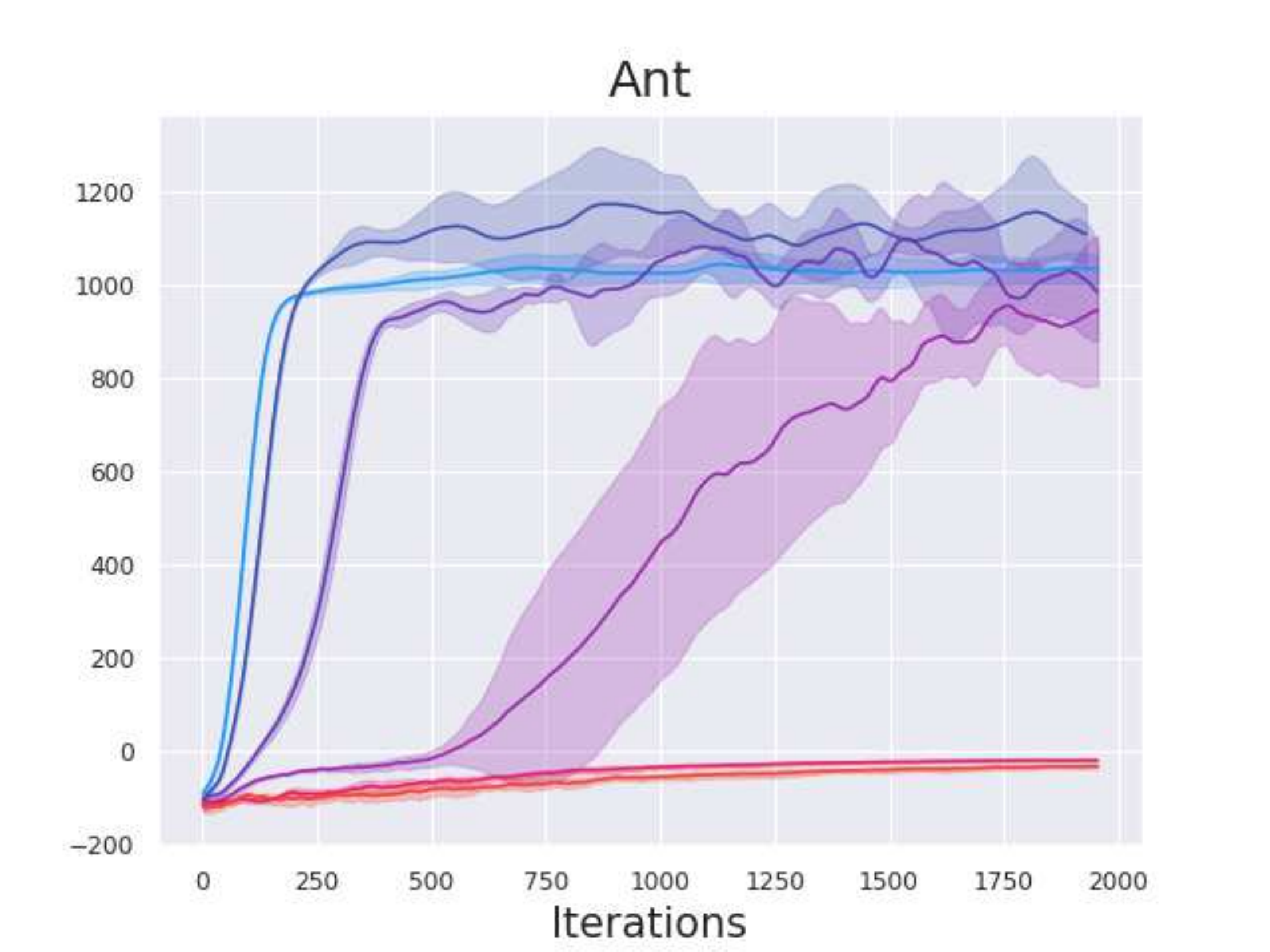} \\
    \includegraphics[scale=0.2]{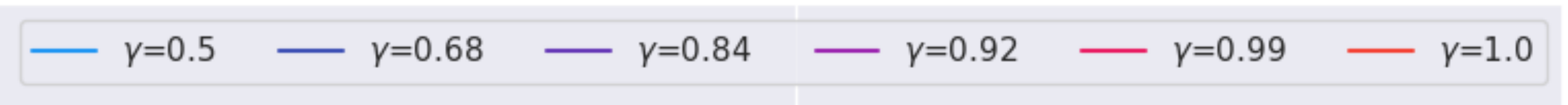}
    \caption{Lowering the discount factor $\gamma$ for Atari domains Breakout and SpaceInvaders and MuJoCo domains Walker-v2 and Ant-v2}
    \label{fig:lower discount}
\end{figure*}

Figure~\ref{fig:Walker-kPI-kVI} shows the training performance of $\kappa$-PI-TRPO and $\kappa$-VI-TRPO on Walker and Ant domains for the best value of $C_{FA}=0.2$, as well as for the `naive' baseline $T_\kappa=1$, or equivalently $N_\kappa=T$. The results on Walker for the other $C_{FA}$ values and the other MuJoCo domains (with $C_{FA}=0.2$) are reported in Appendices~\ref{subsec: TRPO appendix CFA} and \ref{subsec:add-plot-TRPO}, respectively. Table~\ref{tab:k-TRPO} shows the final training performance of $\kappa$-PI-TRPO and $\kappa$-VI-TRPO on the MuJoCo domains with $C_{FA}=0.2$. 

The results exhibit that both $\kappa$-PI-TRPO and $\kappa$-VI-TRPO yield better performance than TRPO ($\kappa=1$). Furthermore, they show that the algorithms with $C_{FA}=0.2$ perform better than with $N_\kappa=T$ for three out of six domains (Walker, Ant and HalfCheetah) and equally well for the remaining three (HumanoidStandup, Swimmer and Hopper).

\section{Discussion and Related Work}
\label{sec:disc-related-work}

{\bf Comparison with GAE:} There is a close connection between the Generalized Advantage Estimation (GAE) algorithm~\cite{schulman2015high} and $\kappa$-PI. In GAE, the policy is updated by the following gradient:
\begin{align}
\nabla_\theta \E_{s\sim \mu}\big[V^{\pi_\theta}(s)\big] &= \E_{s\sim d_\pi^\mu}\big[\nabla_\theta \log \pi_\theta(s) \sum_t (\gamma\lambda)^t \delta_V\big], \nonumber \\
\delta_V &= r_t + \gamma V_{t+1} - V_t,
\label{eq:GAE1}
\end{align}
where $d_\pi^\mu$ is the occupancy measure of policy $\pi$. Eq.~\ref{eq:GAE1} can be interpreted as a gradient in a $\gamma\lambda$-discounted MDP with shaped rewards $\delta_V$, which we refer to as $\mathcal{M}_{\gamma\lambda}(\delta_V)$. As noted in~\citet[Sec.~6]{efroni2018beyond}, an optimal policy $\pi^*_{\gamma\lambda}$ of the MDP $\mathcal{M}_{\gamma\lambda}(\delta_V)$ is also optimal in $\mathcal{M}_{\gamma\kappa}(V)$ with $\kappa=\lambda$. This means that $\pi^*_{\gamma\lambda}$ is the $\kappa$-greedy policy w.r.t.~$V$. 

Comparing the two algorithms, we notice that GAE is conceptually similar to $\kappa$-PI-TRPO with $T_\kappa =1$, or equivalently $N_\kappa=T$. In fact, in the case of $T_\kappa =1$, we can obtain a pseudo-code of GAE by removing Line~5 (no need to estimate the value of the surrogate MDP) and replacing $r_t(\kappa,V_\phi)$ with the TD-error of the original MDP on Line~4 in Algorithm~\ref{alg:kappaPI-TRPO}. 

In Section~\ref{subsec:TRPO-Experiments}, we compared the empirical performance of GAE with that of $\kappa$-PI-TRPO and $\kappa$-VI-TRPO (see Figure~\ref{fig:Walker-kPI-kVI} and Table~\ref{tab:k-TRPO}). The results show that $\kappa$-PI-TRPO and $\kappa$-VI-TRPO perform better than or on par with GAE. Moreover, we observe that in most domains, the performance of GAE is equivalent to that of $\kappa$-PI-TRPO with $N_\kappa=T$. This is in accordance with the description above connecting GAE to our naive baseline implementation. We report all GAE results in Appendix~\ref{subsec:add-plot-TRPO}. 


\begin{remark}[GAE Implementation]
In the OpenAI implementation of GAE, the value network is updated w.r.t.~to the target $\sum_{t} (\gamma\lambda)^t r_t$, whereas in the GAE paper~\cite{schulman2015high}, $\sum_{t} \gamma^t r_t$ is used as the target. We chose the latter form in our implementation of GAE to be in accord with the paper. 
\end{remark}




{\bf Lowering Discount Factor in DQN and TRPO:} To show the advantage of $\kappa$-PI and $\kappa$-VI based algorithms over simply lowering the discount factor $\gamma$, we test the performance of the ``vanilla'' DQN and TRPO algorithms with values of $\gamma$ lower than the one previously used (i.e.,~$\gamma=0.99$). 
%
As evident from Figure~\ref{fig:lower discount}, only in the Ant domain, this approach results in an improved performance (for $\gamma=0.68$). On the other hand, in the Ant domain, the performance of $\kappa$-PI-TRPO, and especially $\kappa$-VI-TRPO, surpasses that of TRPO with the lower value of $\gamma=0.68$. In Breakout, SpaceInvaders, and Walker, the performance of DQN and TRPO worsens or remains unchanged when we lower the discount factor (DQN and TRPO do not benefit from lowering $\gamma$), while our $\kappa$-PI and $\kappa$-VI based algorithms perform better with lowering the value of $\kappa$ (note that our algorithms are reduced to DQN and TRPO for $\kappa=1$).


%
\begin{remark}
While we observe better performance for smaller $\gamma$ values in some MuJoCo domains, e.g.,~$\gamma=0.68$ in Ant, lowering $\gamma$ always results in inferior performance in the Atari domains. This is due to the fact that a number of MuJoCo domains, such as Ant, are inherently short-horizon decision problems, and thus, their performance does not degrade (even sometimes improves) with lowering the discount factor. On the other hand, the Atari problems are generally not short-horizon, and thus, their performance degrades with lowering $\gamma$. 
\end{remark}




\section{Conclusion and Future Work}

In this paper, we studied the use of multi-step greedy policies in model-free RL and showed that in most problems, the algorithms derived from this formulation achieve a better performance than their single-step counterparts. We adopted the $\kappa$-greedy formulation of multi-step greedy policies~\cite{efroni2018beyond} and derived four model-free RL algorithms. The main component of the policy and value iteration algorithms derived from this formulation, $\kappa$-PI and $\kappa$-VI, is solving a surrogate decision problem with a shaped reward and a smaller discount factor. Our algorithms use popular deep RL algorithms, DQN and TRPO, to solve the surrogate decision problem, and thus, we refer to them as $\kappa$-PI-DQN, $\kappa$-VI-DQN, $\kappa$-PI-TRPO, and $\kappa$-VI-TRPO. We empirically evaluated our proposed algorithms and compared them with DQN, TRPO, and GAE on Atari and MuJoCo benchmarks. Our experiments show that for a large range of $\kappa$, our algorithms perform better than DQN and TRPO. Furthermore, we proposed a recipe to allocate the total sample budget to the evaluation and improvement phases of our algorithms, and empirically demonstrated the importance of this allocation. We also showed how GAE can be derived by minor modifications to $\kappa$-PI-TRPO, and thus, is a $\kappa$-greedy RL algorithm. Finally, we showed the advantage of multi-step greedy formulation over lowering the discount factor in DQN and TRPO. Our results indicate that while the performance of DQN and TRPO degrades with lowering the discount factor, our multi-step greedy algorithms improve over DQN and TRPO.

An interesting future direction would be to use other multi-step greedy formulations~\citep{bertsekas1996neuro,bertsekas2018feature,efroni2018beyond,sun2018dual,shani2019exploration} to derive model-free RL algorithms. Another direction is to use multi-step greedy in model-based RL (e.g.,~\citealp{kumar2016optimal,talvitie2017self,luo2018algorithmic,janner2019trust}) and solve the surrogate decision problem with an approximate model. We conjecture that in this case one may set $\kappa$ -- or more generally, the planning horizon -- as a function of the quality of the approximate model: gradually increasing $\kappa$ as the approximate model gets closer to the real one. We leave theoretical and empirical study of this problem for future work. Finally, we believe using adaptive $\kappa$ would greatly improve the performance of our proposed algorithms. We leave verifying this and how $\kappa$ should change as a function of errors in gradient and value estimation for future work. 







\bibliography{main_paper}
\bibliographystyle{icml2020}

\clearpage
\onecolumn


\def\thesection{\Alph{section}}
\clearpage
\newpage

\xpretocmd{\part}{\setcounter{section}{0}}{}{}

\part*{Appendix}
\section{$\kappa$-PI-DQN and $\kappa$-VI-DQN Algorithms}
\label{supp: DQN}

\subsection{Detailed Pseudo-codes} \label{supp: pseudo code k-VI k-PI DQN}

In this section, we report the detailed pseudo-codes of $\kappa$-PI-DQN and $\kappa$-VI-DQN algorithms, described in Section 4.3, side-by-side. 

\begin{algorithm}
    \caption{$\kappa$-PI-DQN}
    \label{alg:kappaPI-DQN-Full}
    \begin{algorithmic}[1]
        \STATE {\bfseries Initialize} replay buffer $\mathcal{D}$; $\;Q$-networks $Q_{\theta}$ and $Q_{\phi}$ with random weights $\theta$ and $\phi$; \\
        \STATE {\bfseries Initialize} target networks $Q'_{\theta}$ and $Q'_{\phi}$ with weights $\;\theta' \leftarrow \theta\;$ and $\;\phi' \leftarrow \phi$; \\
        \FOR{ $i = 0, \ldots, N_\kappa-1$}
            \STATE {\color{gray}\# Policy Improvement}
            \FOR{$t =1,\ldots,T_\kappa$}
                \STATE Select $a_t$ as an $\epsilon$-greedy action w.r.t.~$Q_{\theta}(s_t, a)$; 
                \STATE Execute $a_t$, observe $r_t$ and $s_{t+1}$, and store the tuple $(s_t,a_t,r_t,s_{t+1})$ in $\mathcal{D}$;
                \STATE Sample a random mini-batch $\{(s_j, a_j, r_j, s_{j+1})\}_{j=1}^N$ from $\mathcal{D}$;
                \STATE Update $\theta$ by minimizing the following loss function: \\ 
                \STATE $\qquad \mathcal{L}_{\mathcal{Q}_\theta} = \frac{1}{N} \sum_{j=1}^N \big[Q_\theta(s_j, a_j) - \big(r_j(\kappa, V_\phi) + \gamma \kappa \; \mathrm{max}_{a} Q'_\theta(s_{j+1},a)\big)\big]^2$, $\quad$ where \\
                \STATE $\qquad V_\phi(s_{j+1})=Q_\phi(s_{j+1},\pi_{i-1}(s_{j+1}))\quad$ and $\quad\pi_{i-1}(s_{j+1})\in \arg\max_a Q'_\theta(s_{j+1},a)$;
                \STATE Copy $\theta$ to $\theta'$ occasionally $\quad (\theta'\gets\theta)$;
            \ENDFOR
            \STATE {\color{gray}\# Policy Evaluation}
            \STATE Set $\pi_i(s)\in \arg\max_a Q'_{\theta}(s,a)$;
            \FOR{$t'=1,\ldots,T(\kappa)$}
                \STATE Sample a random mini-batch $\{(s_j, a_j, r_j, s_{j+1})\}_{j=1}^N$ from $\mathcal{D}$;
                \STATE Update $\phi$ by minimizing the following loss function: 
                \STATE $\qquad \mathcal{L}_{\mathcal{Q}_\phi} = \frac{1}{N} \sum_{j=1}^N \big[Q_{\phi}(s_j, a_j) - (r_j + \gamma Q'_{\phi}(s_{j+1}, \pi_i(s_{j+1})))\big]^2$;
                \STATE  Copy $\phi$ to $\phi'$ occasionally $\quad (\phi'\gets\phi)$;
            \ENDFOR
            \ENDFOR
    \end{algorithmic}
\end{algorithm}

\begin{algorithm}
    \caption{$\kappa$-VI-DQN}
    \label{alg:kappaVI-DQN-Full}
    \begin{algorithmic}[1]
        \STATE {\bfseries Initialize} replay buffer $\mathcal{D}$; $\;Q$-networks $Q_{\theta}$ and $Q_{\phi}$ with random weights $\theta$ and $\phi$;  \\
        \STATE {\bfseries Initialize} target network $Q'_\theta$ with weights $\;\theta' \leftarrow \theta$; \\
        \FOR{ $i = 0, \ldots, N_\kappa-1$}
        \STATE {\color{gray}\# Evaluate $T_\kappa V_\phi$ and the $\kappa$-greedy policy w.r.t. $V_\phi$ }
            \FOR{$t=1,\ldots,T_\kappa$}
                \STATE Select $a_t$ as an $\epsilon$-greedy action w.r.t.~$Q_\theta(s_t,a)$; 
                \STATE Execute $a_t$, observe $r_t$ and $s_{t+1}$, and store the tuple $(s_t,a_t,r_t,s_{t+1})$ in $\mathcal{D}$;
                \STATE Sample a random mini-batch $\{(s_j,a_j,r_j,s_{j+1})\}_{j=1}^N$ from $\mathcal{D}$;
                \STATE Update $\theta$ by minimizing the following loss function: \\ 
                \STATE $\qquad \mathcal{L}_{\mathcal{Q}_\theta} = \frac{1}{N} \sum_{j=1}^N \big[Q_\theta(s_j, a_j) - (r_j(\kappa, V_\phi) + \kappa \gamma \; \mathrm{max}_{a} Q'_\theta(s_{j+1},a))\big]^2$, $\quad$ where \\
                \STATE $\qquad V_\phi(s_{j+1}) = Q_\phi(s_{j+1}, \pi(s_{j+1}))\quad$ and $\quad\pi(s_{j+1})\in \arg\max_a Q_{\phi}(s_{j+1},a)$;
                \STATE  Copy $\theta$ to $\theta'$ occasionally $\quad (\theta'\gets\theta)$;
            \ENDFOR
            \STATE Copy $\theta$ to $\phi\quad (\phi\gets \theta)$
            \ENDFOR
    \end{algorithmic}
\end{algorithm}

\newpage


\subsection{Ablation Test for $C_{\text{FA}}$}
\label{subsec: appendix CFA}

\begin{figure*}[ht]
    \centering
    \includegraphics[scale=0.20]{Images/DQN/K-PI/Breakout/Breakout_KPI_nkappa.pdf}
    \includegraphics[scale=0.20]{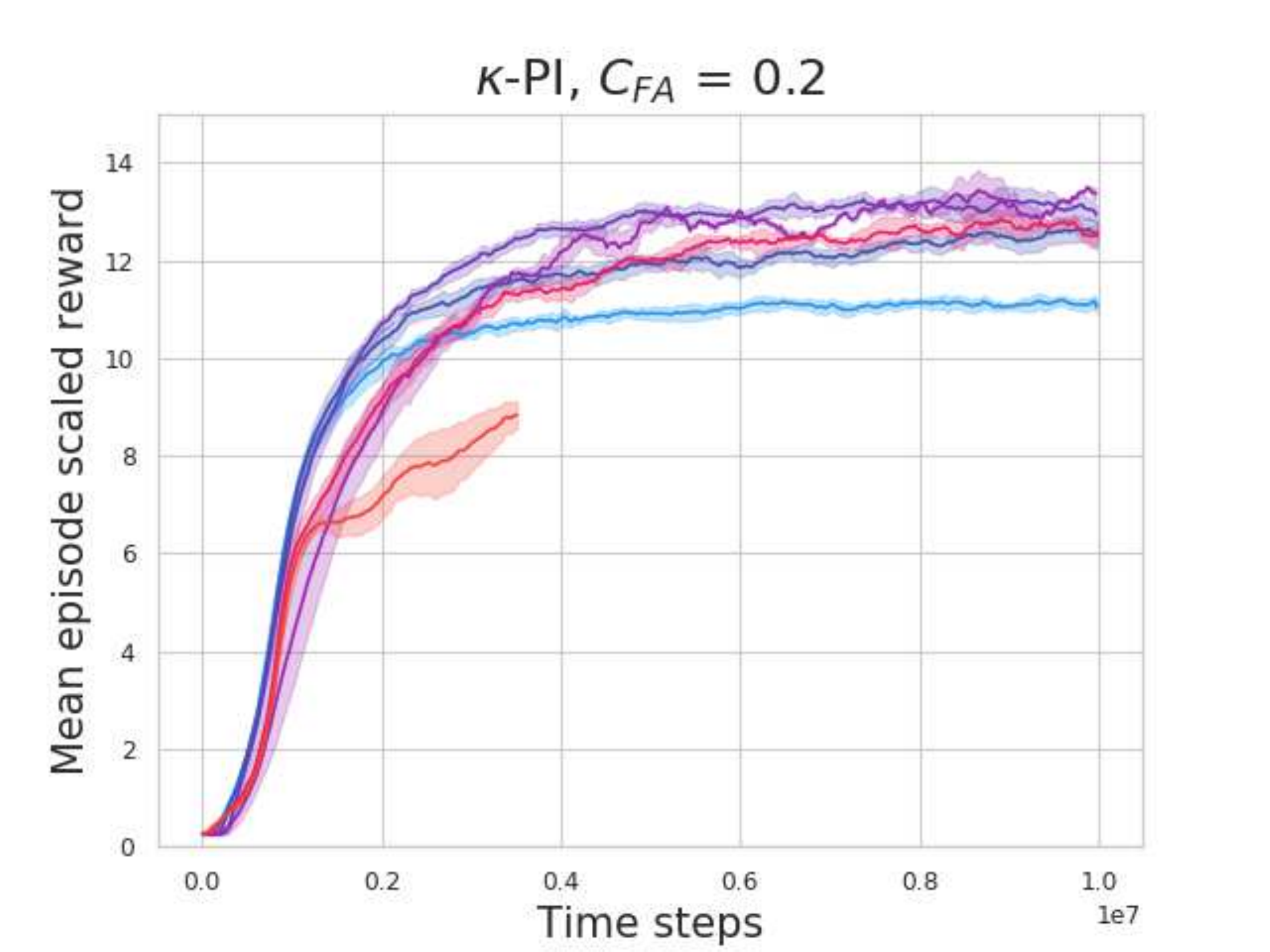}
    \includegraphics[scale=0.20]{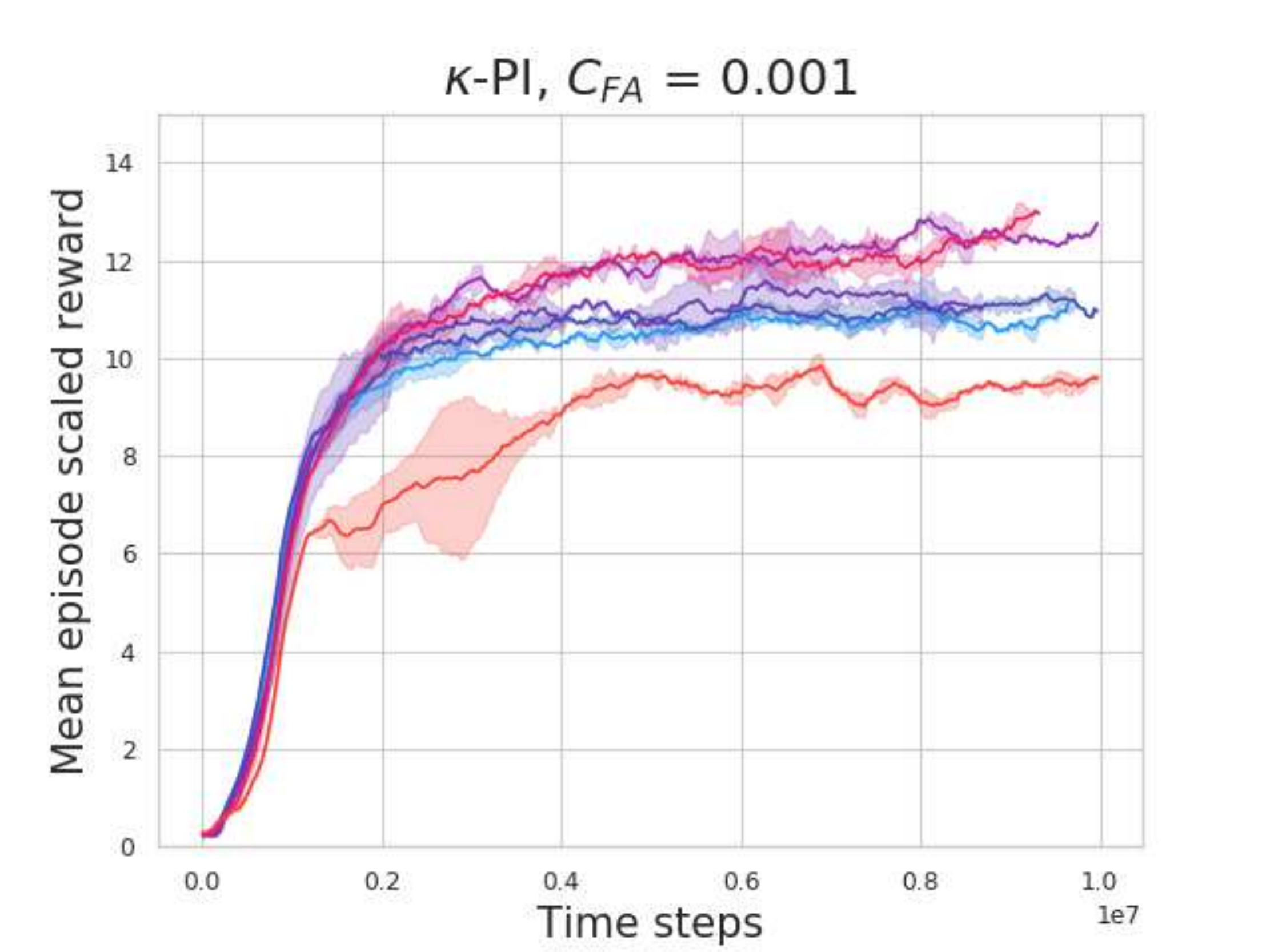}
    \includegraphics[scale=0.3]{Images/TRPO/legend_horizontal.pdf}
    \caption{Performance of $\kappa$-PI-DQN and $\kappa$-VI-DQN on Breakout for different values of $C_{\text{FA}}$.}
    \label{fig:Breakout}
\end{figure*}

\begin{table*}
\small
\begin{center}
\begin{tabular}{ c | c } 
    Hyperparameter & Value \\
    \hline 
    Horizon (T) & 1000 \\
    Adam stepsize & $1 \times 10^{-4}$ \\
    Target network update frequency & 1000 \\
    Replay memory size & 100000 \\
    Discount factor & 0.99 \\
    Total training time steps & 10000000 \\
    Minibatch size & 32 \\
    Initial exploration & 1 \\
    Final exploration & 0.1 \\
    Final exploration frame & 1000000 \\
    \#Runs used for plot averages & 10 \\
    Confidence interval for plot runs & $\sim$ 95\%
\end{tabular}
\end{center}
\caption{Hyperparameters for $\kappa$-PI-DQN and $\kappa$-VI-DQN.}
\label{tab:Hyperparams-DQN}
\end{table*}


\subsection{$\kappa$-PI-DQN and $\kappa$-VI-DQN Plots}
\label{subsec:add-plot-DQN}

In this section, we report additional results of the application of $\kappa$-PI-DQN and $\kappa$-VI-DQN on the Atari domains. A summary of these results has been reported in Table 1 in the main paper. 

\begin{figure*}[ht]
    \centering
    \includegraphics[scale=0.20]{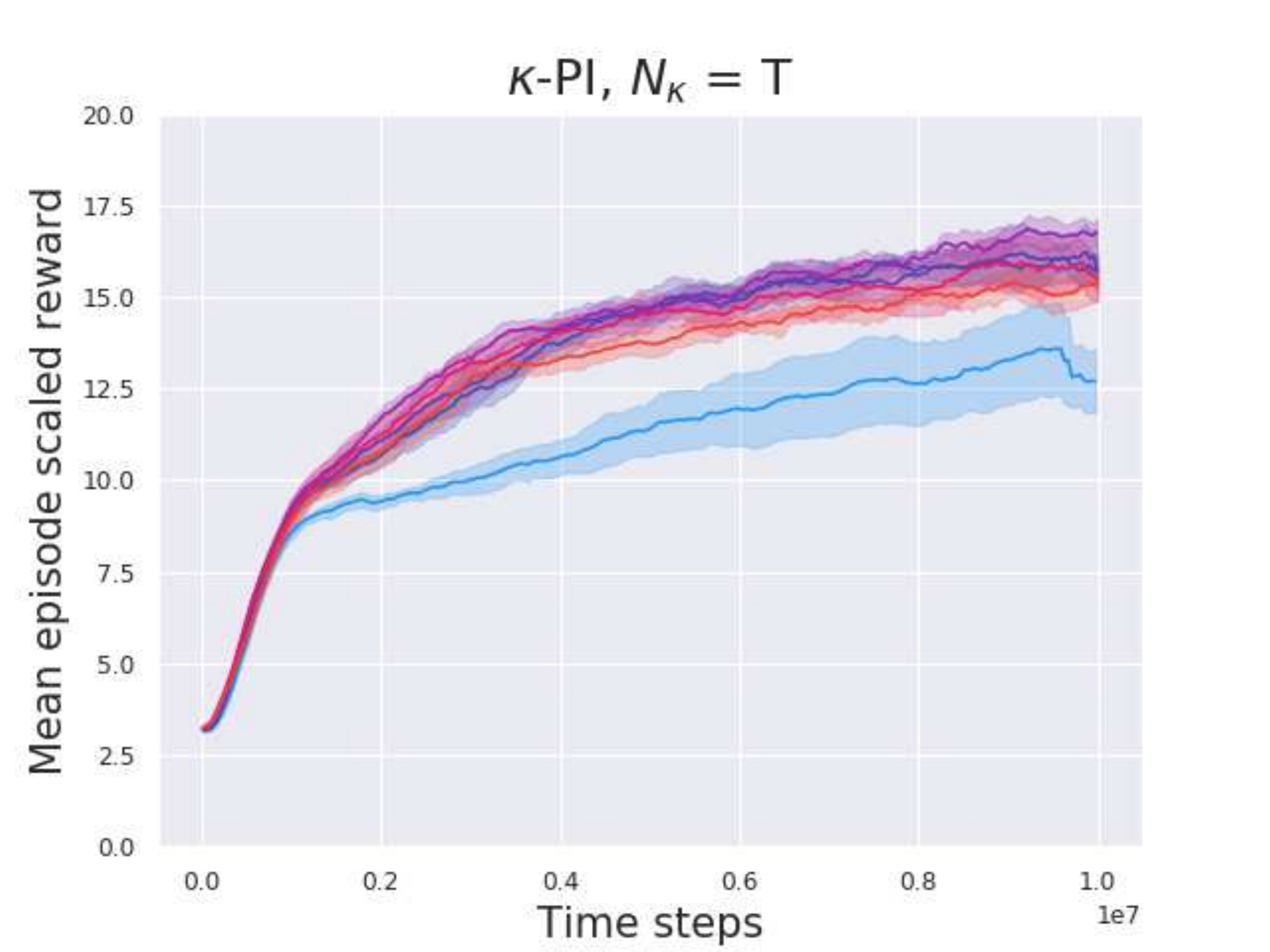}
    \includegraphics[scale=0.20]{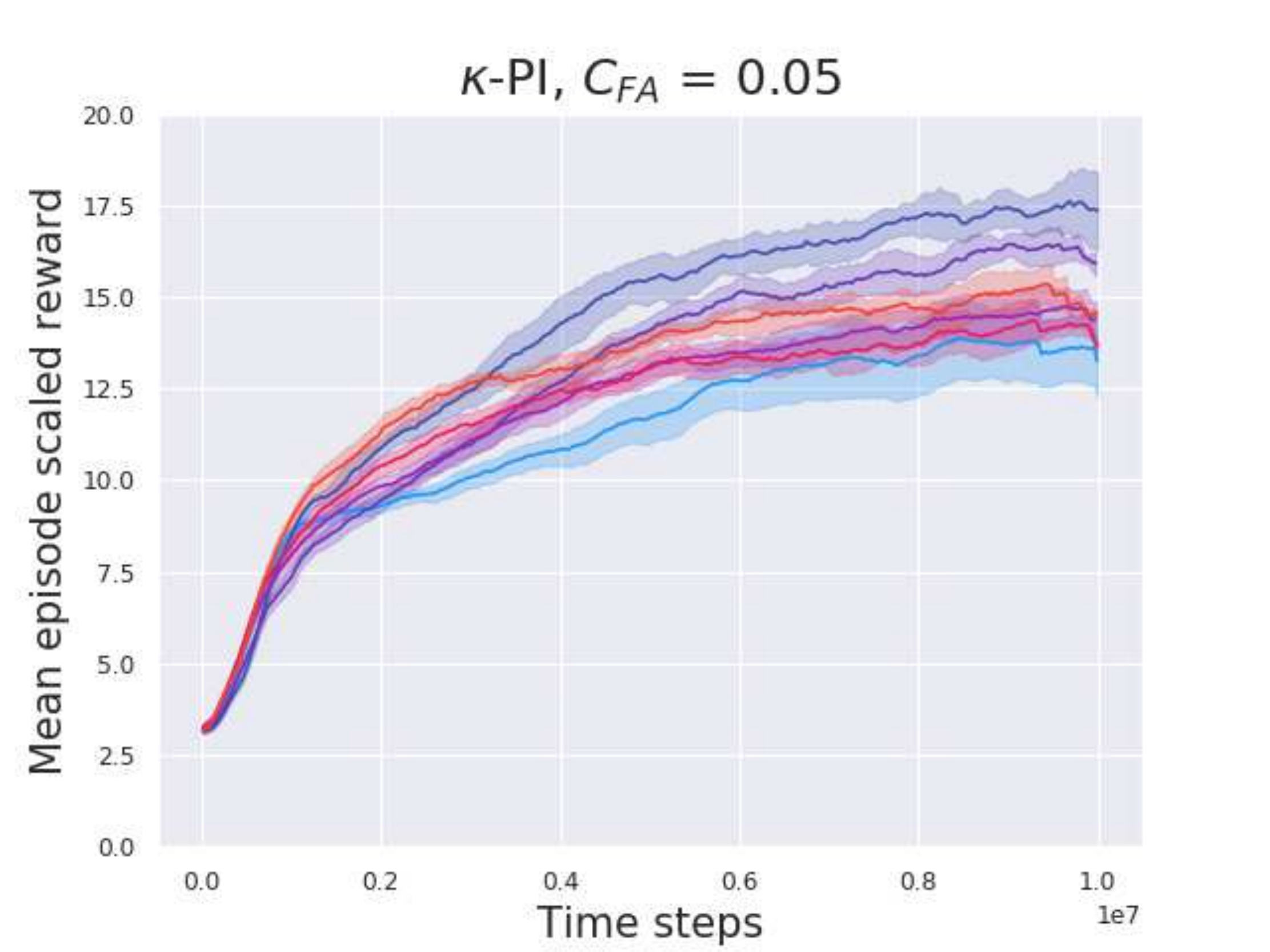}
    \includegraphics[scale=0.20]{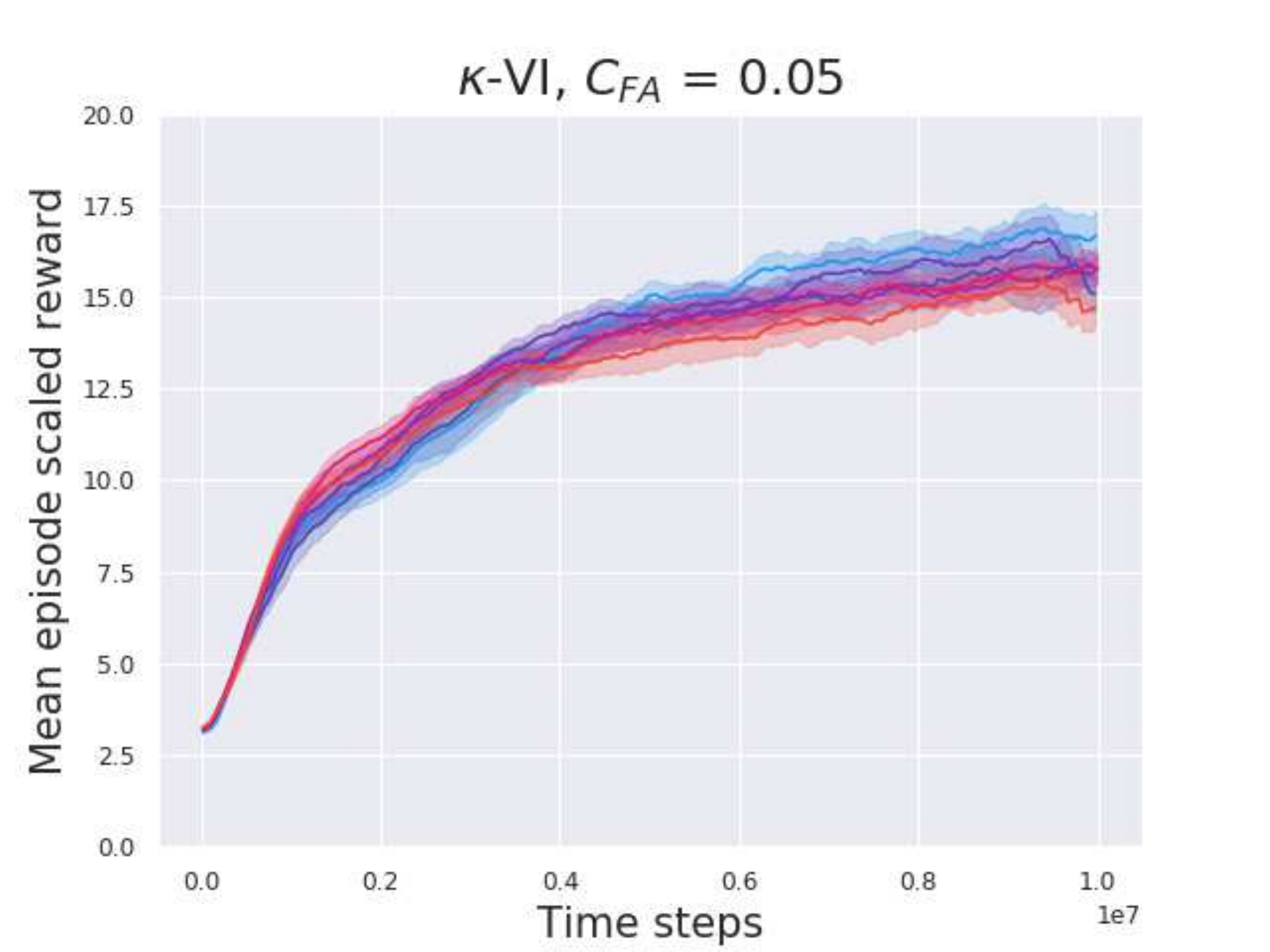}
    \includegraphics[scale=0.3]{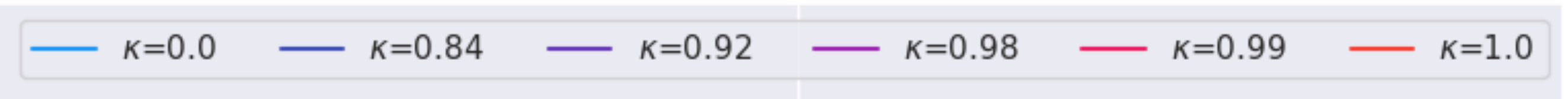}
    \caption{Training performance of the `naive' baseline $N_\kappa=T$ and $\kappa$-PI-DQN, $\kappa$-VI-DQN for $C_{\text{FA}}=0.05$ on SpaceInvaders}
    \label{fig:SpaceInvaders}
\end{figure*}

\begin{figure*}[ht]
    \centering
    \includegraphics[scale=0.20]{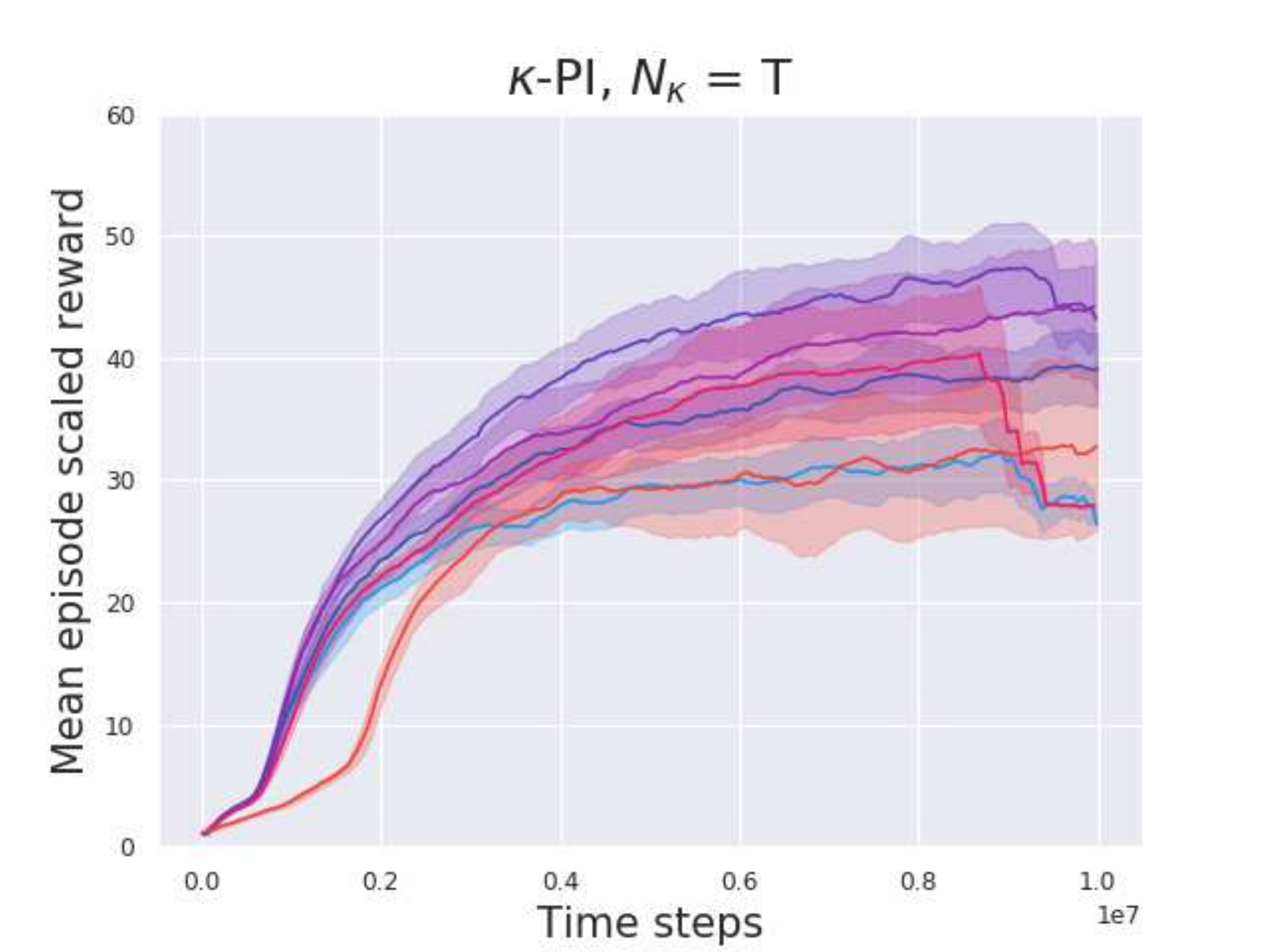}
    \includegraphics[scale=0.20]{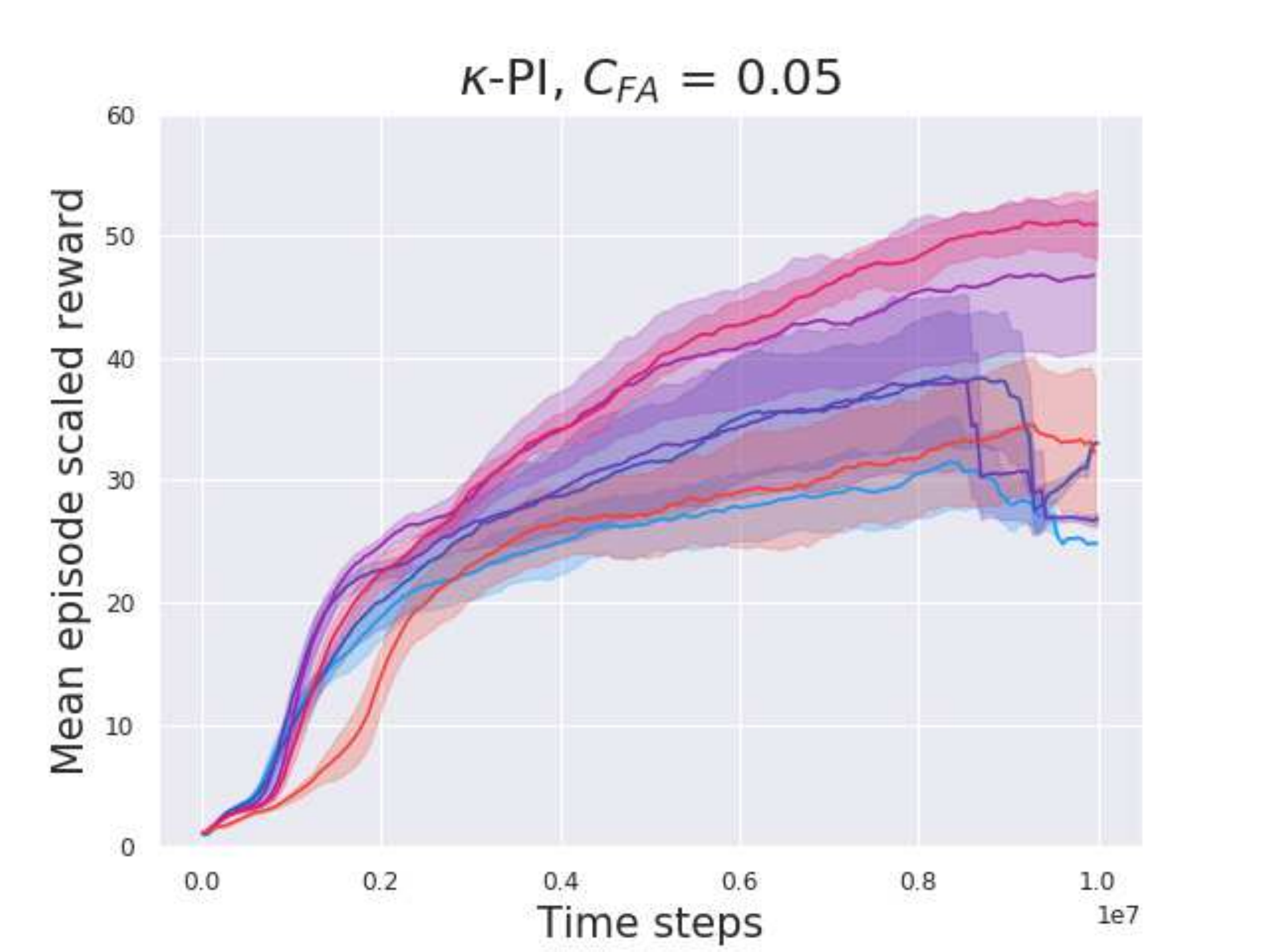}
    \includegraphics[scale=0.20]{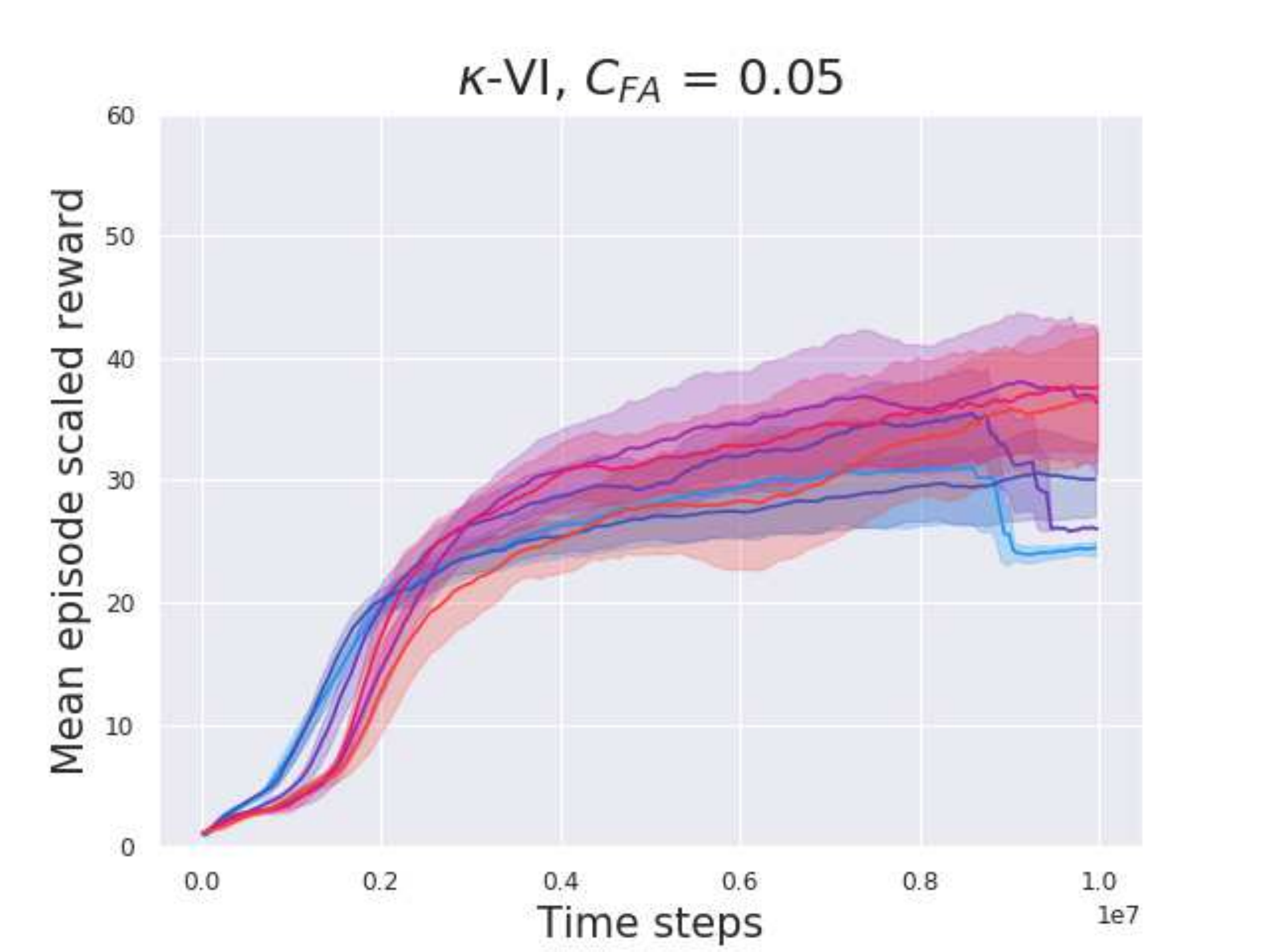}
    \includegraphics[scale=0.3]{Images/TRPO/legend_horizontal.pdf}
    \caption{Training performance of the `naive' baseline $N_\kappa=T$ and $\kappa$-PI-DQN, $\kappa$-VI-DQN for $C_{\text{FA}}=0.05$ on Seaquest}
    \label{fig:Seaquest}
\end{figure*}

\begin{figure*}[ht]
    \centering
    \includegraphics[scale=0.20]{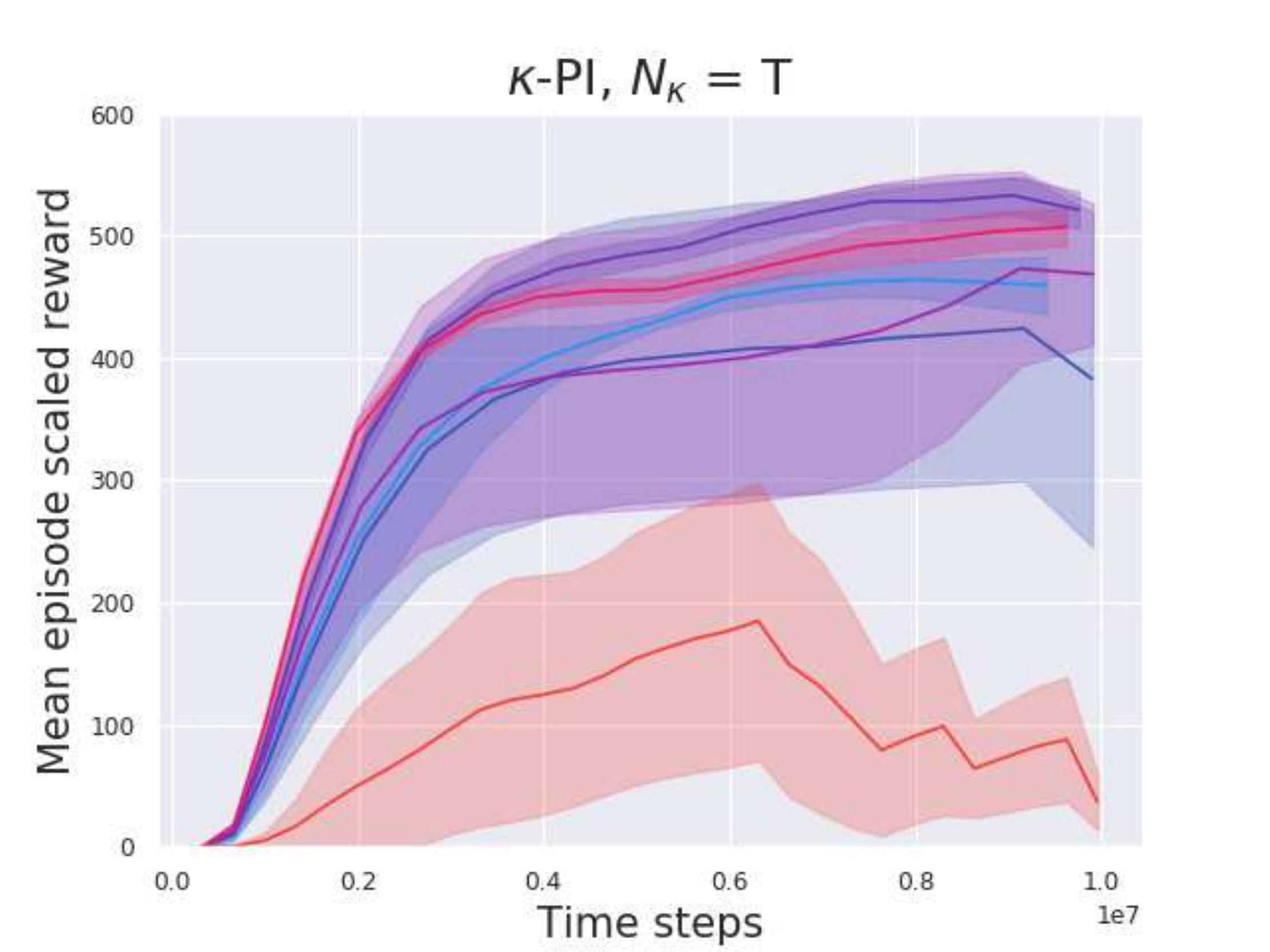}
    \includegraphics[scale=0.20]{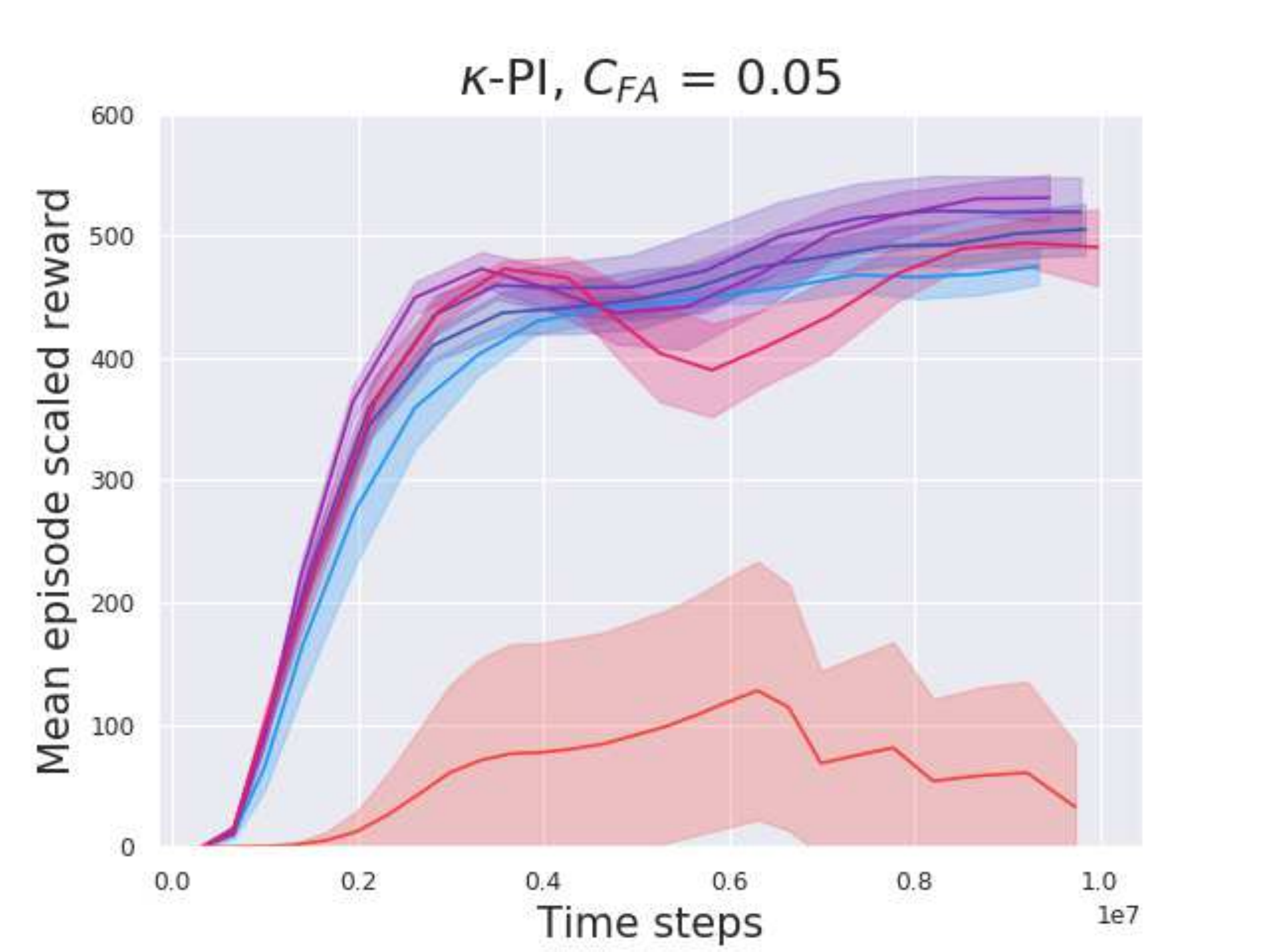}
    \includegraphics[scale=0.20]{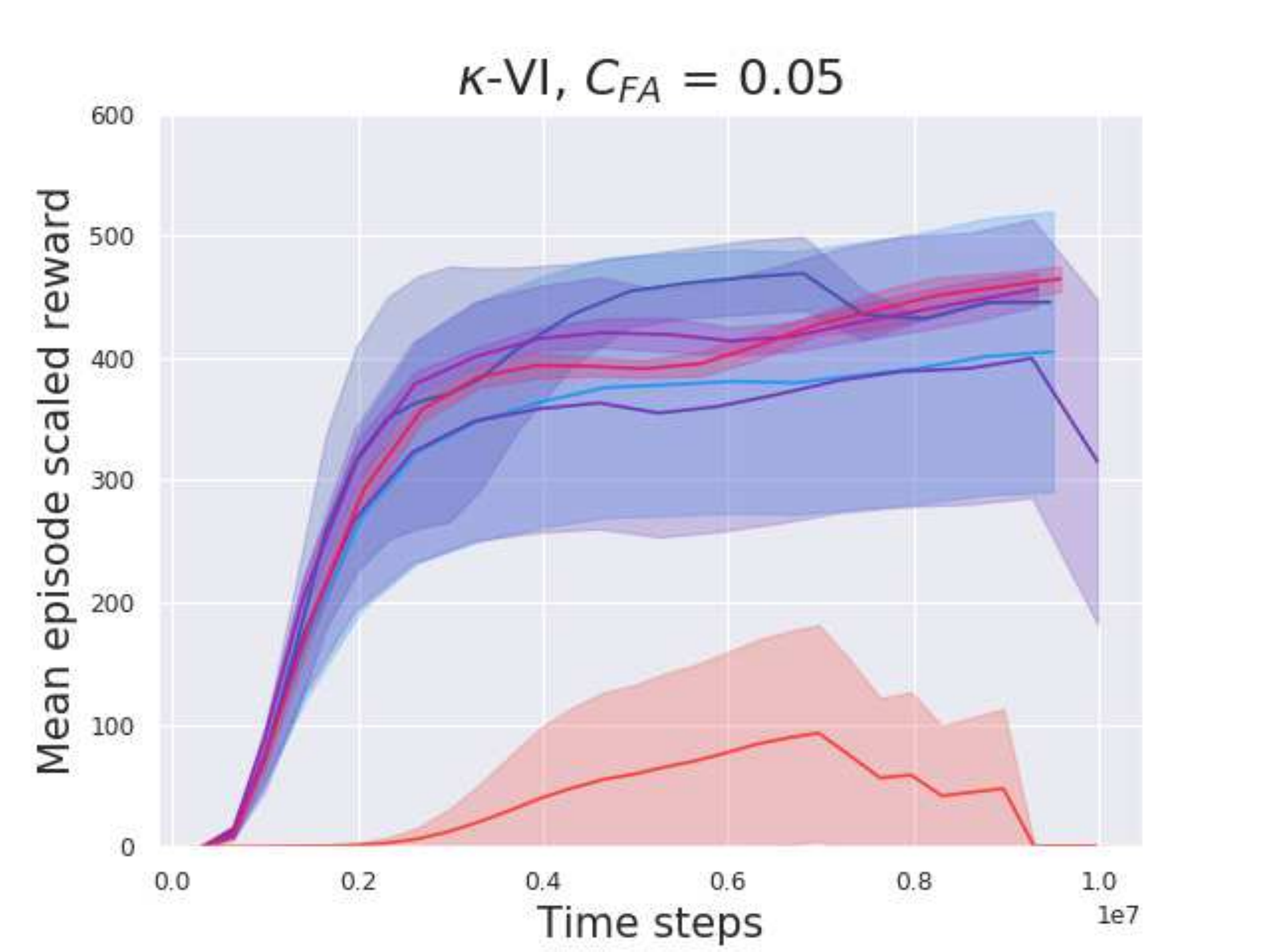}
    \includegraphics[scale=0.3]{Images/TRPO/legend_horizontal.pdf}
    \caption{Training performance of the `naive' baseline $N_\kappa=T$ and $\kappa$-PI-DQN, $\kappa$-VI-DQN for $C_{\text{FA}}=0.05$ on Enduro}
    \label{fig:Enduro}
\end{figure*}

\begin{figure*}[ht]
    \centering
    \includegraphics[scale=0.20]{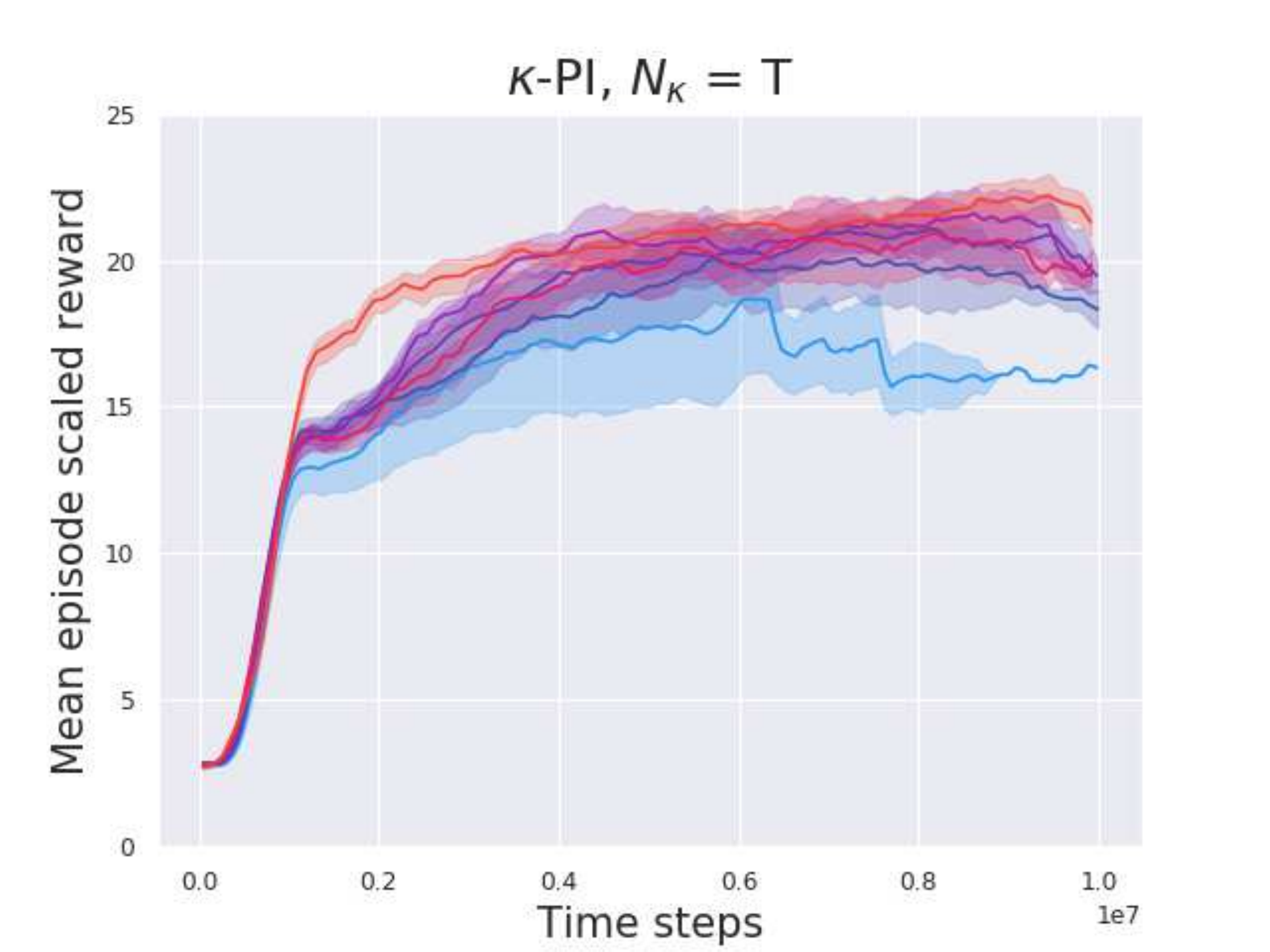}
    \includegraphics[scale=0.20]{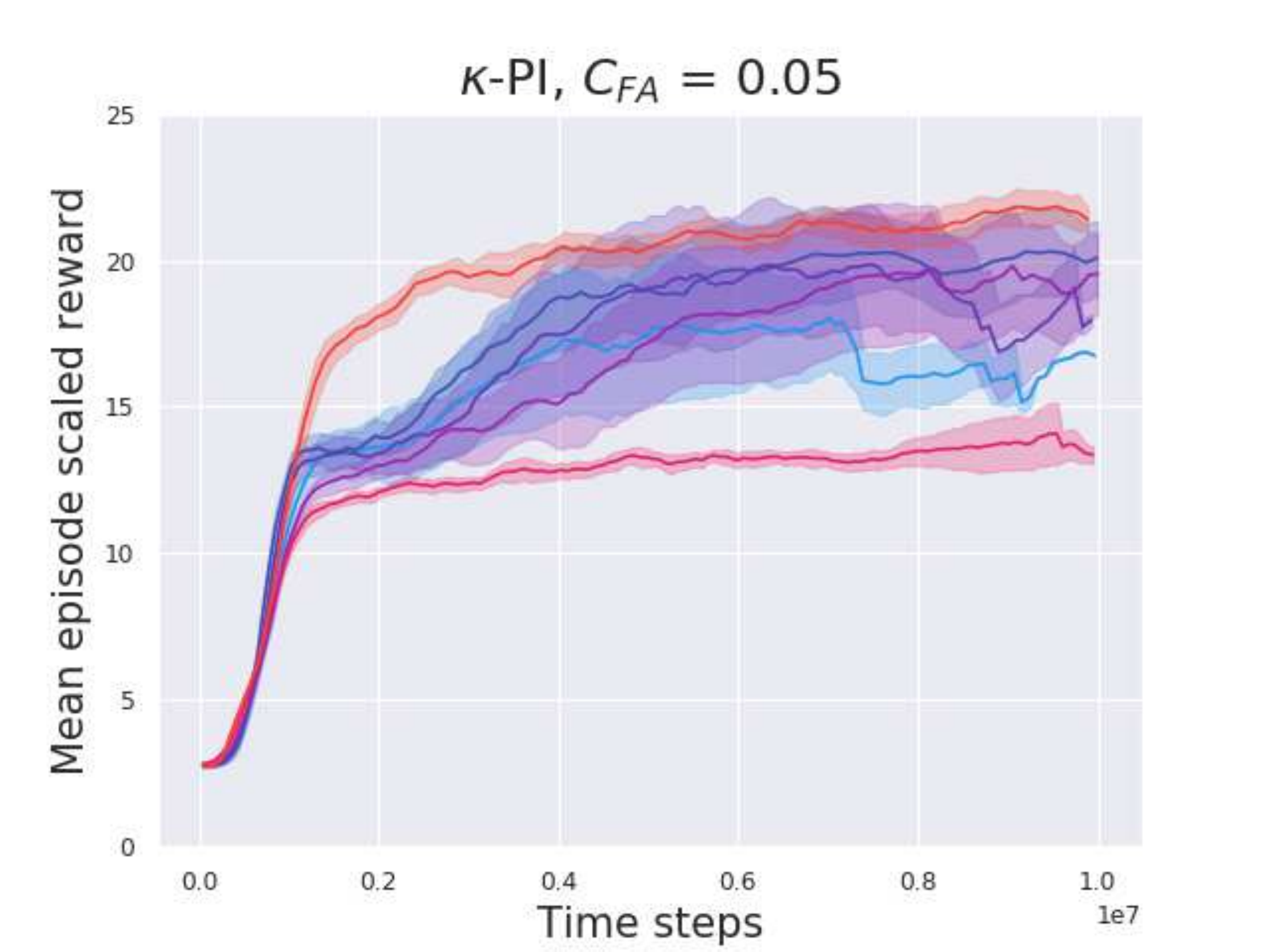}
    \includegraphics[scale=0.20]{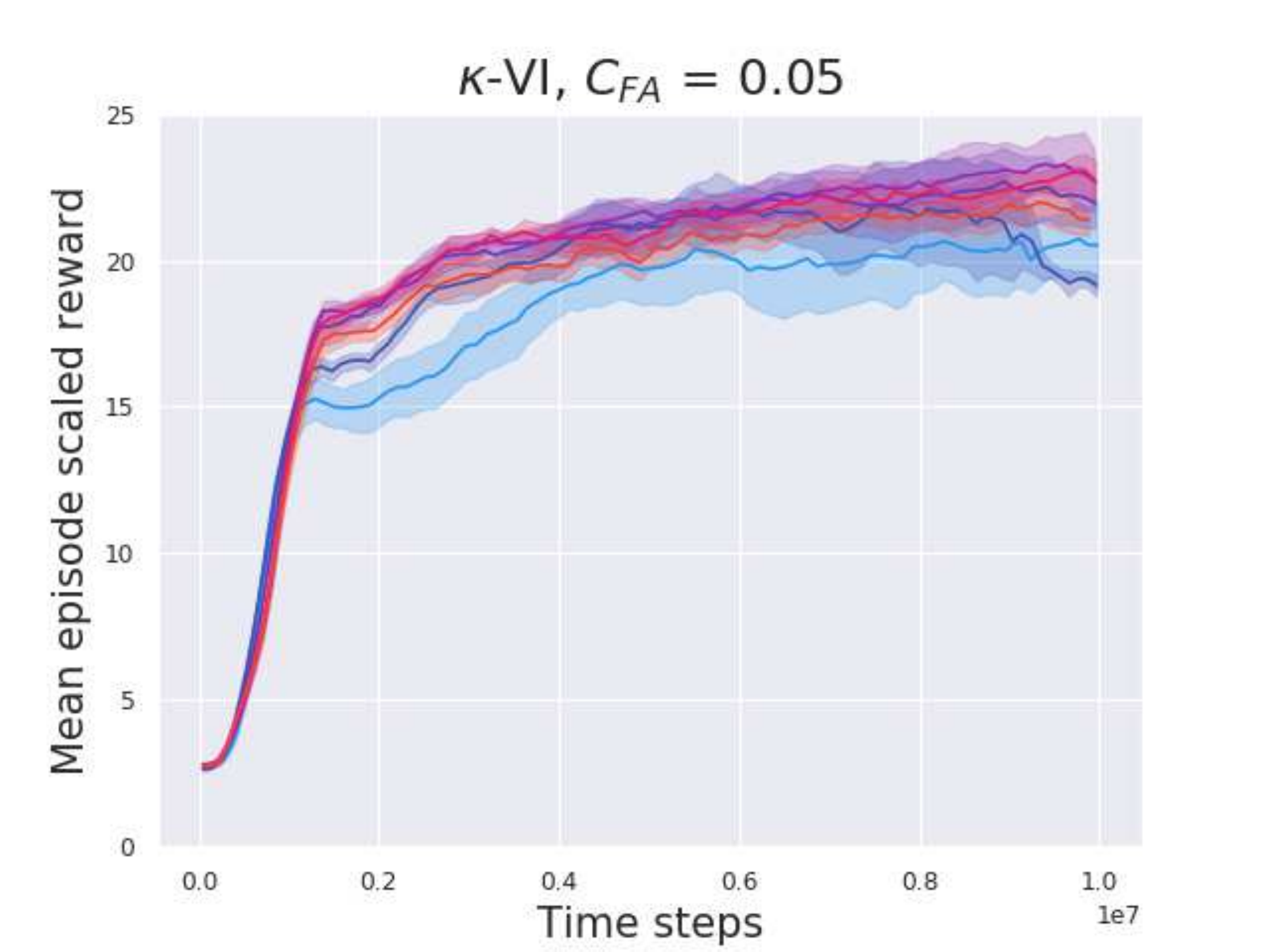}
    \includegraphics[scale=0.3]{Images/TRPO/legend_horizontal.pdf}
    \caption{Training performance of the `naive' baseline $N_\kappa=T$ and $\kappa$-PI-DQN, $\kappa$-VI-DQN for $C_{\text{FA}}=0.05$ on BeamRider}
    \label{fig:BeamRider}
\end{figure*}

\begin{figure*}[ht]
    \centering
    \includegraphics[scale=0.20]{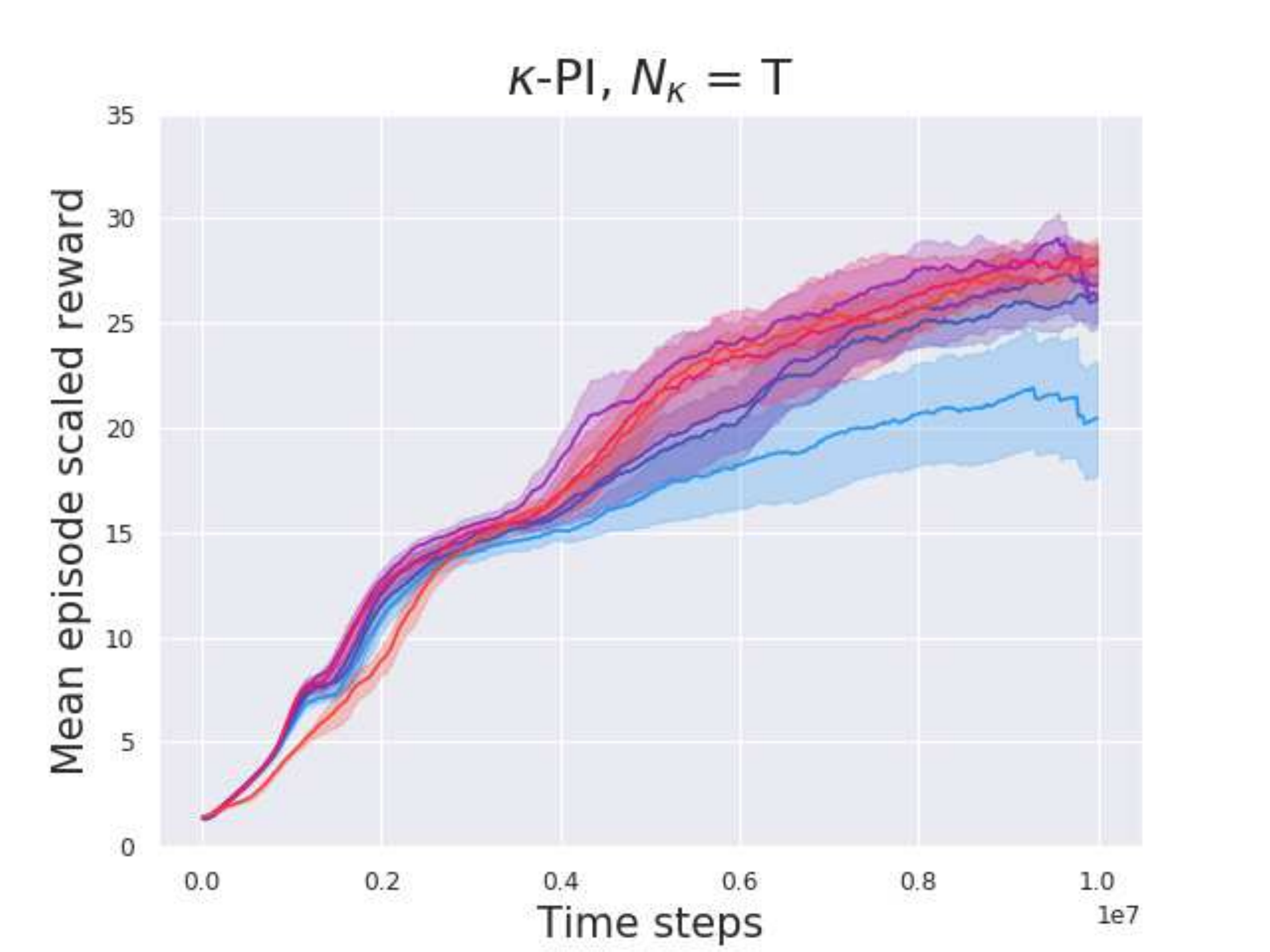}
    \includegraphics[scale=0.20]{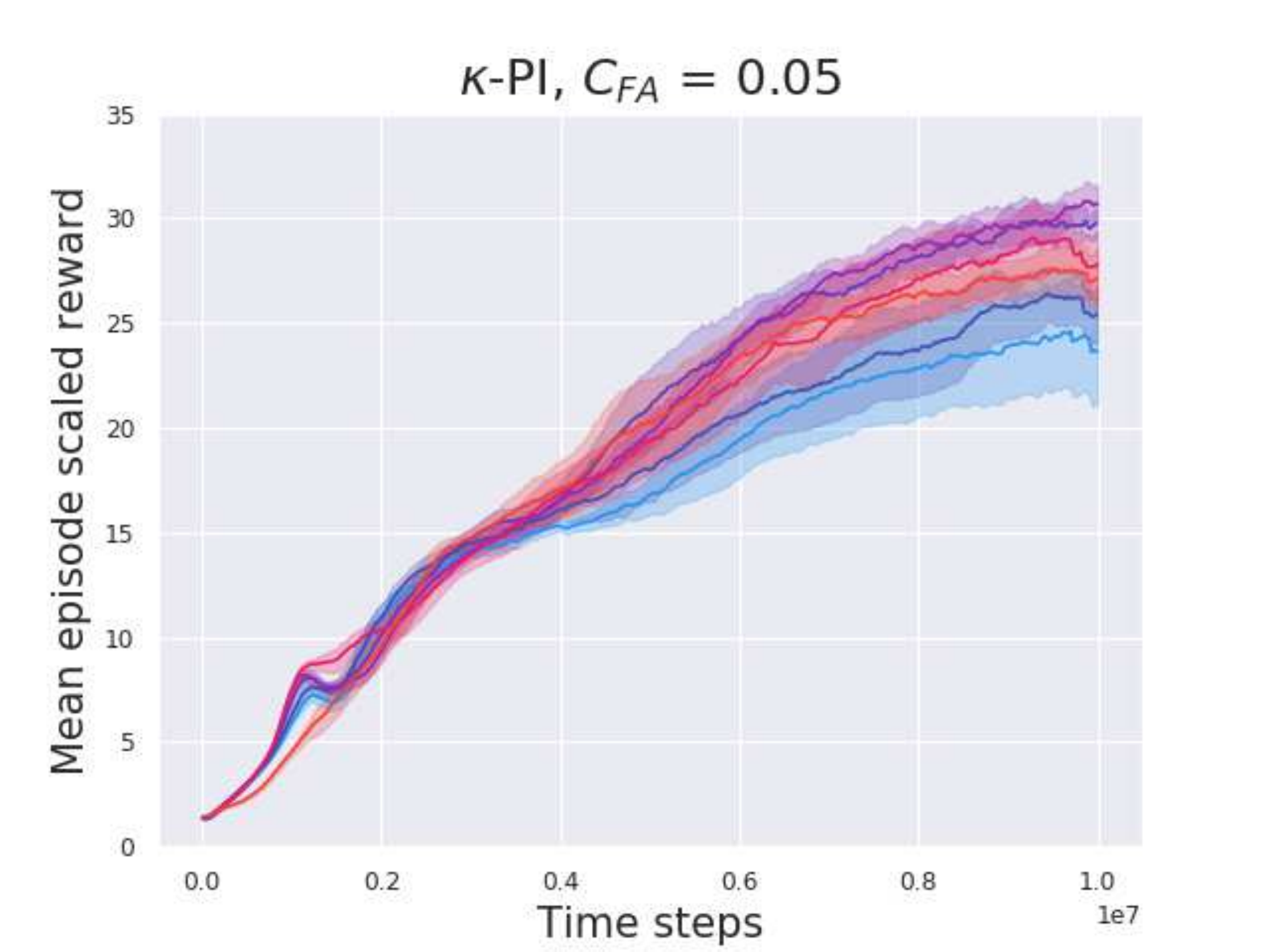}
    \includegraphics[scale=0.20]{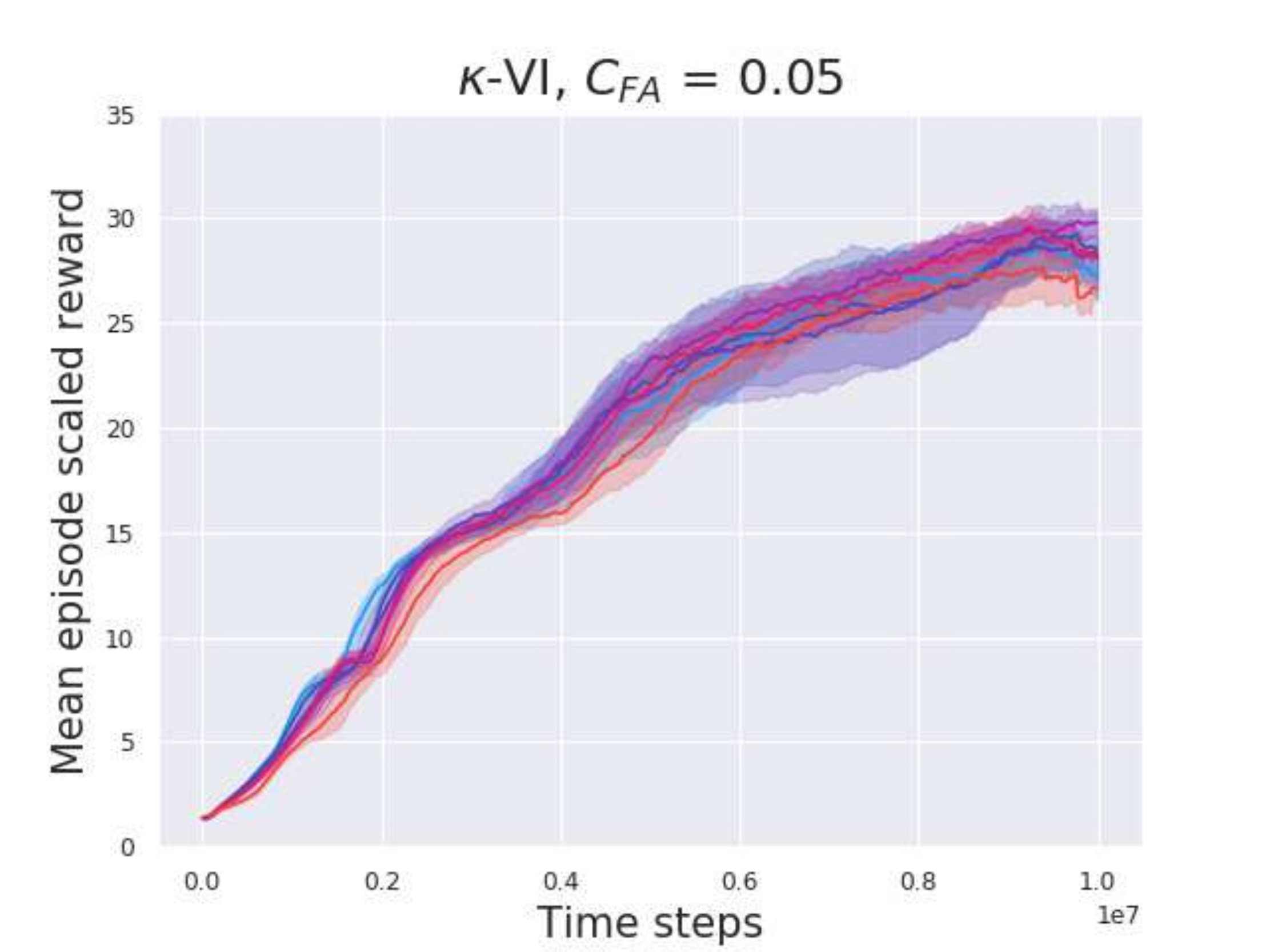}
    \includegraphics[scale=0.3]{Images/TRPO/legend_horizontal.pdf}
    \caption{Training performance of the `naive' baseline $N_\kappa=T$ and $\kappa$-PI-DQN, $\kappa$-VI-DQN for $C_{\text{FA}}=0.05$ on Qbert}
    \label{fig:Qbert}
\end{figure*}

\newpage
\newpage

\clearpage
\section{$\kappa$-PI-TRPO and $\kappa$-VI-TRPO Algorithms}
\label{supp: TRPO}

\subsection{Detailed Pseudo-codes}
\label{supp: pseudo code k-VI k-PI TRPO}

In this section, we report the detailed pseudo-codes of the $\kappa$-PI-TRPO and $\kappa$-VI-TRPO algorithms, described in Section 4.4, side-by-side. 

\begin{algorithm}
    \caption{$\kappa$-PI-TRPO}
    \label{alg:kappaPI-TRPO-Full}
    \begin{algorithmic}[1]
        \STATE {\bfseries Initialize} $V$-networks $V_{\theta}$ and $V_{\phi}$ with random weights $\theta$ and $\phi$; $\;$ policy network $\pi_\psi$ with random weights $\psi$; 
        \FOR{ $i = 0, \ldots, N_\kappa-1$}
            \FOR{$t =1,\ldots,T_\kappa$}
                \STATE Simulate the current policy $\pi_\psi$ for $M$ time-steps; 
                \FOR{$j=1,\ldots,M$}
                    \STATE Calculate $\;R_j(\kappa,V_\phi) = \sum_{t=j}^M (\gamma \kappa)^{t-j} r_t(\kappa,V_\phi)\quad$ and $\quad\rho_j = \sum_{t=j}^M \gamma^{t-j} r_t$;
                \ENDFOR
                \STATE Sample a random mini-batch $\{(s_j, a_j, r_j, s_{j+1})\}_{j=1}^N$ from the simulated $M$ time-steps;
                \STATE Update $\theta$ by minimizing the loss function: $\quad\mathcal{L}_{V_\theta}$ = $\frac{1}{N} \sum_{j=1}^{N}(V_\theta(s_j) - R_{j}(\kappa,V_\phi))^2$;
                \STATE {\color{gray}\# Policy Improvement}
                \STATE Sample a random mini-batch $\{(s_j, a_j, r_j, s_{j+1})\}_{j=1}^N$ from the simulated $M$ time-steps;
                \STATE Update $\psi$ using TRPO with advantage function computed by $\{(R_j(\kappa,V_\phi),V_\theta(s_j))\}_{j=1}^N$;
            \ENDFOR
            \STATE {\color{gray}\# Policy Evaluation}
            \STATE Sample a random mini-batch $\{(s_j, a_j, r_j, s_{j+1})\}_{j=1}^N$ from the simulated $M$ time-steps;
            \STATE Update $\phi$ by minimizing the loss function: $\quad\mathcal{L}_{V_{\phi}}=\frac{1}{N} \sum_{j=1}^N(V_\phi(s_j) - \rho_j)^2$;
        \ENDFOR
    \end{algorithmic}
\end{algorithm}

\begin{algorithm}
    \caption{$\kappa$-VI-TRPO}
    \label{alg:kappaVI-TRPO-Full}
    \begin{algorithmic}[1]
        \STATE {\bfseries Initialize} $V$-networks $V_\theta$ and $V_\phi$ with random weights $\theta$ and $\phi$; $\;$ policy network $\pi_\psi$ with random weights $\psi$;
        \FOR{ $i = 0, \ldots, N_\kappa-1$}
            \STATE {\color{gray}\# Evaluate $T_\kappa V_\phi$ and the $\kappa$-greedy policy w.r.t.~$V_\phi$}
            \FOR{$t=1,\ldots,T_\kappa$}
                \STATE Simulate the current policy $\pi_\psi$ for $M$ time-steps;
                \FOR{$j=1,\ldots,M$}
                    \STATE Calculate $R_j(\kappa,V_\phi) = \sum_{t=j}^M (\gamma \kappa)^{t-j}r_t(\kappa,V_\phi)$;
                \ENDFOR    
                \STATE Sample a random mini-batch $\{(s_j,a_j,r_j,s_{j+1})\}_{j=1}^N$ from the simulated $M$ time-steps;
                \STATE Update $\theta$ by minimizing the loss function: $\quad\mathcal{L}_{V_\theta}=\frac{1}{N}\sum_{j=1}^N(V_\theta(s_j) - R_{j}(\kappa,V_\phi))^2$;
                \STATE Sample a random mini-batch $\{(s_j,a_j,r_j,s_{j+1})\}_{j=1}^N$ from the simulated $M$ time-steps;
                \STATE Update $\psi$ using TRPO with advantage function computed by $\{(R_j(\kappa,V_\phi),V_\theta(s_j))\}_{j=1}^N$;
            \ENDFOR
            \STATE Copy $\theta$ to $\phi\quad(\phi \leftarrow \theta)$;
        \ENDFOR
    \end{algorithmic}
\end{algorithm}

\newpage

\subsection{Ablation Test for $C_{FA}$}
\label{subsec: TRPO appendix CFA}

\begin{figure*}[ht]
    \centering
    \includegraphics[scale=0.20]{Images/TRPO/K-PI/Walker_KPI_nkappa_x7.pdf}
    \includegraphics[scale=0.20]{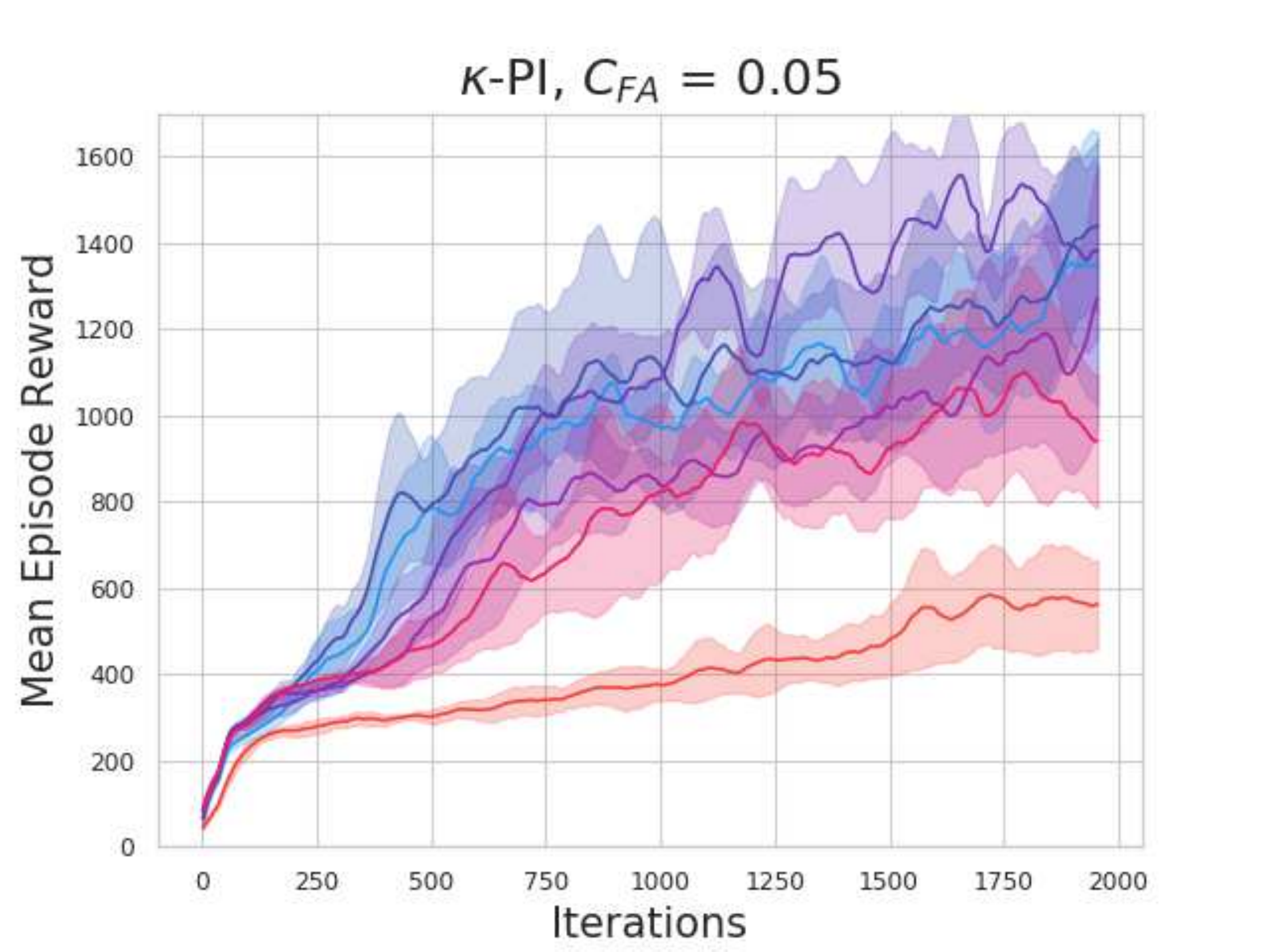}
    \includegraphics[scale=0.20]{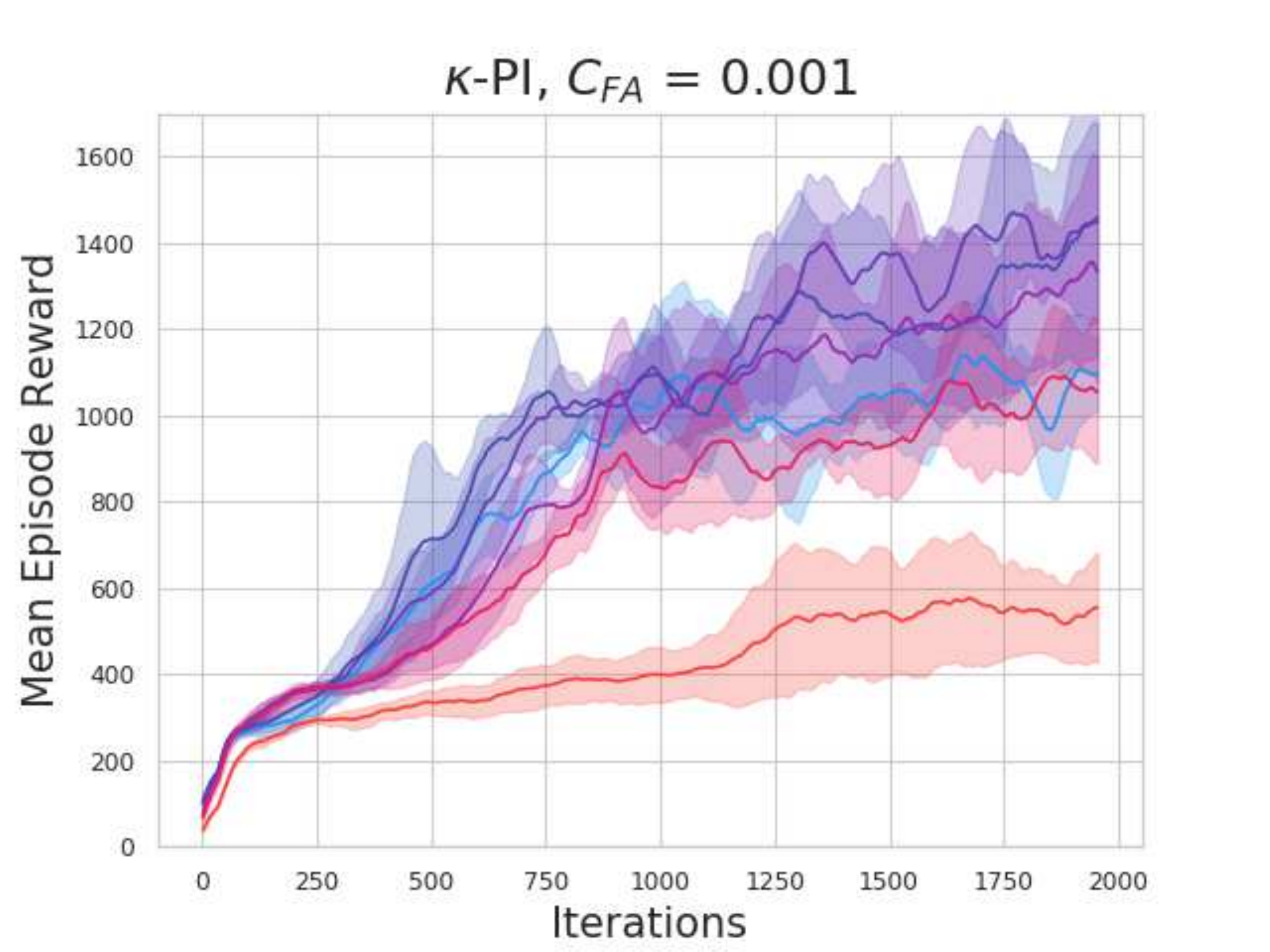}
    \centering
    \includegraphics[scale=0.3]{Images/TRPO/legend_horizontal.pdf}
    \caption{Performance of $\kappa$-PI-TRPO and $\kappa$-VI-TRPO on Walker2d-v2 for different values of $C_{FA}$.}
    \label{fig:Walker}
\end{figure*}

\begin{table*}
\small
\begin{center}
\begin{tabular}{ c | c } 
    Hyperparameter & Value \\
    \hline 
    Horizon (T) & 1000 \\
    Adam stepsize & $1 \times 10^{-3}$ \\
    Number of samples per Iteration & 1024 \\
    Entropy coefficient & 0.01 \\
    Discount factor & 0.99 \\
    Number of Iterations & 2000 \\
    Minibatch size & 128 \\
    \#Runs used for plot averages & 10 \\
    Confidence interval for plot runs & $\sim$ 95\%
\end{tabular}
\end{center}
\caption{Hyper-parameters of $\kappa$-PI-TRPO and $\kappa$-VI-TRPO on the MuJoCo domains.}
\label{tab:Hyperparams-TRPO}
\end{table*}

\subsection{$\kappa$-PI-TRPO and $\kappa$-VI-TRPO Plots}
\label{subsec:add-plot-TRPO}

In this section, we report additional results of the application of $\kappa$-PI-TRPO and $\kappa$-VI-TRPO on the MuJoCo domains. A summary of these results has been reported in Table 2 in the main paper. 

\begin{figure*}[ht]
    \centering
    \hspace{-0.6cm}\includegraphics[scale=0.20]{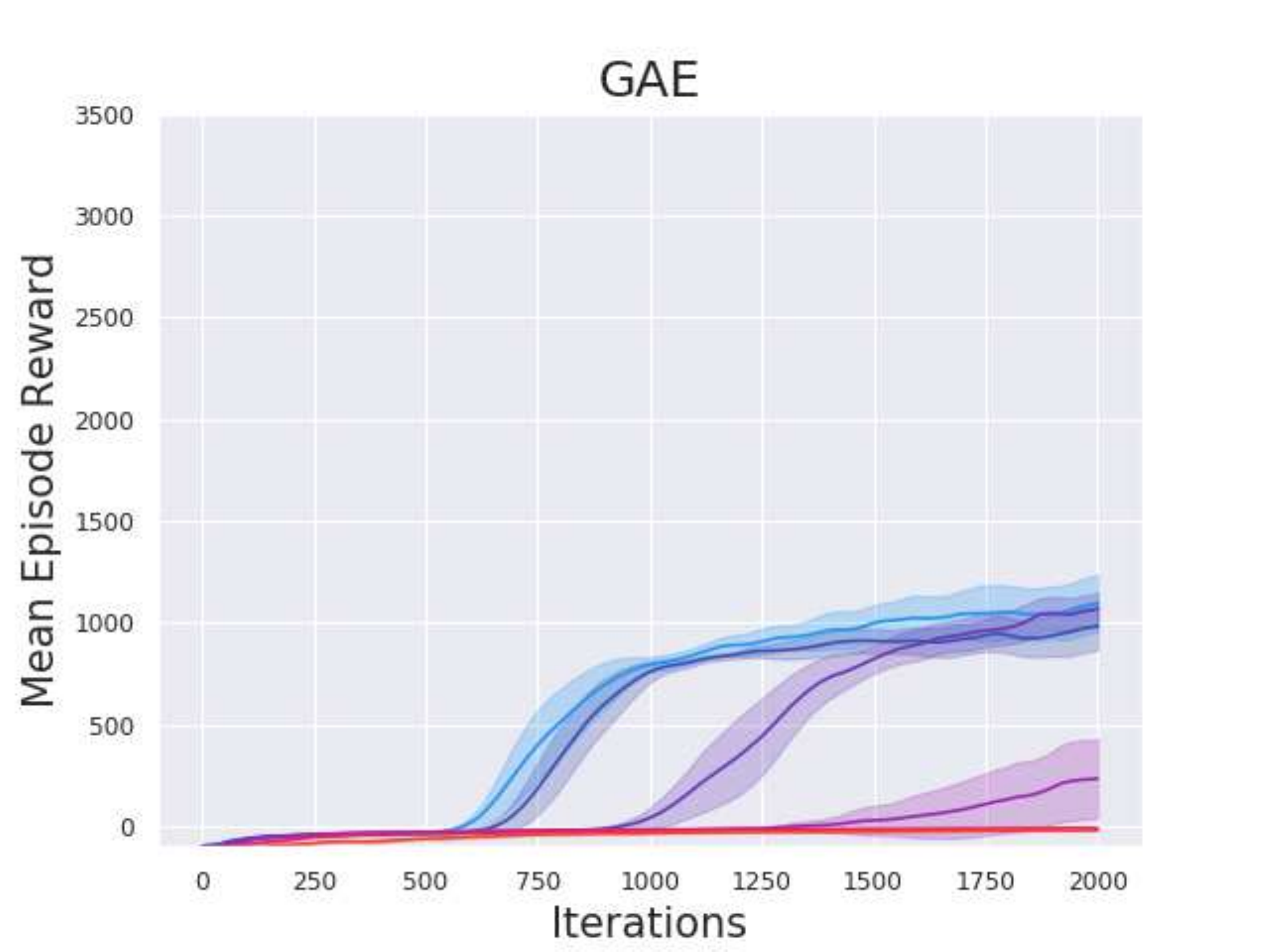}
    \hspace{-0.6cm} \includegraphics[scale=0.20]{Images/TRPO/K-PI/Ant_KPI_nkappa_1.pdf} 
    \hspace{-0.6cm}\includegraphics[scale=0.20]{Images/TRPO/K-PI/Ant_KPI_nkappa_x7.pdf}
    \hspace{-0.6cm}\includegraphics[scale=0.20]{Images/TRPO/K-VI/Ant_KVI_nkappa_x7.pdf} \\
    \centering
    \includegraphics[scale=0.3]{Images/TRPO/legend_horizontal.pdf}
    \caption{Performance of GAE, `Naive' baseline and $\kappa$-PI-TRPO, $\kappa$-VI-TRPO on Ant-v2.}
    \label{fig:Ant}
\end{figure*}

\begin{figure*}[ht]
    \centering
    \hspace{-0.6cm}\includegraphics[scale=0.20]{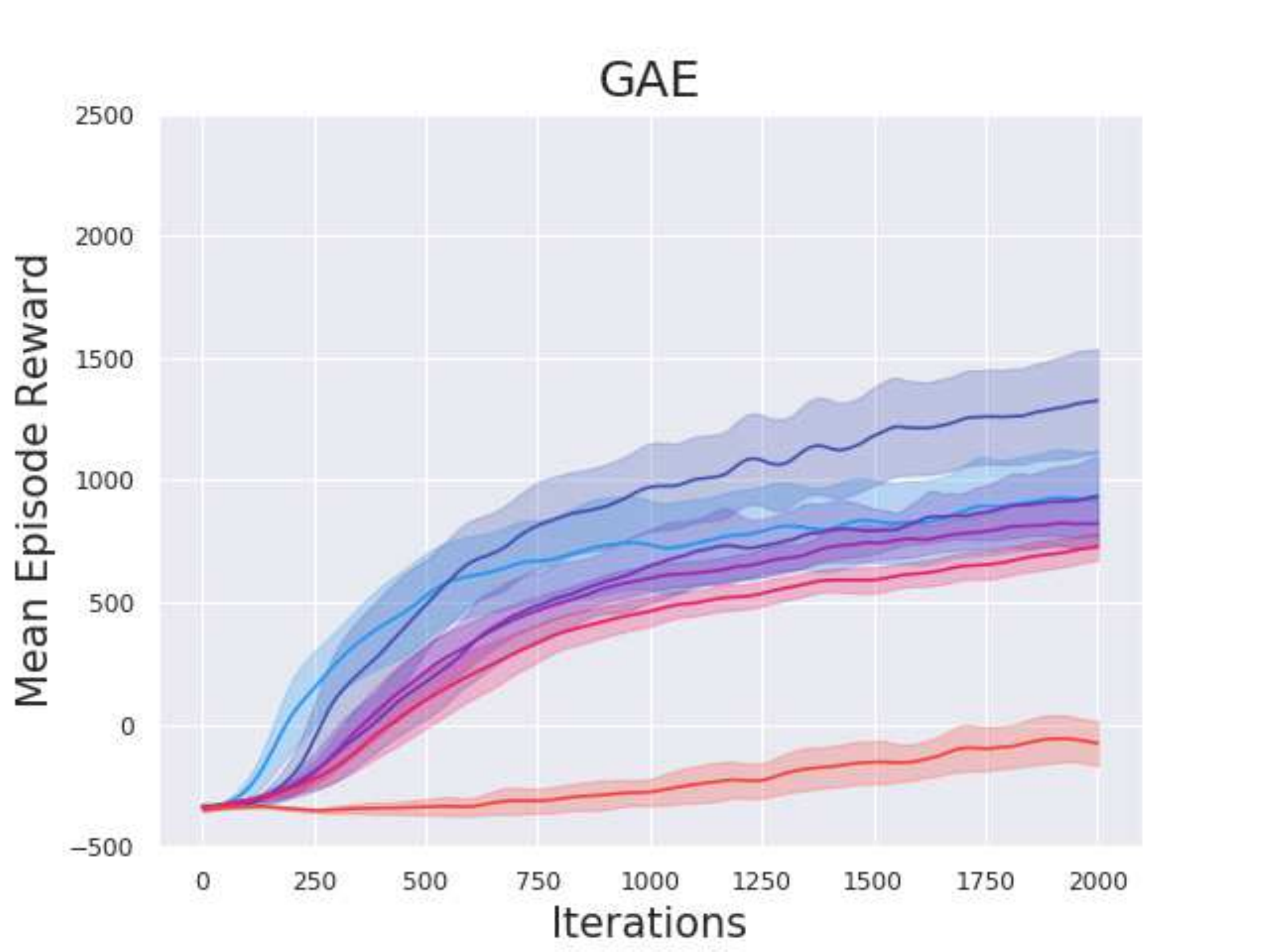}
    \hspace{-0.6cm}\includegraphics[scale=0.20]{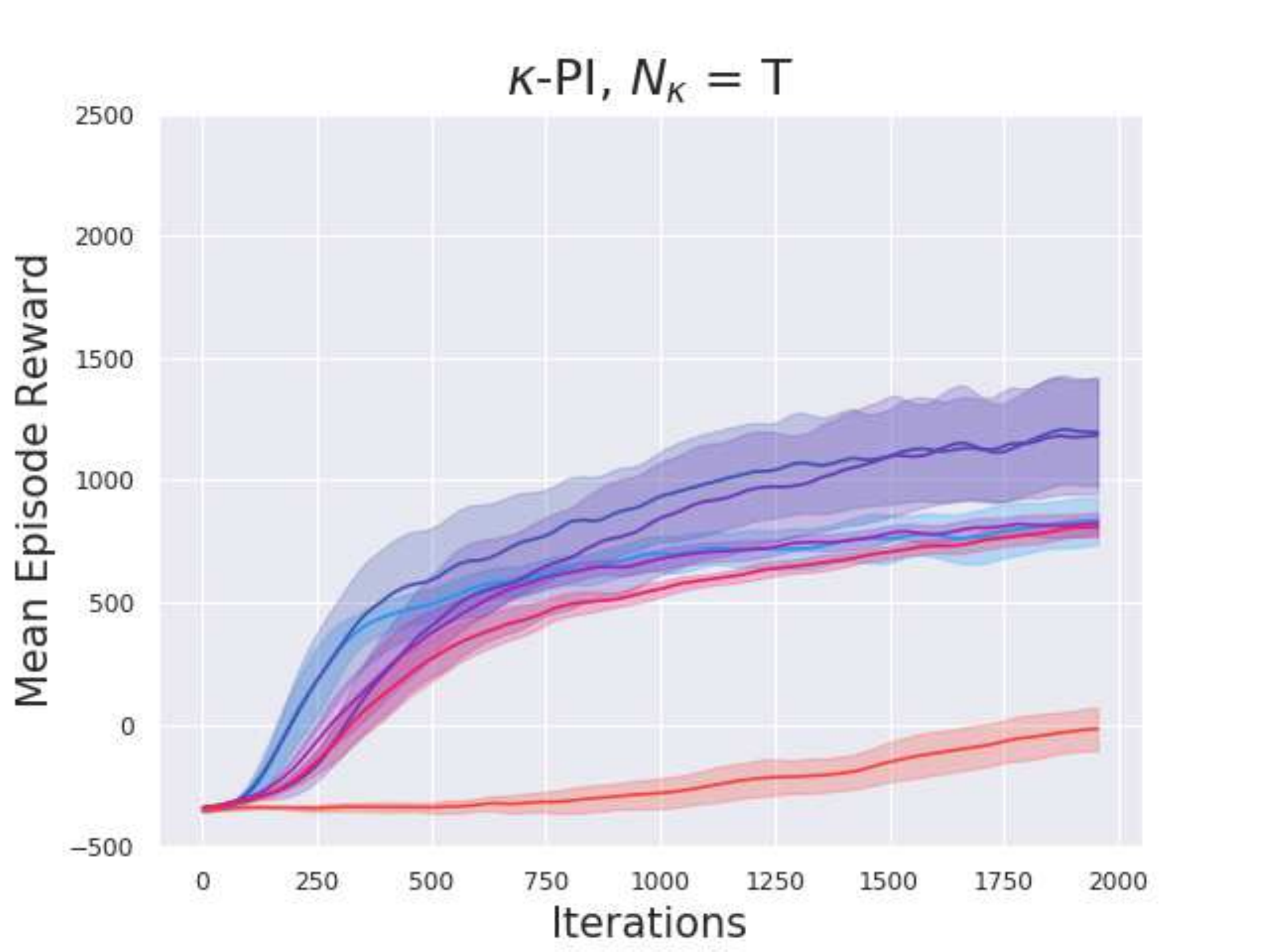} 
    \hspace{-0.6cm}\includegraphics[scale=0.20]{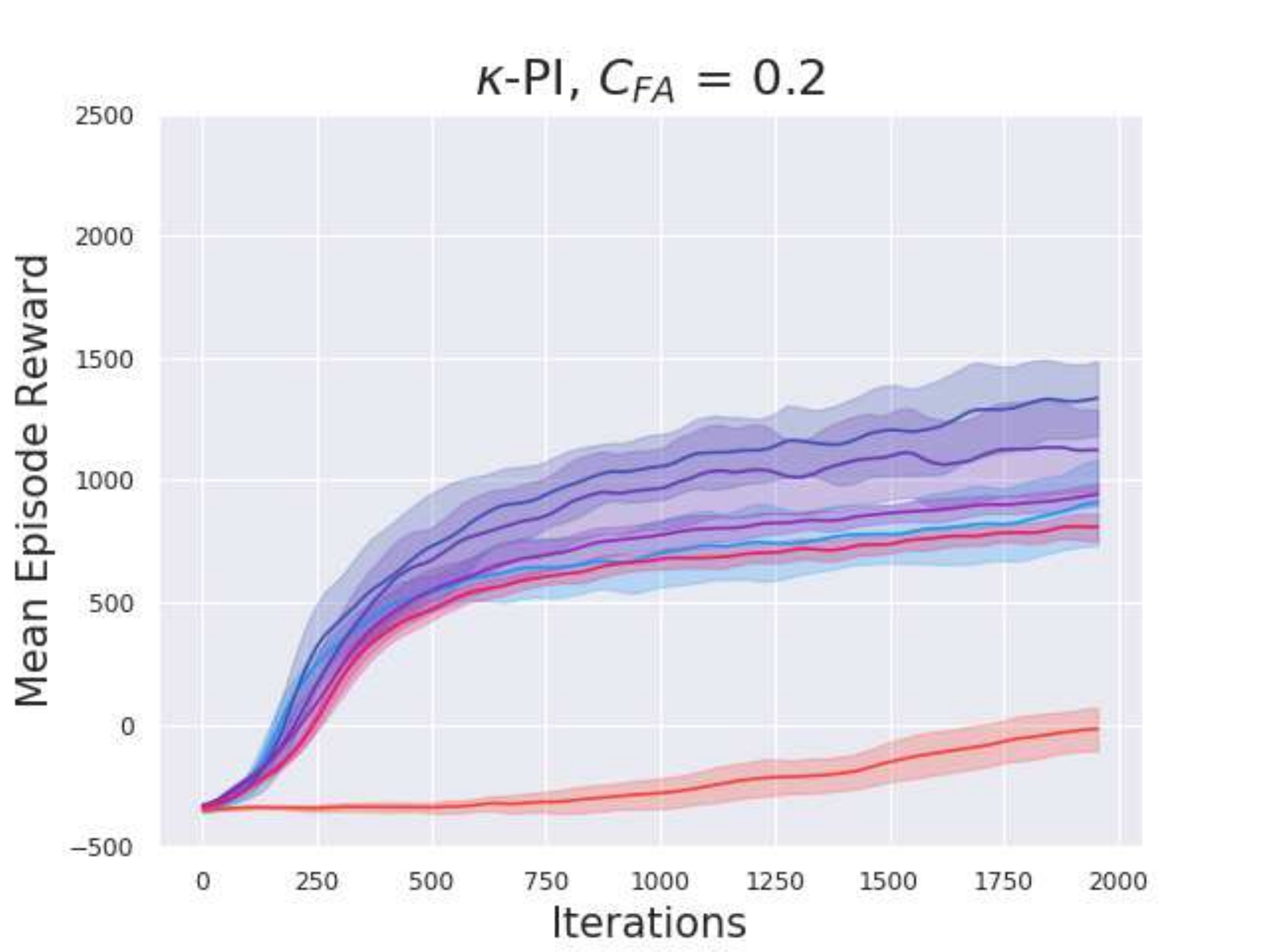}
    \hspace{-0.6cm}\includegraphics[scale=0.20]{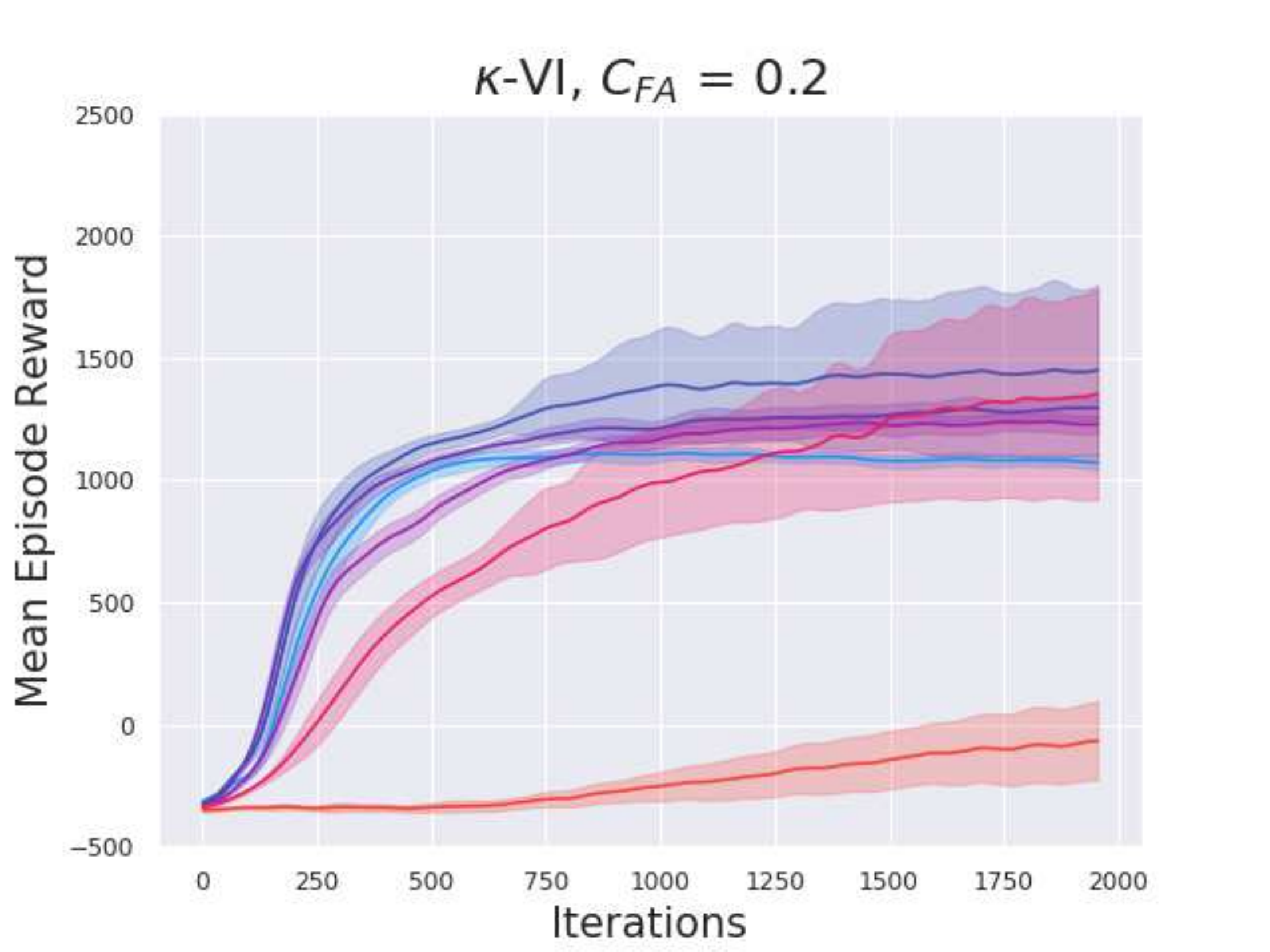} \\
    \centering
    \includegraphics[scale=0.3]{Images/TRPO/legend_horizontal.pdf}
    \caption{Performance of GAE, `Naive' baseline and $\kappa$-PI-TRPO, $\kappa$-VI-TRPO on HalfCheetah-v2.}
    \label{fig:HalfCheetah}
\end{figure*}

\begin{figure*}[ht]
    \centering
    \hspace{-0.6cm}\includegraphics[scale=0.20]{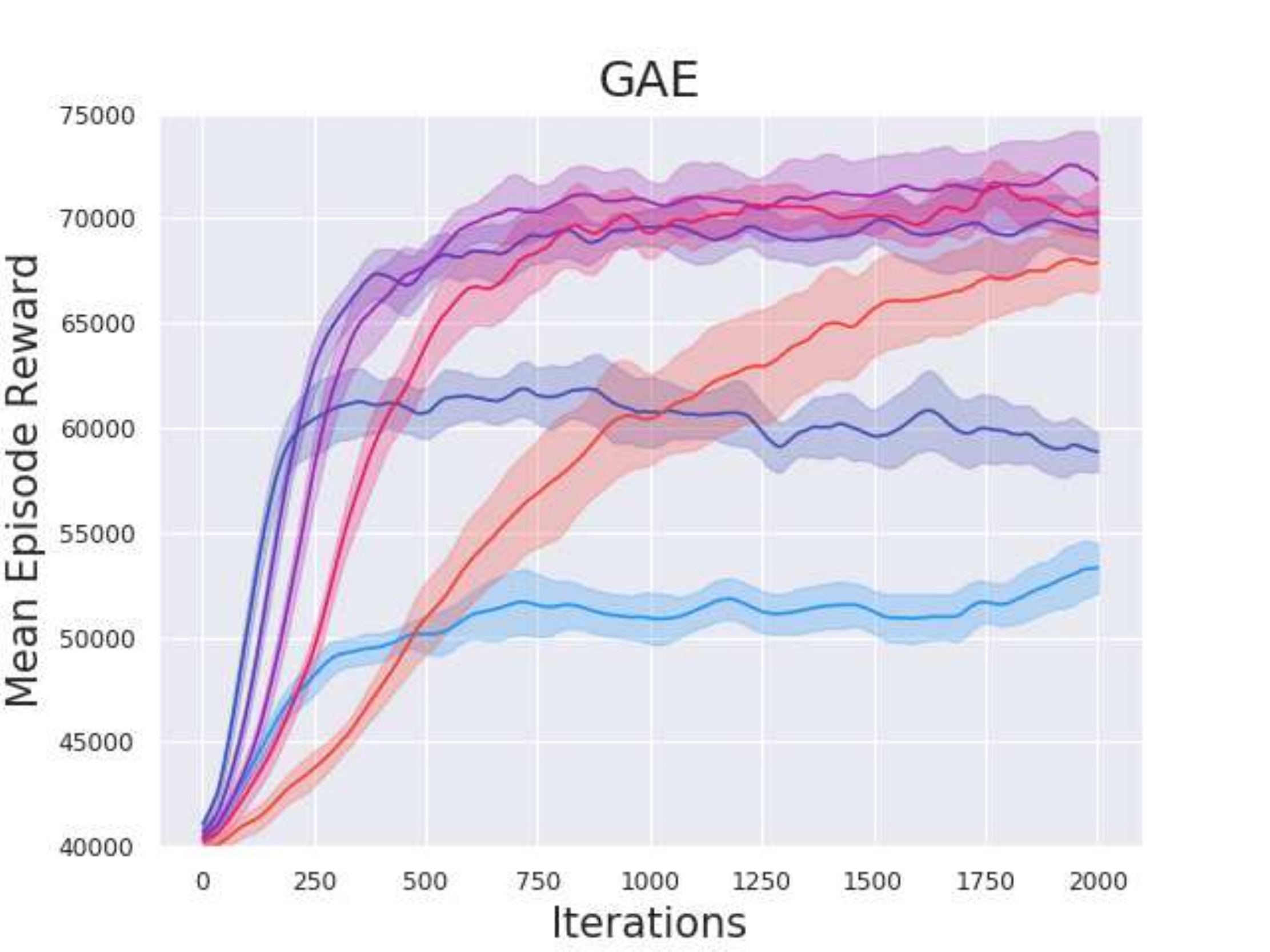}
    \hspace{-0.6cm}\includegraphics[scale=0.20]{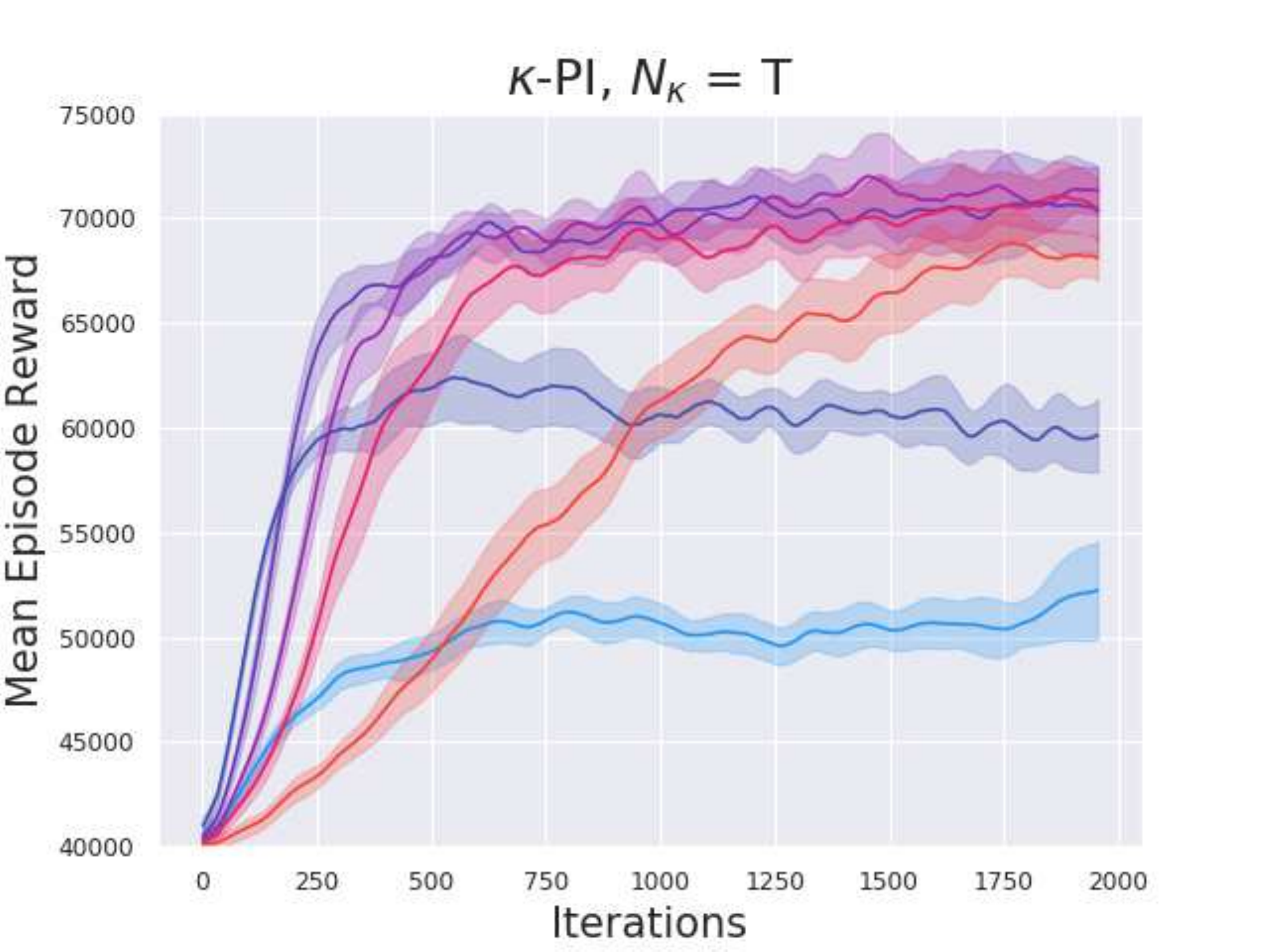} 
    \hspace{-0.6cm}\includegraphics[scale=0.20]{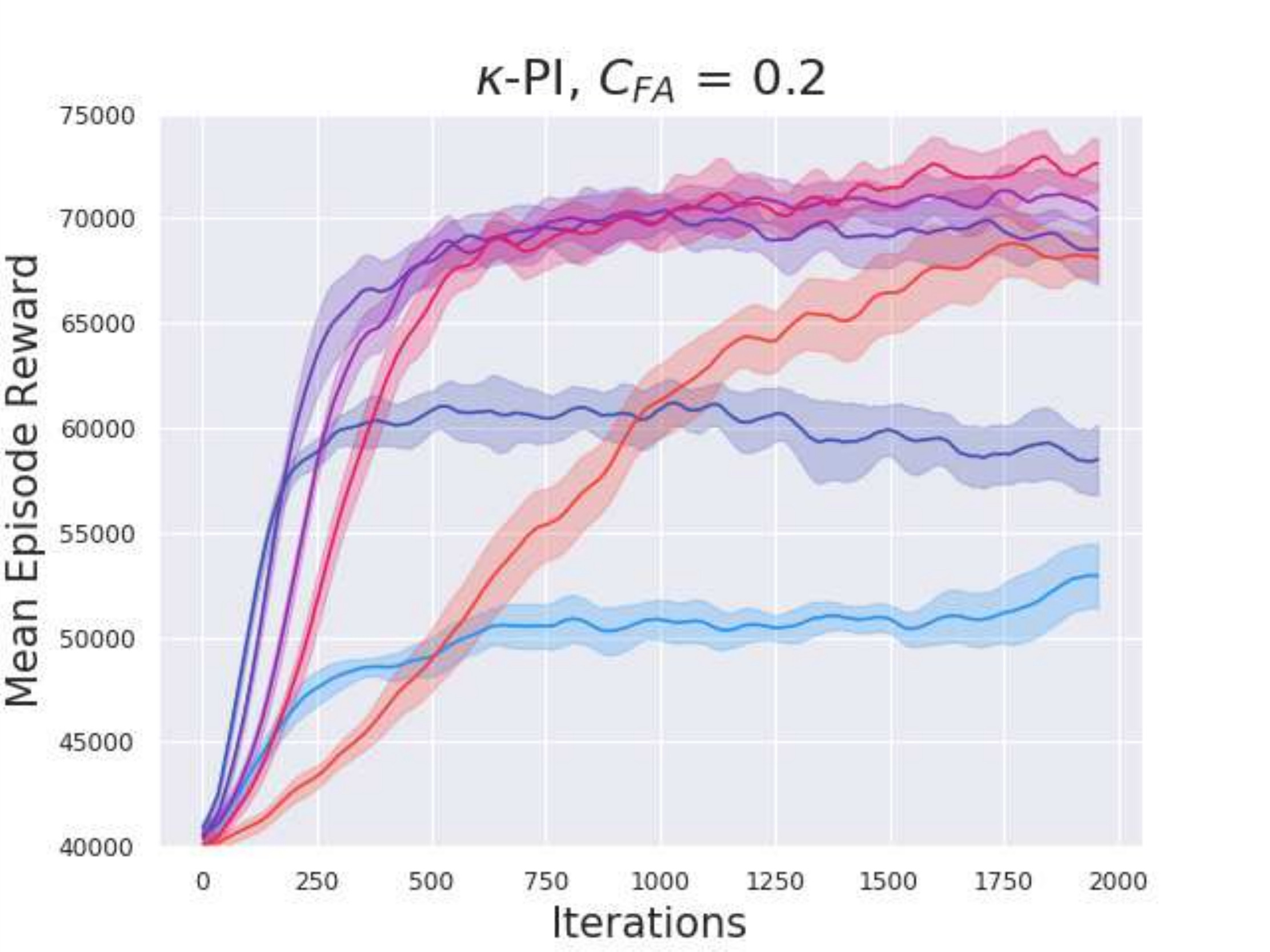}
    \hspace{-0.6cm}\includegraphics[scale=0.20]{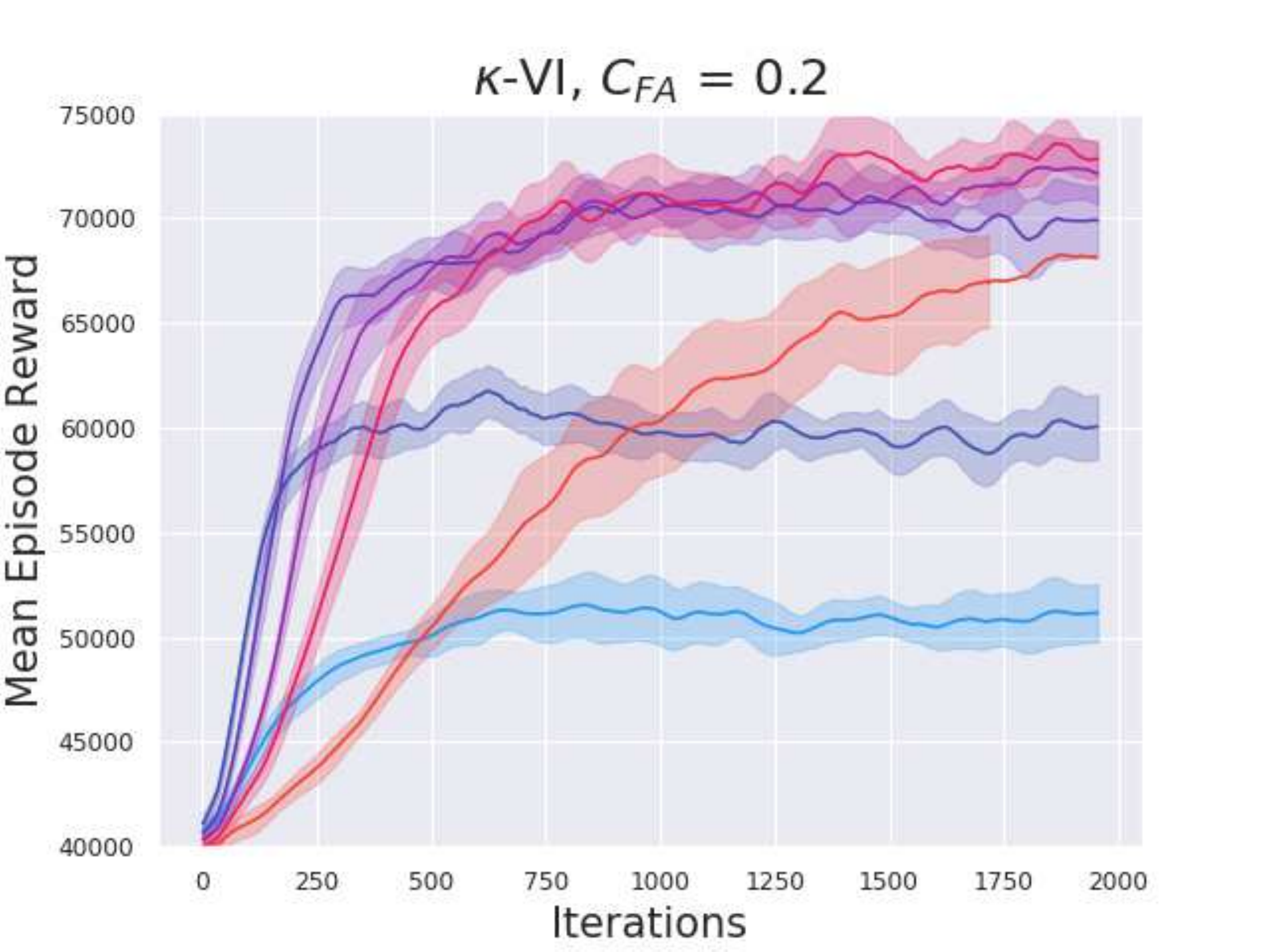} \\
    \centering
    \includegraphics[scale=0.3]{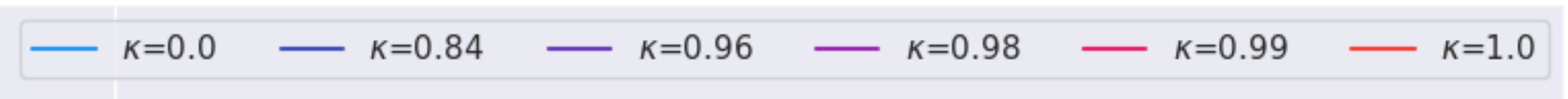}
    \caption{Performance of GAE, `Naive' baseline and $\kappa$-PI-TRPO, $\kappa$-VI-TRPO on HumanoidStandup-v2.}
    \label{fig:HumanoidStandup}
\end{figure*}

\begin{figure*}[ht]
    \centering
    \hspace{-0.6cm}\includegraphics[scale=0.20]{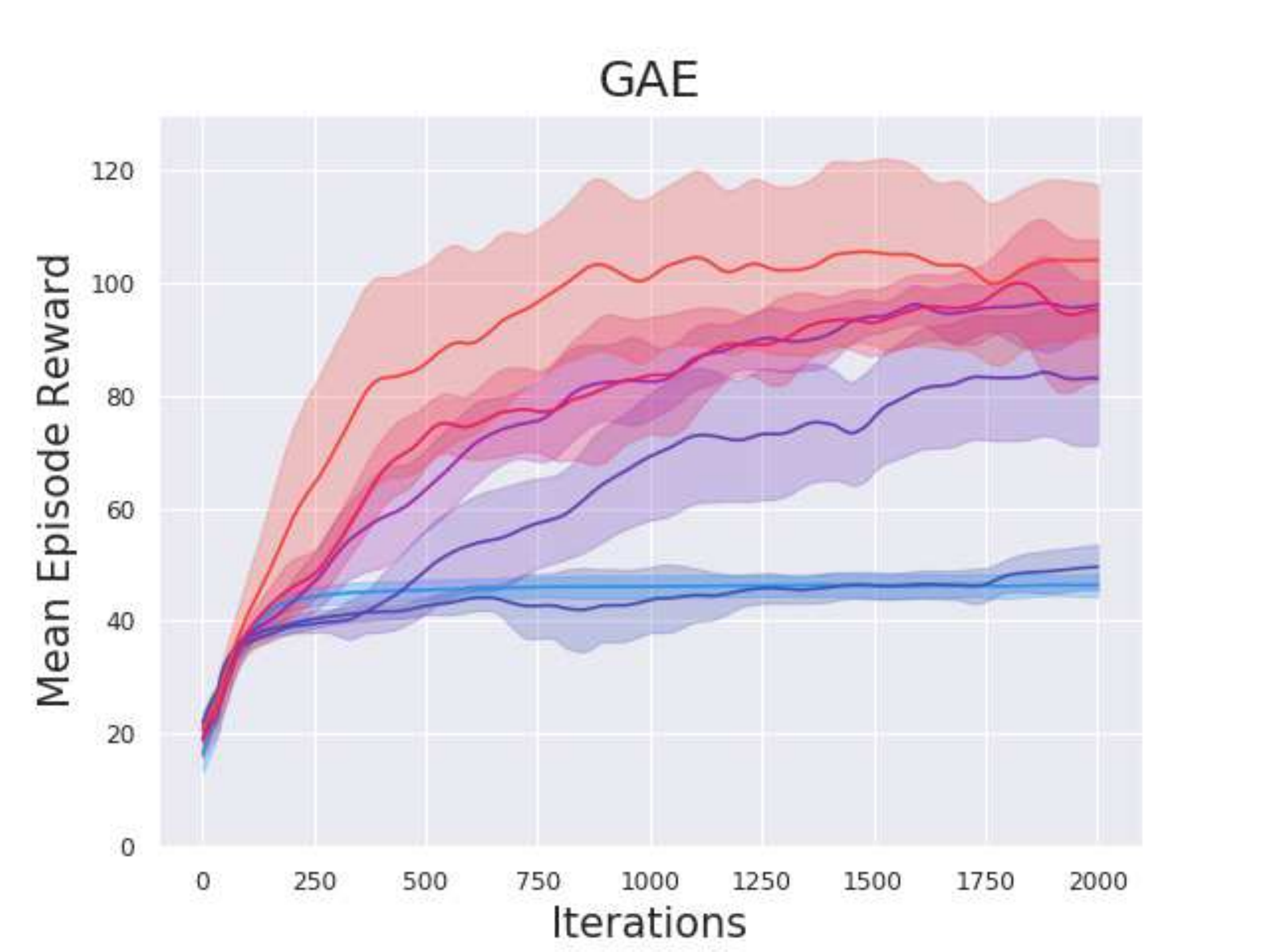}
    \hspace{-0.6cm}\includegraphics[scale=0.20]{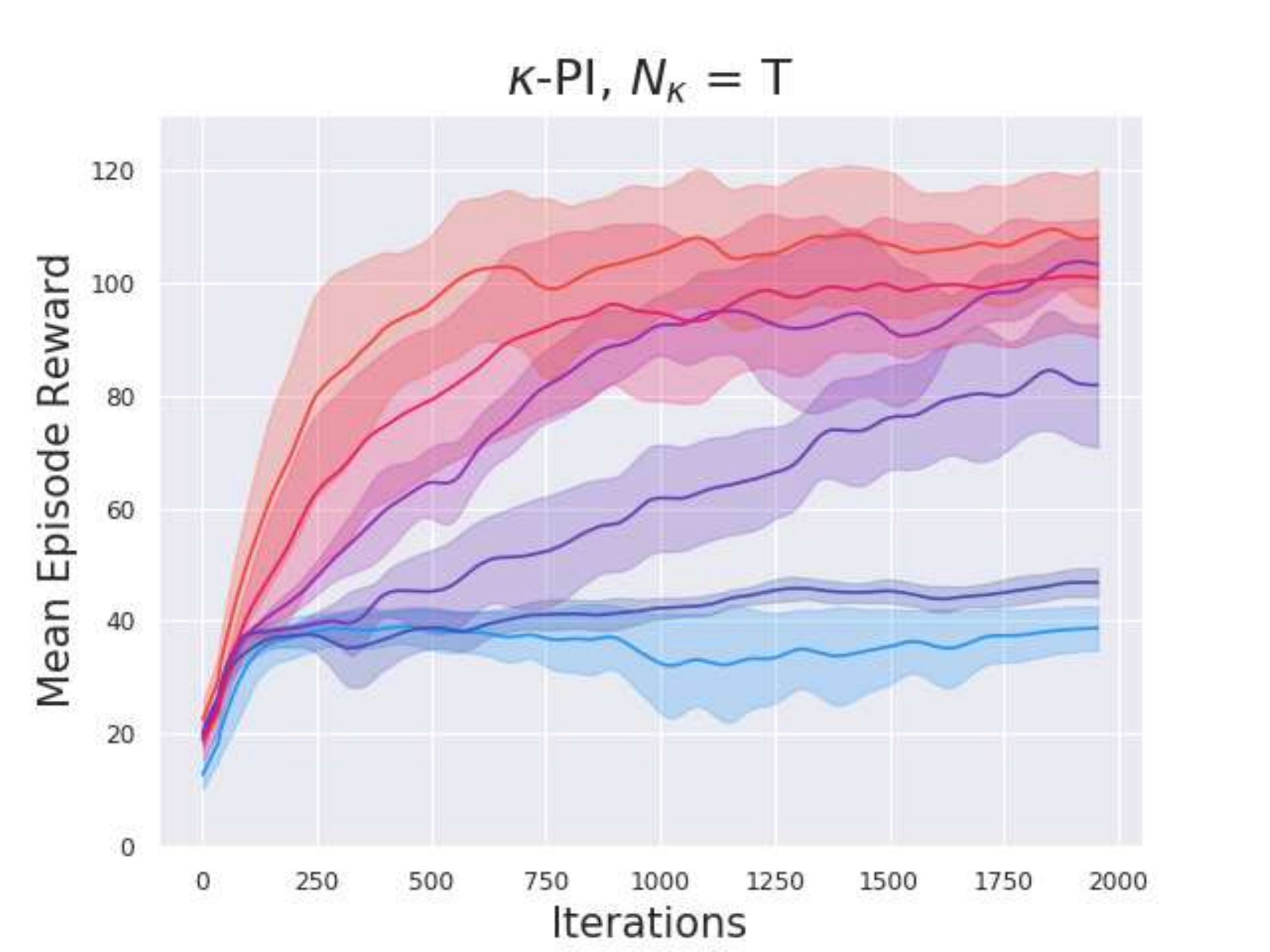} 
    \hspace{-0.6cm}\includegraphics[scale=0.20]{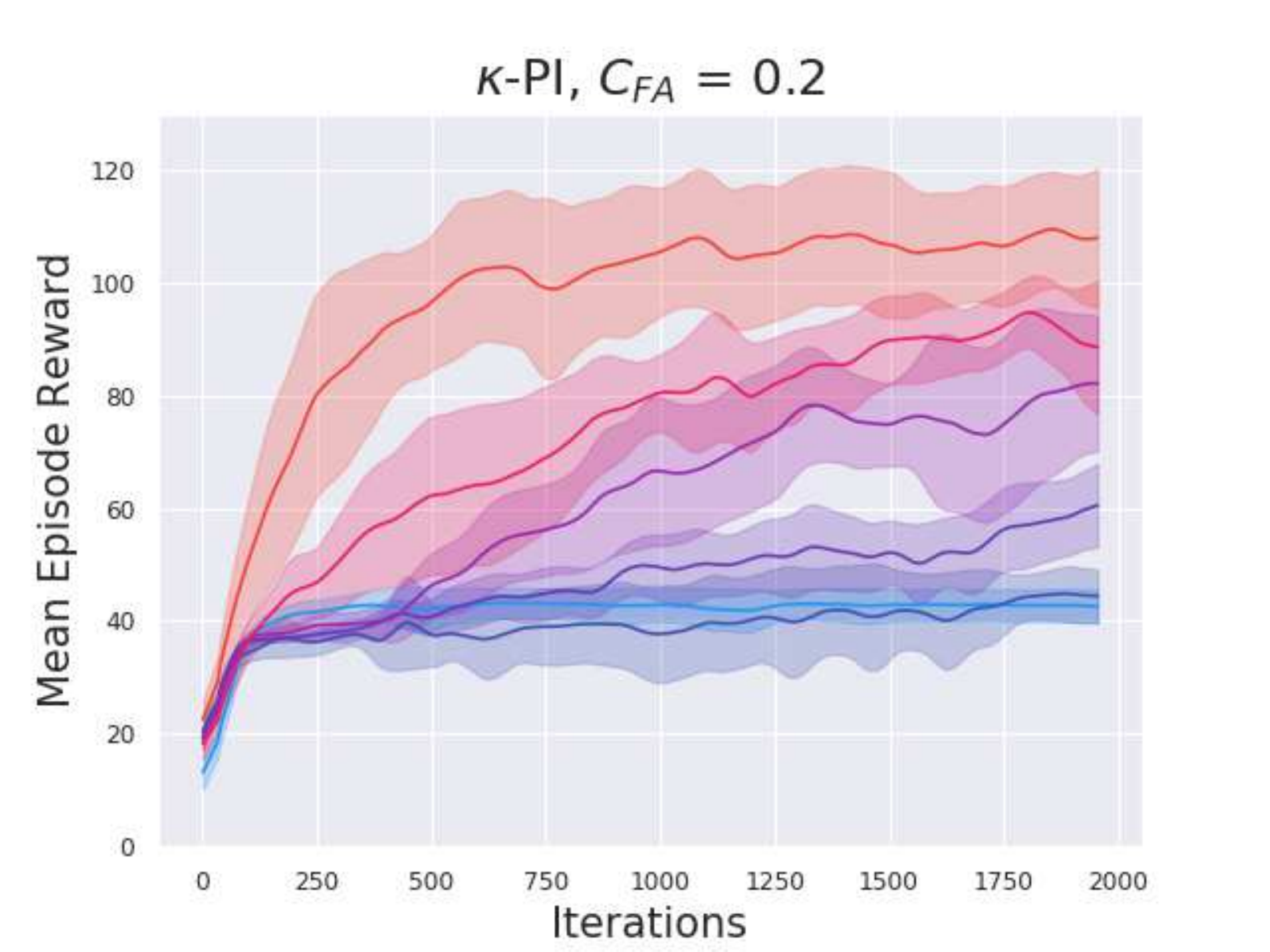}
    \hspace{-0.6cm}\includegraphics[scale=0.20]{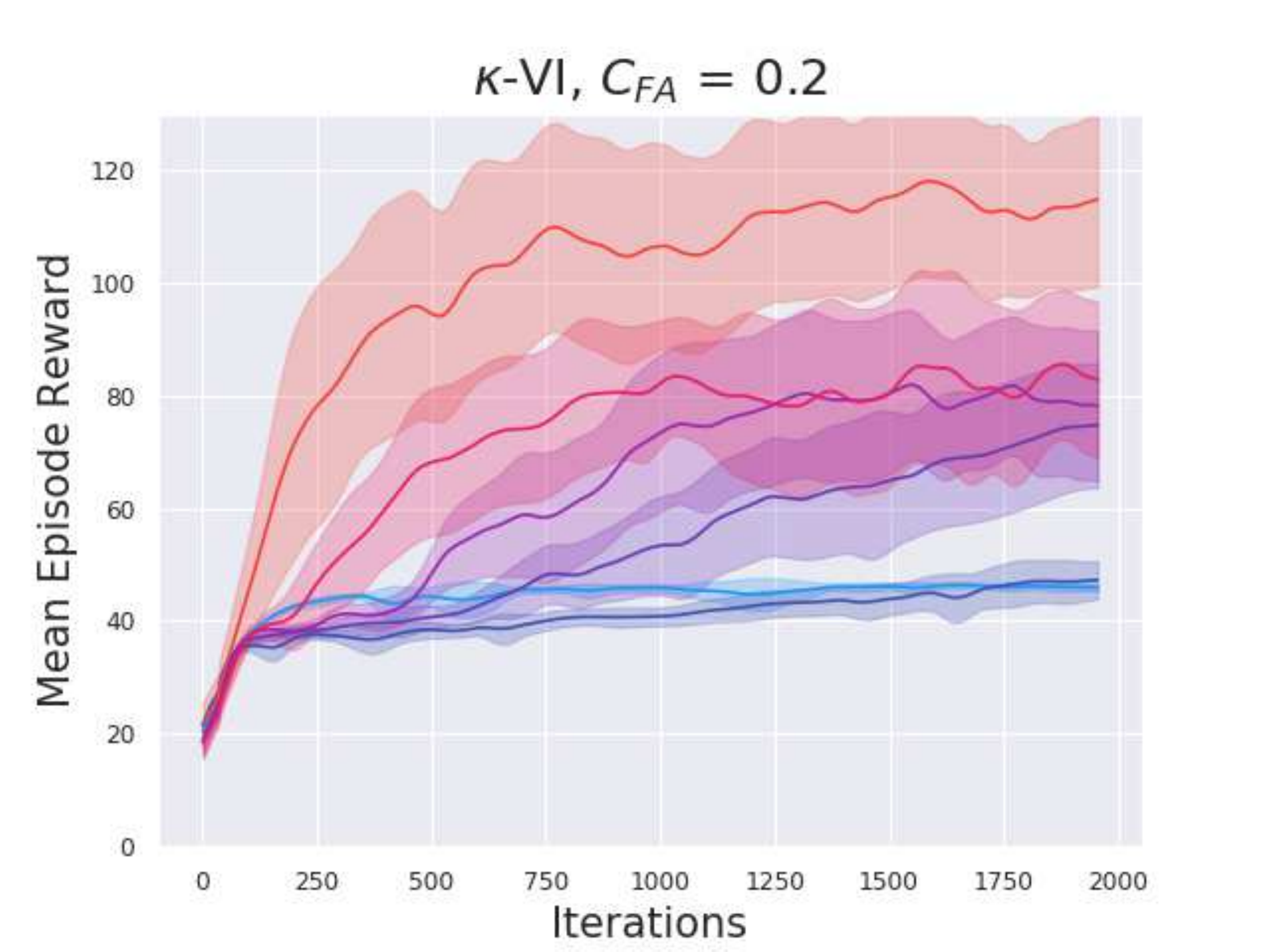} \\
    \centering
    \includegraphics[scale=0.3]{Images/TRPO/HumanoidStandup_legend_horizontal.pdf}
    \caption{Performance of GAE, `Naive' baseline and $\kappa$-PI-TRPO, $\kappa$-VI-TRPO on Swimmer-v2.}
    \label{fig:Swimmer}
\end{figure*}

\begin{figure*}[ht]
    \centering
    \hspace{-0.6cm}\includegraphics[scale=0.20]{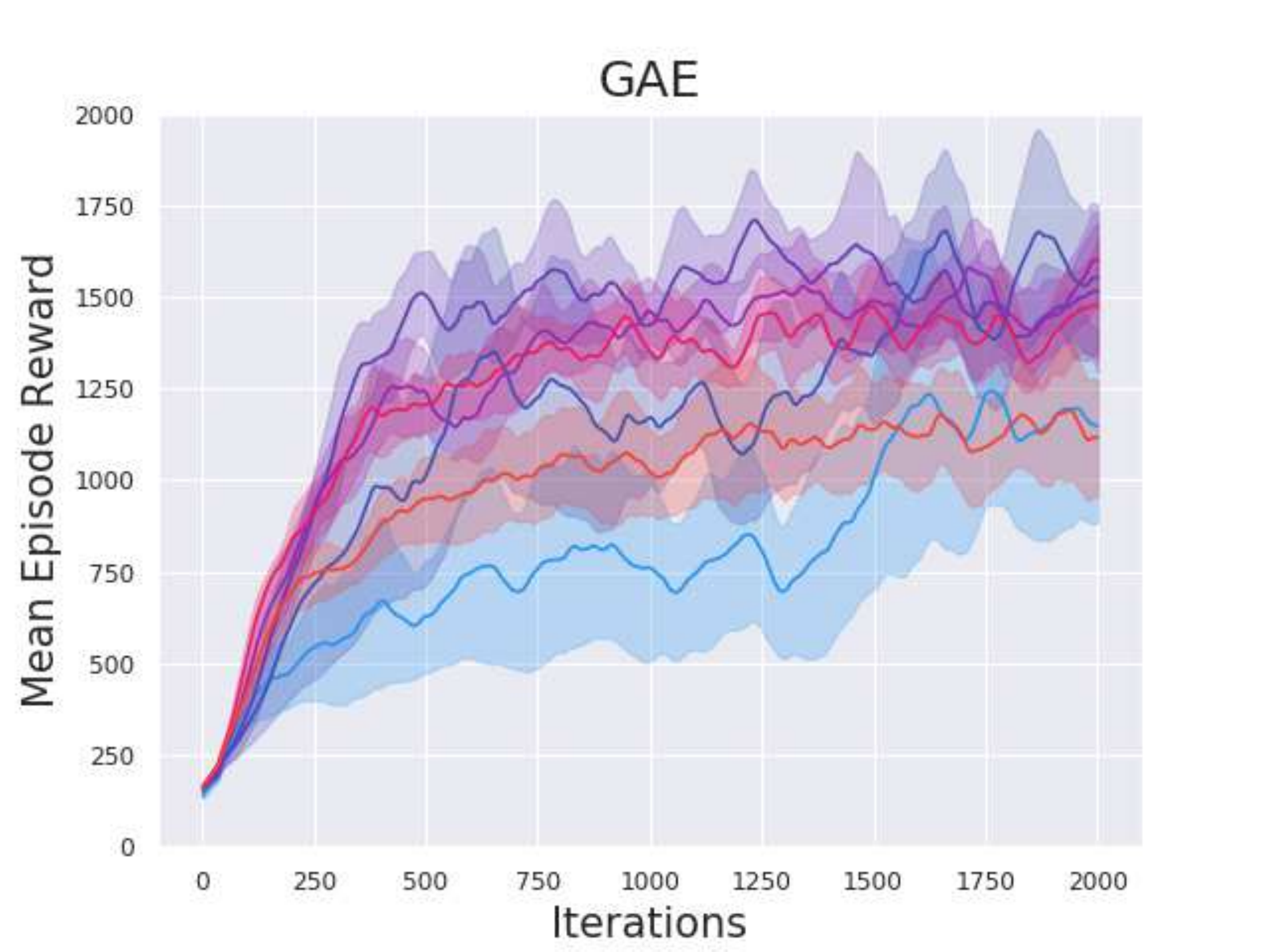}
    \hspace{-0.6cm}\includegraphics[scale=0.20]{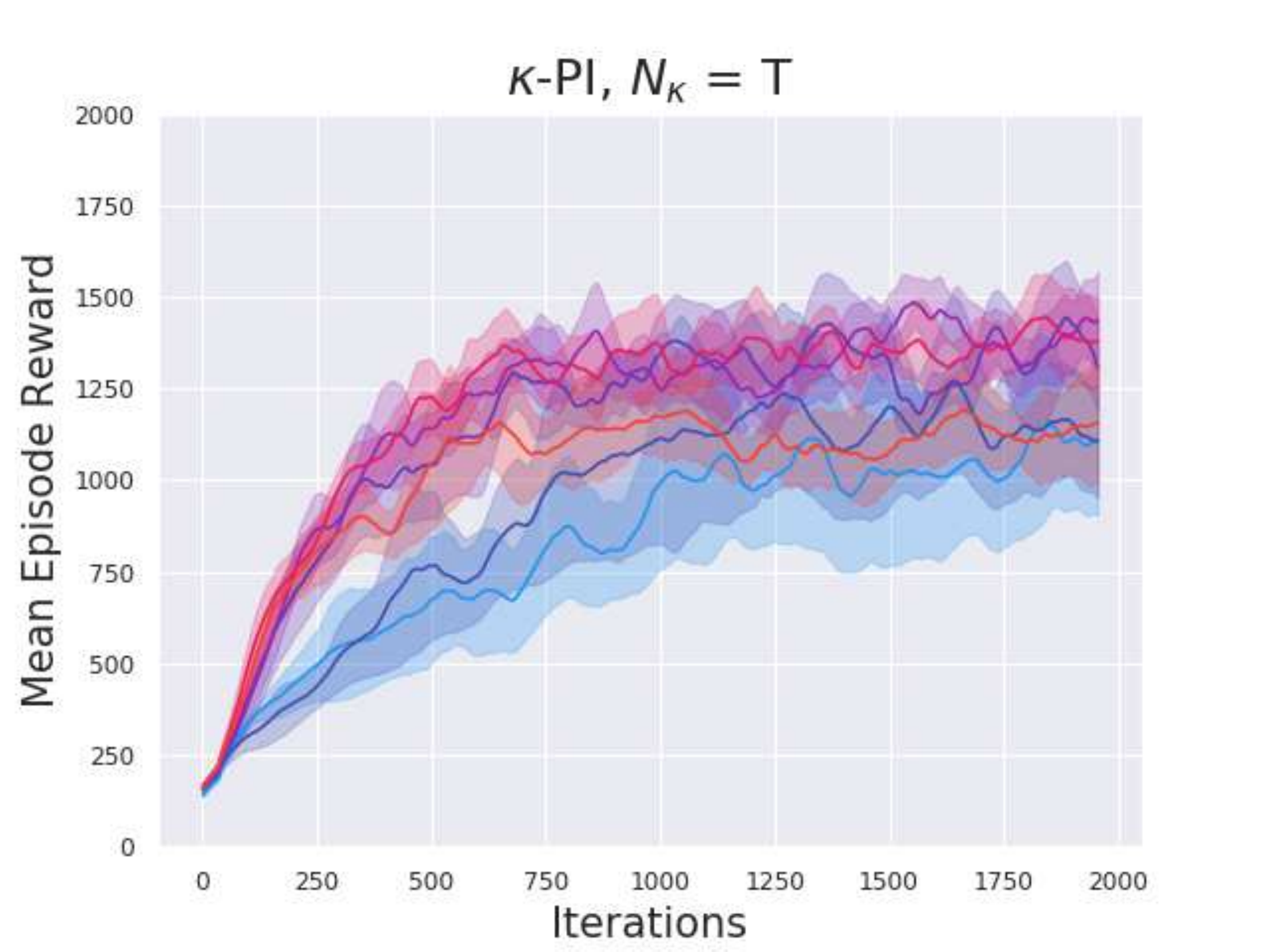} 
    \hspace{-0.6cm}\includegraphics[scale=0.20]{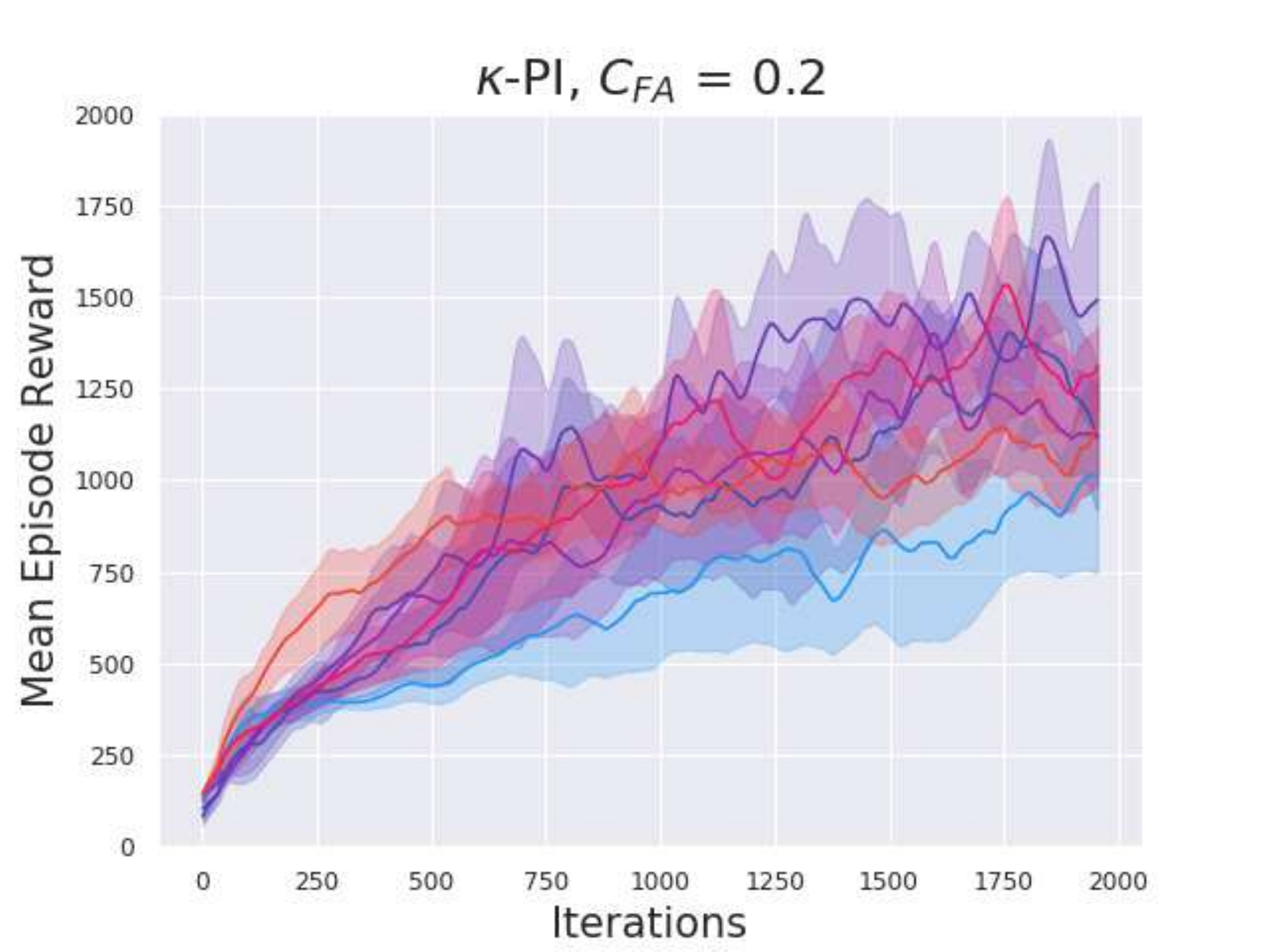}
    \hspace{-0.6cm}\includegraphics[scale=0.20]{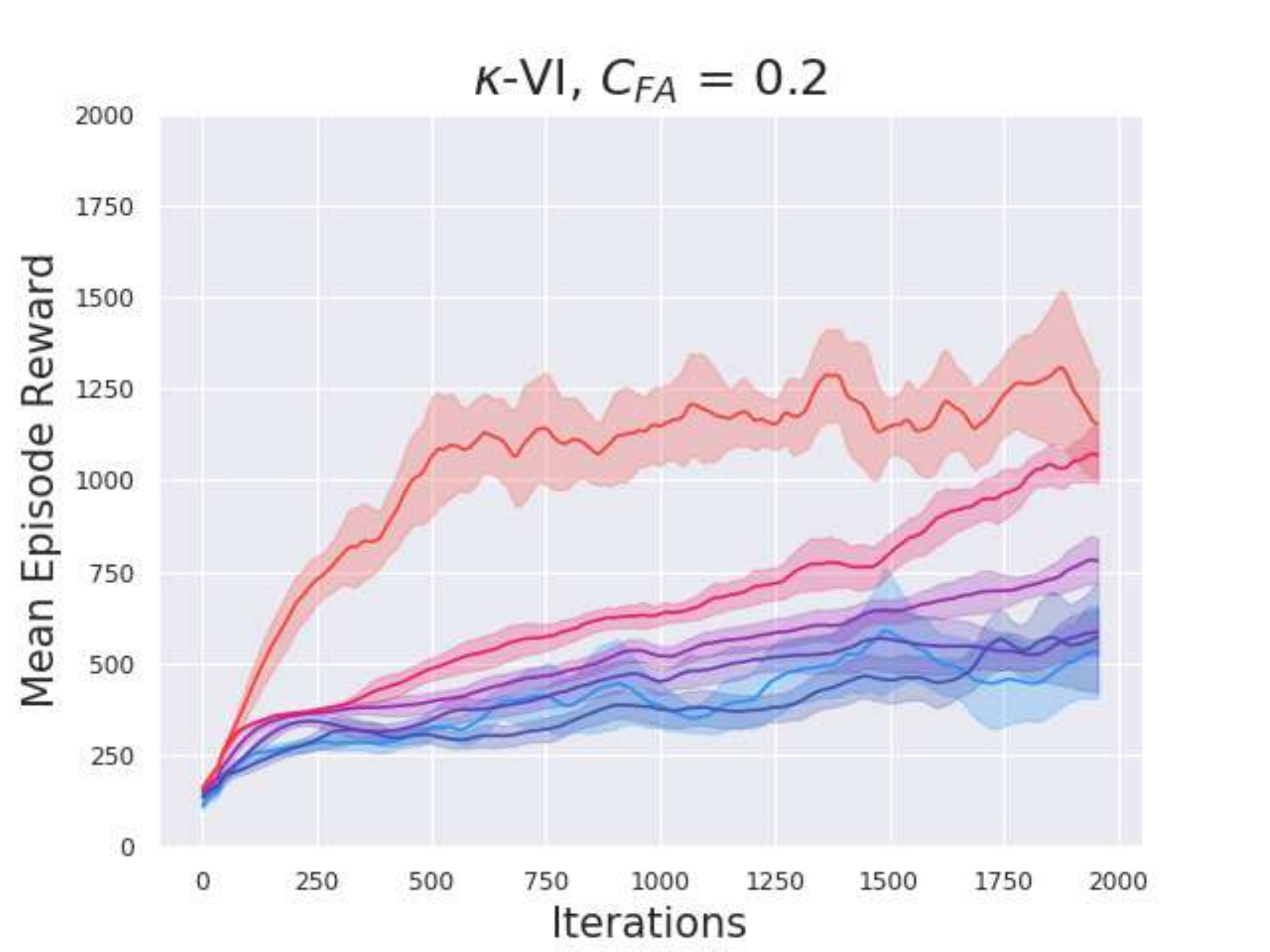} \\
    \centering
    \includegraphics[scale=0.3]{Images/TRPO/legend_horizontal.pdf}
    \caption{Performance of GAE, `Naive' baseline and $\kappa$-PI-TRPO, $\kappa$-VI-TRPO on Hopper-v2.}
    \label{fig:Hopper}
\end{figure*}

\newpage


\end{document}